\documentclass[runningheads]{llncs}
\pdfoutput=1

\usepackage{eccv}

\usepackage{eccvabbrv}

\usepackage{graphicx}
\usepackage{booktabs}
\usepackage{siunitx}
\usepackage{caption}
\usepackage{subcaption}
\usepackage{placeins}
\usepackage{dirtytalk}
\usepackage{wrapfig}
\usepackage{xcolor}
\usepackage{colortbl}
\usepackage{makecell}
\usepackage{multirow}
\usepackage{bm}

\usepackage[accsupp]{axessibility}  

\usepackage{hyperref}

\usepackage{orcidlink}

\newcommand{\std}[1]{\textcolor[gray]{0.3}{\tiny$\pm$#1}}
\newcommand{\training}[1]{\textsuperscript{\tiny\textbf{#1}}}
\newcommand{\nih}{\training{\textcolor{RedOrange}{N}}}
\newcommand{\vindr}{\training{\textcolor{OliveGreen}{V}}}
\newcommand{\mimic}{\training{\textcolor{MidnightBlue}{M}}}
\newcommand{\other}{\training{\textcolor{Black}{O}}}
\definecolor{figBlue}{HTML}{6C8EBF}
\definecolor{figPurple}{HTML}{9673A6}
\definecolor{figGreen}{HTML}{82B366}
\definecolor{figYellow}{HTML}{D6B656}

\begin{document}

\title{ChEX: Interactive Localization and Region Description in Chest X-rays} 

\titlerunning{ChEX}

\author{Philip Müller\inst{1}\orcidlink{0000-0001-8186-6479} \and
Georgios Kaissis\inst{1,2}\orcidlink{0000-0001-8382-8062} \and
Daniel Rueckert\inst{1,3}\orcidlink{0000-0002-5683-5889}}

\authorrunning{P.~Müller et al.}
\institute{Technical University of Munich, Germany \and
Helmholtz Munich, Germany \and
Imperial College London, UK\\
\email{\{philip.j.mueller,g.kaissis,daniel.rueckert\}@tum.de}}

\maketitle

\begin{abstract}
Report generation models offer fine-grained textual interpretations of medical images like chest X-rays, yet they often lack \emph{interactivity} (\ie the ability to steer the generation process through user queries) and \emph{localized interpretability} (\ie visually grounding their predictions), which we deem essential for future adoption in clinical practice. 
While there have been efforts to tackle these issues, they are either limited in their interactivity by not supporting textual queries or fail to also offer localized interpretability. 
Therefore, we propose
a novel multitask architecture and training paradigm integrating textual prompts and bounding boxes for diverse aspects like anatomical regions and pathologies. We call this approach the \emph{Chest X-Ray Explainer (ChEX)}. Evaluations across a heterogeneous set of 9 chest X-ray tasks, including localized image interpretation and report generation, showcase its competitiveness with SOTA models while additional analysis demonstrates ChEX’s interactive capabilities. Code: \url{https://github.com/philip-mueller/chex}.
\end{abstract}

\section{Introduction}
The automatic interpretation of medical images, such as chest X-rays, holds immense promise for enhancing healthcare. 
Report generation models enable detailed textual interpretations beyond the capabilities of traditional image classification techniques alone. 
Despite significant progress in enhancing their accuracy\cite{maira}, incorporating them in clinical practice
remains a formidable challenge. A primary concern stems from the lack of transparency and interpretability surrounding the decision-making processes of these models, which poses obstacles for medical professionals seeking to validate their predictions, thereby hindering widespread adoption\cite{miller2019explanation,geis2019ethics}.
Additionally, the non-interactive nature of most current models introduces risks in cases of inaccurate predictions, while interactive involvement in the generation process, \eg through user prompts, may encourage the user to manually intervene in such cases.

Therefore, we advocate for two important aspects that can enhance the utility of these models in clinical practice: \emph{(localized) interpretability and interactivity}.
One pioneering effort in this direction is the work of Tanida \etal \cite{rgrg}. By incorporating bounding boxes of anatomical regions into the report generation process, their model offers enhanced interpretability. 
Furthermore, it supports bounding boxes as input queries, facilitating user interactivity. 
However, the model exhibits two major limitations: it lacks support for textual user prompts and focuses exclusively on anatomical regions during both training and inference. This narrow focus can lead to suboptimal predictions for other aspects, such as pathologies. 
In contrast, models like RaDialog \cite{radialog}, Med-PaLM M \cite{medpalmm}, and OmniFM-DR \cite{omnifm} support interactivity through textual prompts but do not provide bounding boxes for their textual answers or support them as queries.

\begin{figure}[t]
    \centering
    \includegraphics[width=.95\textwidth]{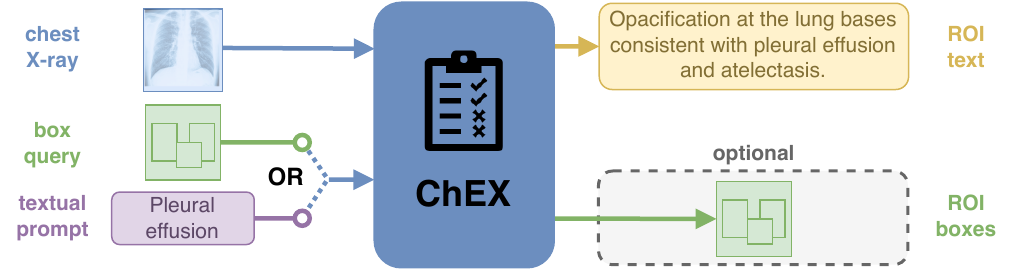}
    \caption{
    Overview of ChEX. Given a chest X-ray and a user query, either as a textual prompt (\eg, a pathology name, an anatomical region, or both) or as a bounding box, the model predicts a textual description of the queried region or aspect. For textual user prompts, it additionally predicts relevant bounding boxes. Thus, ChEX facilitates the interactive interpretation of chest X-rays while providing (localized) interpretability.}
    \label{fig:main}
\end{figure}

In this work, we address these limitations by proposing a novel multitask architecture and training paradigm. Our approach integrates textual prompts and bounding boxes for various aspects, including anatomical regions, pathologies, and report sentences. We call this approach the \emph{Chest X-Ray Explainer (ChEX)}. As illustrated in Figure \ref{fig:main}, ChEX can be queried using textual prompts or user-defined bounding boxes and predicts individual descriptions for each specified region or aspect. For textual queries, the predictions are supplemented by bounding boxes to localize relevant regions for each answer. 
Therefore, ChEX offers a unique combination of interactivity -- \emph{reacting to user prompts} -- and interpretability -- \emph{visually grounding the answers} -- not provided by other report generation models. Furthermore, it additionally supports localized tasks beyond report generation (RG), including pathology (object) detection (OD), sentence grounding (SG), region classification (RC), and region explanation (RE).

Our contributions are as follows:
\begin{itemize}
    \item We propose ChEX, an interactive and interpretable model for predicting visually grounded textual descriptions of chest X-rays based on user queries.
    \item We propose a multitask training paradigm, enabling ChEX to be jointly trained with diverse types of supervision from different datasets. 
    \item We evaluate ChEX across 9 diverse chest X-ray tasks, 
    spanning localized image interpretation and report generation functionalities. ChEX demonstrates competitive performance with specialized and general state-of-the-art (SOTA) models 
    despite being significantly smaller than some of these models, which have up to 80 times the size of ChEX. Notably, none of the baseline models is capable of performing all the tasks covered by ChEX.
    \item We conduct a thorough analysis of ChEX's interactive capabilities, demonstrating its proficiency in responding to specific user prompts. 
\end{itemize}

\section{Related Work}
\subsubsection{Medical Image-Text Models}
One prevalent category of medical image-text models adopts a CLIP-style \cite{clip} framework, leveraging contrastive image-text learning 
\cite{gloria,convirt,lovt,biovil,chexzero,liao_2021,seibold_2022,wang_2022_neurips,wang2022medclip}. 
While most works focus on global image-text alignment, only some works consider more localized elements such as individual sentences or image patches \cite{gloria,lovt,biovil,liao_2021,seibold_2022,wang_2022_neurips}. 

Report generation has recently received significant attention \cite{promptmrg,hou_2023_organ,Sun_2022,rgrg,Gu_2024_comg,radialog,medpalmm,omnifm,li_2023_cvpr,
maira,wang_metrans_2023,
nicolson_2023,miura2021improving,wang2022inclusive}.
Some of these approaches target specific aspects such as pathologies \cite{promptmrg,hou_2023_organ,Sun_2022}, anatomical regions \cite{rgrg,Gu_2024_comg}, or both \cite{li_2023_cvpr}. However, support for interactive user queries remains rare among report generation models, with only the model RGRG \cite{rgrg} enabling queries via bounding boxes and a few others \cite{radialog,medpalmm,omnifm} supporting textual queries, albeit without the ability to jointly predict bounding boxes and textual descriptions. Notably, the model OmniFM-DR \cite{omnifm} supports the prediction of bounding boxes for textual prompts but without describing the content therein.

The ability of responding to textual prompts characterizes visual question answering (VQA) models, with some multitask models \cite{medpalmm,omnifm,llavamed,xu2023elixr} supporting zero-shot medical VQA, while others require fine-tuning \cite{zhang2024biomedgpt,he2024pefomed,sonsbeek_2023,liu_2023_vqa,eslami_2021}. However, unlike ChEX, these models lack localization capabilities for their responses. Moreover, it is important to note that while these methods rely on question-answer pairs for training, ChEX indirectly acquires its interactive capabilities through multitask training. 
Although localization tasks have been integrated into multitask training for natural images \cite{beit3,glip2,yang2022_unitab},
such approaches remain scarce in the medical domain, with OmniFM-DR \cite{omnifm} being a notable exception.

\subsubsection{Prompt-based Localization}
DETR \cite{detr} pioneered the use of token vectors for object detection, sparking a series of subsequent models following this approach \cite{codetr,zhu2020deformable,liu2022dabdetr,conditional-detr,samdetr}. However, these models typically employ tokens that are not input-dependent, often relying on learned or position-based tokens \cite{liu2022dabdetr,samdetr,conditional-detr}.
Visual grounding models \cite{transvg,transvgpp,du2022visual,li2021referring} predict bounding boxes for given textual phrases, relating them to image regions, and have been applied in medical imaging \cite{chen2023_miccai,ichinose2023_miccai}. GLIP \cite{glip,glip2} unifies object detection and visual grounding using textual prompts, with applications emerging in the medical domain \cite{guo_2023_miccai,wu_2023_miccai}.
Open-vocabulary object detection extends object detection models to unseen classes using textual prompts \cite{ovr,regionclip,detic,cora,vild,ovdetr,liu2023grounding,Hanoona2022Bridging,Maaz2022Multimodal}. Similarly, prompt-based segmentation techniques have been explored \cite{sam} with applications in the medical domain \cite{huang2024_mia,ma2024segment}.

\section{ChEX: A Localized and Interactive Chest X-ray Description Model}

\begin{figure}[t]
    \centering
    \includegraphics[width=.95\textwidth]{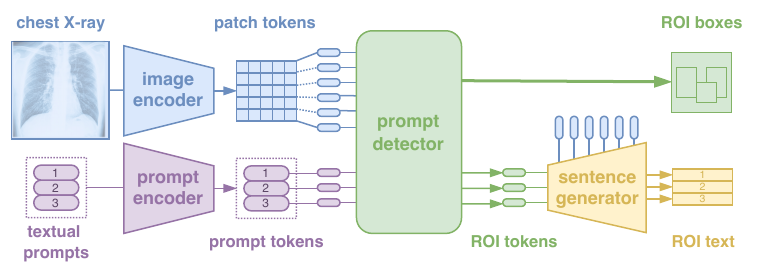}
    \caption{Architecture of ChEX. The DETR-style \textcolor{figGreen}{prompt detector} predicts bounding boxes and features for ROIs based on prompt tokens (textual prompts encoded by the \textcolor{figPurple}{prompt encoder}) and patch features (from the \textcolor{figBlue}{image encoder}). The \textcolor{figYellow}{sentence generator} is then used to predict textual descriptions for each ROI independently.}
    \label{fig:architecture}
\end{figure}

Our model ChEX supports a wide range of tasks, spanning from localization to text generation, while considering different aspects like anatomical regions and pathologies. 
To facilitate efficient training and zero-shot inference across all tasks, ChEX employs a simple yet versatile architecture, outlined in Figure \ref{fig:architecture}.
First, an \textcolor{figBlue}{\textbf{image encoder}} is used to extract patch tokens of the given chest X-ray, while each textual prompt is encoded using the \textcolor{figPurple}{\textbf{prompt encoder}}, a frozen text encoder. Next, the \textcolor{figGreen}{\textbf{prompt detector}}, a DETR\cite{detr}-style object detector, localizes the prompts within the image,
predicting a set of bounding boxes for each prompt along with a single Region-of-interest (ROI) token vector per prompt.
Finally, the \textcolor{figYellow}{\textbf{sentence generator}} 
is conditioned on each ROI token independently as well as on all patches to predict a concise description of each ROI. 
Further details are given in the following \cref{sec:architecture,sec:training,sec:inference}. 
For reproducibility, we provide comprehensive implementation details in the supp.\@ material.

\subsection{Model Architecture}\label{sec:architecture}

\subsubsection{Image and Prompt Encoder}
We use CLIP\cite{clip} with ViT-B/32\cite{vit} and the default text encoder, both pre-trained\cite{chexzero} on chest X-ray/report pairs from MIMIC-CXR\cite{mimic-cxr}.
Instead of using the \texttt{CLS} token, we use all patch tokens from the ViT and project them individually into the shared image-text space using the pre-trained projector from CLIP. 
Given $Q$ textual prompts, each of them is embedded independently using the prompt encoder, \ie the text encoder from CLIP, and projected to the shared image-text space.
During training, we freeze the complete prompt encoder and the image encoder up to the 8th encoder layer.

\subsubsection{Prompt Detector}
Using the $Q$ prompt tokens from the output of the prompt encoder and the patch tokens from the image encoder, the prompt detector independently localizes each prompt by predicting $M=3$ bounding boxes per prompt, following the DETR \cite{detr} decoder with 6 layers. To support multiple boxes per prompt, 
we adopt \cite{ovdetr} and additively combine each prompt token with $M$ learned tokens, leading to a total of $Q \times M$ decoder query tokens. 
After decoding, we apply an MLP-based box predictor on each token feature to predict individual bounding boxes and box scores. We then use the bounding box parameters as a regional bias and compute box features by applying Gaussian ROI pooling \cite{wsrpn} on the patch tokens. Additionally, we introduce a random skip connection to provide a direct path from the decoder layers. 
Finally, we compute the $Q$ ROI tokens by aggregating the $M$ box features per prompt using a weighted average based on box scores. 
When using bounding box queries instead of textual prompts, ROI features are directly computed with Gaussian ROI pooling on patch tokens using the given bounding box parameters and bypassing all decoder layers.

\subsubsection{Sentence Generator}
For sentence generation, we use the GPT2-medium \cite{radford2019language} model pre-trained on PubMed abstracts and condition it on each ROI token independently using P-tuning v2 \cite{p-tuning} with MLP projection and without freezing parameters. To incorporate additional global context, we apply a \emph{post decoder} comprising three transformer decoder layers, with the ROI tokens as queries and the patches as keys and values before feeding the features into the GPT2 model.

\subsection{Multitask Training}\label{sec:training}
We train ChEX in a multitask setting, where each sample provides one or more types of targets, including bounding boxes for pathologies or anatomical structures, pathology classification labels (per sample or per region), and report sentences (per sample or per region). To enable training with such a wide range of targets, we use three types of prompt tokens:
\begin{enumerate}
    \item \textbf{Pathology tokens:} We define textual prompts for each pathology class (\eg, \say{pleural effusion}) and encode them using the prompt encoder.
    \item \textbf{Anatomy tokens:} We define textual prompts for each anatomical region (\eg, \say{right lung}) and encode them using the prompt encoder.
    \item \textbf{Sentence tokens:} We encode each individual sentence of the radiology report provided with the current sample.
\end{enumerate}
For each sample, we only use the token types for which there are targets available. 
Using the encoded prompt tokens, the prompt detector predicts bounding boxes and ROI tokens for each of them, before the sentence generator (conditioned on the ROI tokens) predicts the target sentences.

ChEX does not only support textual queries, \ie prompts, but also bounding box queries. Therefore, in some (randomly selected) batches, we train ChEX with query bounding boxes instead of prompt tokens. In such cases, we compute the ROI tokens based on the target bounding boxes directly using Gaussian ROI pooling, skipping the textual tokens and box prediction process.

\subsubsection{Loss Functions}
We use the following loss functions to train our model:
\begin{enumerate}
    \item \textbf{Bounding boxes:} 
    We use a modified DETR loss \cite{detr}, applying Hungarian matching independently per \emph{pathology} or \emph{anatomy} token until all predicted boxes are matched. We retain the L1 and gIoU losses, but omit the cross-entropy loss. Box scores are instead trained only for pathologies, using the focal loss \cite{focalloss} with positive targets for boxes matched in the first iteration.
    \item \textbf{Pathology class labels:} 
    We use an  InfoNCE contrastive loss \cite{CPC}, pairing ROI tokens with textual pathology prompts according to the class labels.
    For example, if the pathology pleural effusion is present in the image (\emph{pathology token}) or region (\emph{anatomy token}), we build a positive pair with the prompt \say{pleural effusion}, while negative pairs use the prompt \say{no pleural effusion} and prompts of non-present pathologies (\eg, \say{pneumothorax}, \dots).
    \item \textbf{Report sentences:} We apply autoregressive language modeling to predict region sentences (\emph{anatomy token}) or to reconstruct the sentence (\emph{sentence token}). For \emph{sentence tokens}, we additionally use contrastive learning between ROI tokens  and their corresponding textual sentence features. We also apply the CLIP loss \cite{clip} on average-pooled image patch and sentence features.
\end{enumerate}

\subsection{Zero-shot Inference}\label{sec:inference}
During inference, textual prompts like predefined pathology names, sentences, or user queries are encoded, and the prompt detector predicts bounding boxes and ROI tokens. These tokens are used by the sentence generator to predict descriptions of the detected regions. If regions require classification, such as for object detection or region classification, ROI tokens are classified based on cosine similarity with encoded prompts. When bounding boxes are given as queries, they're used for Gaussian ROI pooling, omitting other parts of the prompt detector and encoder if no textual prompts are provided. For full report generation, we use pre-defined sets of textual prompts and concatenate the predicted descriptions.

\section{Experimental Setup and Evaluation}
\subsection{Training Dataset and Pre-processing}
We train on the frontal chest X-rays from MIMIC-CXR\cite{mimic-cxr,mimic-cxr-2,physionet} and VinDr-CXR\cite{vindr,vindr2,physionet}. 
MIMIC-CXR comes paired with radiology reports, of which we use the findings and impression sections and split them into individual sentences. We use additional supervision for the MIMIC-CXR images provided by the Chest ImaGenome (CIG) \cite{cig,cig2,physionet} dataset. 
It includes bounding boxes for 29 unique
anatomical regions and additionally assigns report sentences as well as 53 unique findings and pathology labels to each of these regions. VinDr-CXR includes bounding boxes for 22 unique findings and pathologies. Overall, we train on 227,382 images from MIMIC-CXR and 15,000 images from VinDr-CXR, from their official train splits, but oversample VinDr-CXR samples to simulate equal size of both datasets.
We randomly crop and resize all images to a resolution of $224\times224$ and then apply random horizontal flips, random affine transformations, contrast/brightness jittering, and random Gaussian blurring.

\subsection{Benchmark Tasks}
We evaluate the zero-shot performance of our model ChEX across the 9 chest X-ray tasks shown in \cref{tab:tasks}. 
For details on the evaluation and dataset preparation, we refer to the supp.\@ material.
While no task-specific fine-tuning is applied, 
post-processing (\eg box suppression and scaling of boxes) for ChEX and all baselines is adjusted to consider differences in annotation practices of datasets.
\begin{table}[t]
    \centering
    \caption{Benchmark tasks with their datasets and evaluation metrics}
    \label{tab:tasks}
    \scriptsize
    \setlength{\tabcolsep}{5pt}
    \renewcommand{\arraystretch}{1.3}
    \renewcommand\cellgape{\Gape[1.5pt]}
    \begin{tabular}{lrrl}
    \toprule
    \textbf{Task Dataset} & \textbf{\#Samples} & \textbf{\#Classes} & \textbf{Evaluation Metrics} \\
    \midrule
    \rowcolor[gray]{.9}\multicolumn{4}{l}{\textbf{Sentence Grounding (SG):} \textcolor[gray]{.4}{Predicting bounding boxes for given sentences}} \\
    MS-CXR\cite{biovil} & 169 & none & mIoU, mAP \\
    \rowcolor[gray]{.9}\multicolumn{4}{l}{\textbf{Pathology Detection (OD):} \textcolor[gray]{.4}{Object detection of pathologies}}\\
    VinDrCXR\cite{vindr} & 1,500 & top 15 & mAP \\
    NIH ChestXray (NIH8)\cite{nih8} & 448 & 8 & mAP \\
     MS-CXR\cite{biovil} & 169 & 8 & mAP \\
    \rowcolor[gray]{.9}\multicolumn{4}{l}{\textbf{Region Classification (RC):} \textcolor[gray]{.4}{Classifying regions defined by given bounding boxes}} \\
    MS-CXR\cite{biovil} & 169 & 8 & AUROC \\
    Chest ImaGenome (CIG)\cite{cig} & 3,402 & 53 & weighted AUROC (wAUROC) \\
    \rowcolor[gray]{.9}\multicolumn{4}{l}{\textbf{Region Explanation (RE):} \textcolor[gray]{.4}{Predicting descriptions for regions defined by  bounding boxes}} \\
    MS-CXR\cite{biovil} & 169 & none & \makecell[l]{METEOR\cite{meteor}\\Mic-F1-14\textsuperscript{\textdagger}, Mac-F1-14\textsuperscript{\textdagger}} \\
    Chest ImaGenome (CIG)\cite{cig} & 3,402 & none & \makecell[l]{METEOR\cite{meteor}\\Mic-F1-14\textsuperscript{\textdagger}, Mac-F1-14\textsuperscript{\textdagger}} \\
    \rowcolor[gray]{.9}\multicolumn{4}{l}{\textbf{Full Report Generation (RG):} \textcolor[gray]{.4}{Predicting full reports from chest X-rays}} \\
    MIMIC-CXR\cite{mimic-cxr} & 3,082 & none & \makecell[l]{METEOR\cite{meteor}\\Mic-F1-14\textsuperscript{\textdagger}, Mac-F1-14\textsuperscript{\textdagger}, Ex-F1-14\textsuperscript{\textdagger}\\Mic-F1-5+\textsuperscript{\textdagger}, Mac-F1-5+\textsuperscript{\textdagger}} \\
    \bottomrule
    \end{tabular}\\
    \textsuperscript{\textdagger} Clinical efficacy (CE) metrics based on the CheXbert\cite{chexbert} classifier, micro-, macro-, and example-level-averaged over all 14 classes following Nicolson \etal\cite{nicolson_2023} (Mic-F1-14, Mac-F1-14, Ex-F1-14), and averaged over 5 classes following Miura \etal\cite{miura2021improving} (Mic-F1-5+, Mac-F1-5+).
\end{table}

\subsection{Benchmark Baselines}
We consider a wide variety of specialist and generalist baselines in our benchmark. 
For SG, we consider the standard supervised visual grounding (SupVG) model TransVG\cite{transvg} and the generative multitask model OmniFM-DR\cite{omnifm}.
For OD and RC tasks, we compare against standard supervised object detection (SupOD) models including Faster R-CNN\cite{frcnn} and DETR-style models \cite{detr,conditional-detr,codetr,zhu2020deformable}, and weakly-supervised object detection (WSupOD) including ADPD\cite{adpd} and CheXNet\cite{chexnet}, each of them with different (pre-trained) backbones. Additionally, we study the zero-shot capabilities of standard CLIP-style models for chest X-rays (BioVIL\cite{biovil} and CheXzero\cite{chexzero}) on all SG, OD, and RC tasks. 
To the best of our knowledge, there is only one model supporting the RE tasks out-of-the-box, namely RGRG\cite{rgrg}. We also use the two CLIP-style baselines as image-text retrieval models and evaluate their performance on these tasks. For RG, we compare against several report generation models, including recent models like MAIRA-1\cite{maira}, Med-PaLM M \cite{medpalmm}, Prompt-MRG\cite{promptmrg}, and RaDialog\cite{radialog}. 

\section{Results}
We compare ChEX with SOTA models (\cref{sec:results_sota}). Additionally, we study how ChEX reacts to (interactive) user queries (\cref{sec:inter_prompting,sec:precise_prompts}), show its interpretable and customizable report generation capabilities (\cref{sec:custom_reportgen}), and study technical design choices (\cref{sec:technical_insights}).
ChEX is competitive with SOTA models while providing a high degree of interpretability and interactivity, therefore offering a promising path towards clinical application.

\begin{wrapfigure}[27]{R}{.5\textwidth}
\centering 
  \includegraphics[width=.49\textwidth,trim=0 0 0 108]{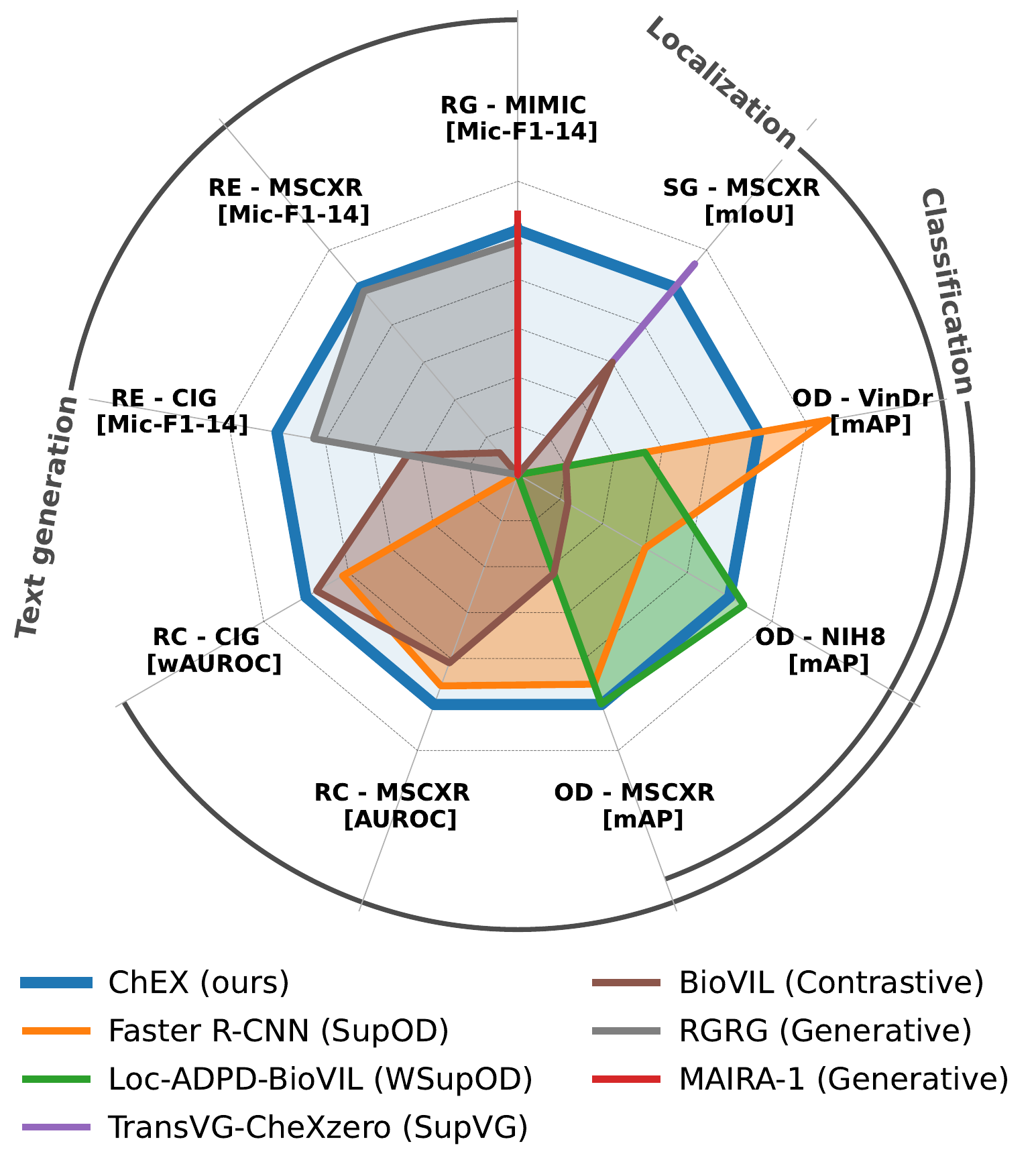}
  \caption{Comparison of ChEX with specialized SOTA and common multitask models on 9 chest X-ray tasks, 
  including sentence grounding (SG), pathology detection (OD), region classification (RC), region explanation (RE), and full report generation (RG).
  ChEX shows excellent performance on this wide range of tasks while none of the baselines is capable of even performing all of them. To improve readability, values are scaled relative to the results of ChEX.}
  \label{fig:results}
\end{wrapfigure}

\subsection{Comparison with SOTA}\label{sec:results_sota}
\cref{fig:results} and \cref{tab:restuls_overview} provide an overview of the performance of our model ChEX compared to the best baselines, including specialized SOTA and common multitask models.  
On 8 of the 9 tasks, ChEX is competitive (within 1-std) or better than the best baseline on at least one metric. Only on pathology detection (OD) on VinDR-CXR is it outperformed by a specialized supervised object detection model. 
Note that none of the baselines is capable of performing all the tasks, as most of them are either focused solely on localization or generative tasks, but not both. Only the contrastive image-text (\ie CLIP-style) models can perform a wide range of tasks but rely on retrieval for the generative tasks, thus showing poor performance on these tasks. Overall, our model ChEX shows excellent performance on a wide range of tasks, covering both localization as well as text generation, and is capable of replacing specialized models without major performance drops on most tasks. 

\begin{table}[t]
    \caption{Comparison of ChEX with the best-performing baselines for different types of models. On 8 of the 9 tasks, ChEX is competitive (within 1-std) or better than the best baseline on at least one metric, highlighting that ChEX is capable of replacing specialized models without major performance drops on most tasks. We indicate variability by std computed using bootstrapping and mark the best results as well as those within 1-std in bold. For detailed results, we refer to the supp.\@ material.}
    \label{tab:restuls_overview}
\centering\tiny
\setlength{\tabcolsep}{1.5pt}
\begin{tabular}{llS@{}lS@{}lS@{}lS@{}lS@{}lS@{}l}
\toprule
\scriptsize\textbf{Task} & \scriptsize\textbf{Metric} & \multicolumn{2}{c}{\scriptsize\textbf{ChEX}} & \multicolumn{2}{c}{\scriptsize\textbf{SupOD}} & \multicolumn{2}{c}{\scriptsize\textbf{WSupOD}} & \multicolumn{2}{c}{\scriptsize\textbf{SupVG}} & \multicolumn{2}{c}{\scriptsize\textbf{Contrastive}} & \multicolumn{2}{c}{\scriptsize\textbf{Generative}} \\
\midrule
\multicolumn{5}{l}{\scriptsize\textbf{Sentence Grounding (SG)}} \\
MS-CXR & [mIoU] & 47.52&\std{1.45} & \multicolumn{2}{c}{\text{{--}}} & \multicolumn{2}{c}{\text{{--}}} & \bfseries53.51 & \std{1.53} & 28.57 & \std{1.31} & 46.2 &  \\
& [mAP] & \bfseries 44.47 & \std{2.21} & \multicolumn{2}{c}{\text{{--}}} & \multicolumn{2}{c}{\text{{--}}} & \bfseries44.05 & \std{2.63} & 18.62 & \std{1.37} & \multicolumn{2}{c}{\text{{--}}} \\
\midrule
\multicolumn{5}{l}{\scriptsize\textbf{Pathology Detection (OD)}} \\
VinDrCXR & [mAP] & 14.12 & \std{0.95} & \bfseries18.21 & \std{1.20} & 7.44 & \std{0.88} & \multicolumn{2}{c}{\text{{--}}}  & 2.82 & \std{0.25} & \multicolumn{2}{c}{\text{{--}}}  \\
NIH8 & [mAP] & \bfseries11.14 & \std{1.05} & 6.69 & \std{0.82} & \bfseries11.89 & \std{0.88} & \multicolumn{2}{c}{\text{{--}}} & 2.63 & \std{0.26} & \multicolumn{2}{c}{\text{{--}}} \\
MS-CXR & [mAP] & \bfseries16.60 & \std{1.38} & \bfseries15.83 & \std{1.42} & \bfseries16.56 & \std{1.06} & \multicolumn{2}{c}{\text{{--}}} & 7.15 & \std{0.52} & \multicolumn{2}{c}{\text{{--}}}  \\
\midrule
\multicolumn{5}{l}{\scriptsize\textbf{Region Classification (RC)}} \\
MS-CXR & [AUROC] & \bfseries82.33 & \std{2.80} & 76.13 & \std{2.55} & 61.46 & \std{3.41} & \multicolumn{2}{c}{\text{{--}}} & 67.41 & \std{2.80} & \multicolumn{2}{c}{\text{{--}}} \\
CIG & [wAUROC] & \bfseries70.46 & \std{0.36} & 58.28 & \std{0.22} & 60.02 & \std{0.21} & \multicolumn{2}{c}{\text{{--}}} & 66.96 & \std{0.32} & \multicolumn{2}{c}{\text{{--}}} \\
\midrule
\multicolumn{5}{l}{\scriptsize\textbf{Region Explanation (RE)}} \\
MS-CXR & [Mic-F1-14] & \bfseries49.97 & \std{2.24} & \multicolumn{2}{c}{\text{{--}}}  & \multicolumn{2}{c}{\text{{--}}}  & \multicolumn{2}{c}{\text{{--}}} & 5.86 & \std{1.41} & \bfseries48.97 & \std{2.50} \\ 
 & [Mac-F1-14] & \bfseries20.50 & \std{1.54} & \multicolumn{2}{c}{\text{{--}}} & \multicolumn{2}{c}{\text{{--}}}  & \multicolumn{2}{c}{\text{{--}}} &  3.69 & \std{0.67} & 16.37 & \std{2.00} \\ 
 & [METEOR] & \bfseries8.79 & \std{0.54} & \multicolumn{2}{c}{\text{{--}}} & \multicolumn{2}{c}{\text{{--}}} & \multicolumn{2}{c}{\text{{--}}} & 4.26 & \std{0.36} & 8.15 & \std{0.78} \\ 
CIG & [Mic-F1-14] & \bfseries53.34 & \std{0.43} & \multicolumn{2}{c}{\text{{--}}} & \multicolumn{2}{c}{\text{{--}}} & \multicolumn{2}{c}{\text{{--}}} & 24.40 & \std{0.38} & 45.26 & \std{0.44} \\ 
 & [Mac-F1-14] & \bfseries29.13 & \std{0.35} & \multicolumn{2}{c}{\text{{--}}} & \multicolumn{2}{c}{\text{{--}}} & \multicolumn{2}{c}{\text{{--}}} & 8.93 & \std{0.15} & 20.88 & \std{0.19} \\ 
 & [METEOR] & \bfseries10.18 & \std{0.13} & \multicolumn{2}{c}{\text{{--}}} & \multicolumn{2}{c}{\text{{--}}} & \multicolumn{2}{c}{\text{{--}}} & 3.82 & \std{0.03} & 7.88 & \std{0.10} \\ 
\midrule
\multicolumn{5}{l}{\scriptsize\textbf{Full Report Generation (RG)}} \\
MIMIC-CXR\textsuperscript{\textdagger} & [Mic-F1-14] & 52.32 & \std{0.51} & \multicolumn{2}{c}{\text{{--}}} & \multicolumn{2}{c}{\text{{--}}} & \multicolumn{2}{c}{\text{{--}}} & \multicolumn{2}{c}{\text{{--}}} & \bfseries55.7 \\ 
 & [Mac-F1-14] & 32.56 & \std{0.51} & \multicolumn{2}{c}{\text{{--}}} & \multicolumn{2}{c}{\text{{--}}} & \multicolumn{2}{c}{\text{{--}}} & \multicolumn{2}{c}{\text{{--}}} & \bfseries39.83 \\ 
 & [Ex-F1-14] & \bfseries58.76 & \std{0.42} & \multicolumn{2}{c}{\text{{--}}} & \multicolumn{2}{c}{\text{{--}}} & \multicolumn{2}{c}{\text{{--}}} & \multicolumn{2}{c}{\text{{--}}} & 47.6 \\ 
 & [Mic-F1-5+] & \bfseries61.03 & \std{0.56} & \multicolumn{2}{c}{\text{{--}}} & \multicolumn{2}{c}{\text{{--}}} & \multicolumn{2}{c}{\text{{--}}} & \multicolumn{2}{c}{\text{{--}}} & 58.8\\
 & [Mac-F1-5+] & \bfseries55.85 & \std{0.57} & \multicolumn{2}{c}{\text{{--}}} & \multicolumn{2}{c}{\text{{--}}} & \multicolumn{2}{c}{\text{{--}}} & \multicolumn{2}{c}{\text{{--}}} & 51.7 \\
 & [METEOR] & 13.26 &\std{0.10} & \multicolumn{2}{c}{\text{{--}}} & \multicolumn{2}{c}{\text{{--}}} & \multicolumn{2}{c}{\text{{--}}} & \multicolumn{2}{c}{\text{{--}}} & \bfseries33.3 \\ 
\bottomrule
\end{tabular}\\
\scriptsize
\textsuperscript{\textdagger} Test splits and pre-processing can differ between models, leading to limitations in the exact comparison of results, as also acknowledged by \cite{maira,rgrg}. 

\end{table}

\subsubsection{Localization and Region Classification}
In \emph{sentence grounding (SG)}, ChEX performs similarly to the generative model OmniFM-DR and the SupVG model TransVG (with TransVG showing an advantage in mIoU), despite TransVG being trained explicitly on this task.
In \emph{pathology detection (OD)}, ChEX is competitive on 2 out of 3 tasks. On VinDr-CXR, the SupOD model Faster R-CNN performs best, while on NIH8, ChEX almost doubles the performance of the best SupOD model. On MS-CXR, ChEX is within 1-std of the best SupOD and WSupOD models. Zero-shot contrastive models perform poorly, relying on thresholding of noisy similarity maps.
On \emph{region classification (RC)} tasks, ChEX outperforms all baselines, with improvements of 8\% on MS-CXR and 5\% on CIG. 

\subsubsection{Text Generation}
On \emph{Region explanation (RE)} tasks, ChEX performs similar or better than RGRG. On MS-CXR, ChEX improves by 25\% on Mac-F1-14. On CIG, ChEX improves by 18\% on Mic-F1-14, 40\% on Mac-F1-14, and 29\% on METEOR, although RGRG was explicitly trained on this task. Other report generation models cannot provide region-level descriptions while sentence-retrieval with contrastive baselines performs poorly on these tasks.
\begin{figure}[t]
    \centering
    \begin{subfigure}[t]{0.63\textwidth}
        \includegraphics[trim=0 -20 0 0, clip, width=\textwidth]{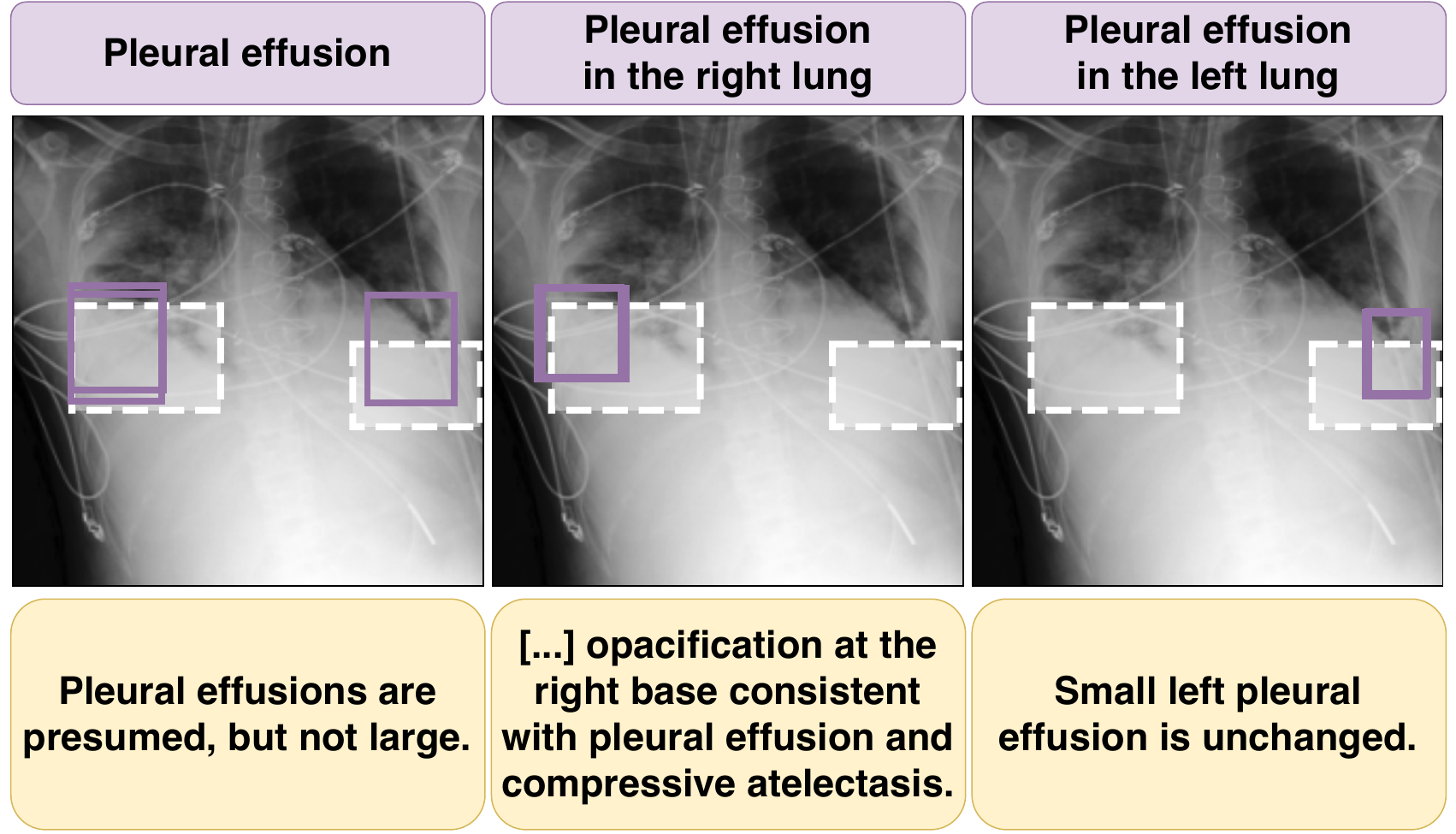} 
        \caption{Example chest X-ray with pleural effusion in both lungs. Ground-truth boxes are marked in white, queries are shown above, their predicted boxes in purple, and their predicted descriptions below.}
    \end{subfigure}
    \begin{subfigure}[t]{0.36\textwidth}
        \includegraphics[width=\textwidth]{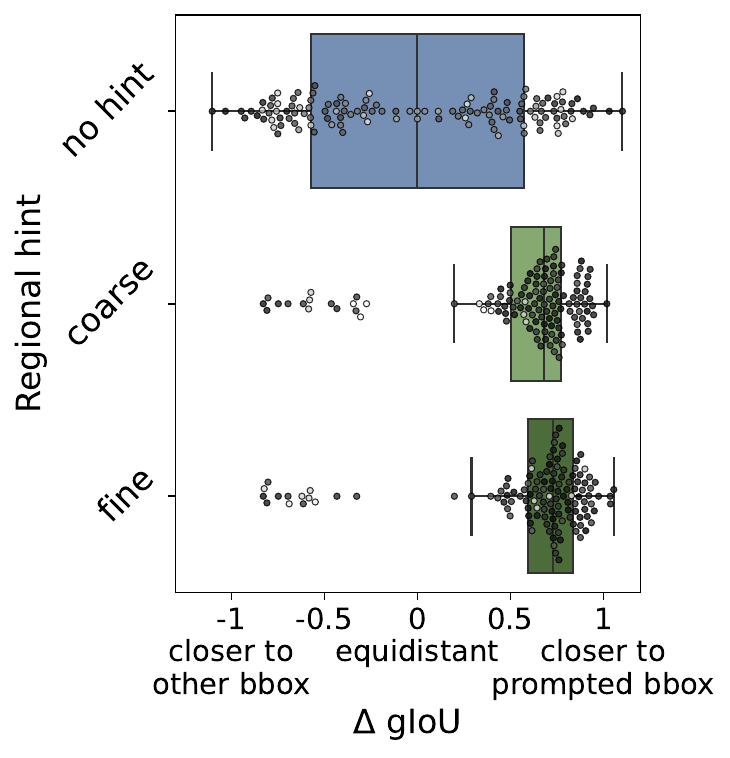} 
        \caption{Relative distance (gIoU) between the queried and the other pathology instance. Darker points mark higher box scores.}
        \label{fig:inter_prompting_quant}
    \end{subfigure}
    \caption{Effect of interactive prompting for multiregion disambiguation. In samples with the presence of the same pathology (\eg, pleural effusion) in both lungs, using no regional hints in the textual query (\protect\say{pleural effusion}) detects both pathology instances equally well while adding a course regional hint (\protect\say{pleural effusion in the right lung}) or a fine regional hint (\protect\say{pleural effusion in the right lower lung}) steers the models towards selecting the queried pathology instance. 
    }
    \label{fig:interactive_prompting}
\end{figure}
For \emph{full report generation (RG)} on MIMIC-CXR, ChEX sets a new state-of-the-art on the metrics Ex-F1-14 ($+23\%$), Mic-F1-5+ ($+4\%$), and Mac-F1-5+ ($+8\%$). On the commonly used Mic-F1-14 metric, ChEX outperforms the 10 times larger model Med-PaLM M 12B and is only slightly outperformed by the 7 times larger SOTA model MAIRA-1. 
While ChEX shows limitations on Mac-F1-14, 
it still outperforms RGRG. However, ChEX performs relatively low on language-based metrics like METEOR due to its query-based report generation approach. Overall, ChEX demonstrates strong report generation performance, achieving results that are close or even better than SOTA models of up to 80 times the size of ChEX.

\subsection{Interactive Prompting}\label{sec:inter_prompting}
\begin{figure}[t]
    \centering
    \begin{subfigure}[t]{0.63\textwidth}
        \includegraphics[trim=0 -20 0 0, clip, width=\textwidth]{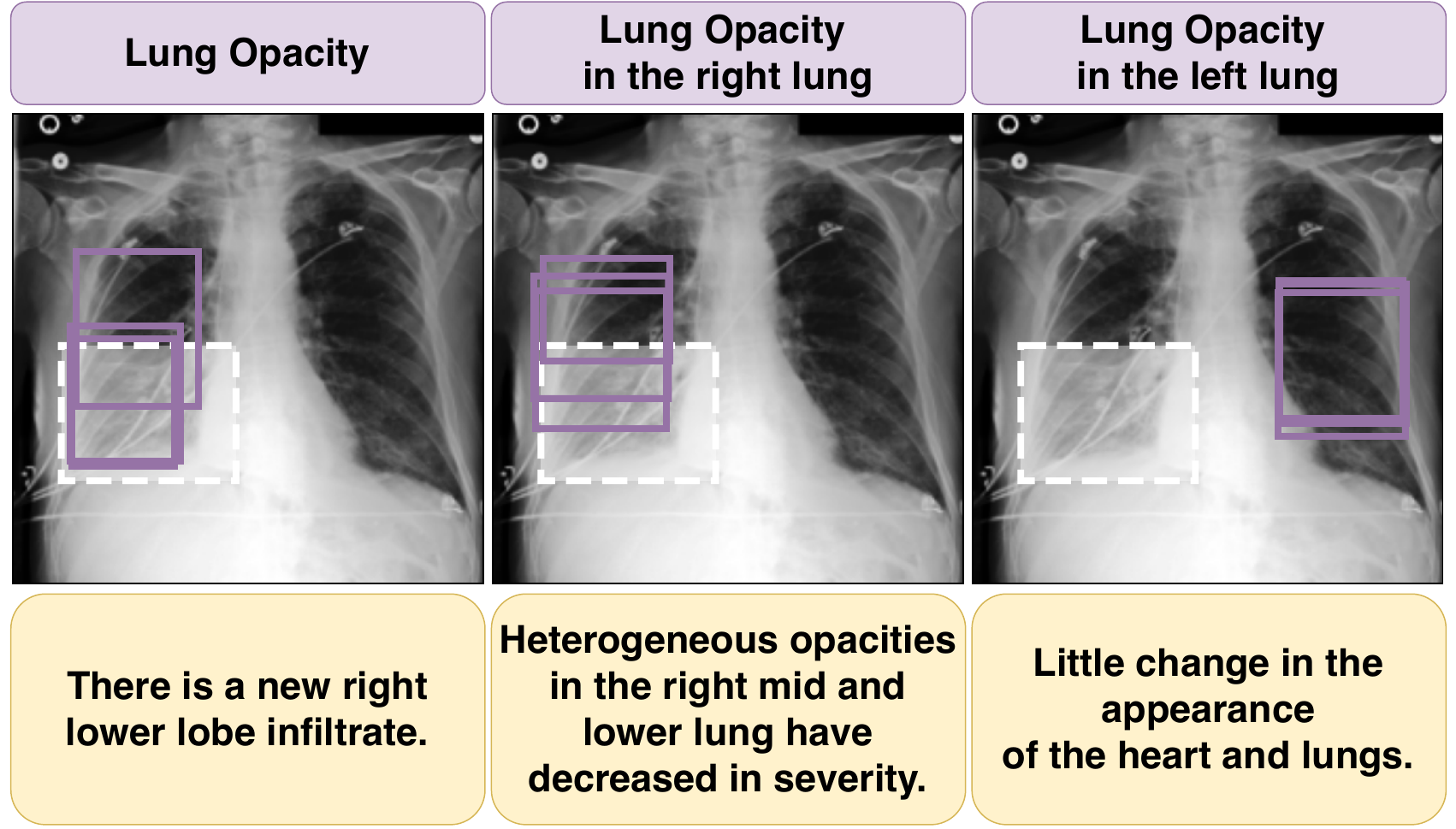}
        \caption{Example chest X-ray with a lung opacity only in the right lung. Ground-truth boxes are marked in white, queries are shown above, their predicted boxes are in purple, and their predicted descriptions are shown below.}
    \end{subfigure}
    \begin{subfigure}[t]{0.35\textwidth}
         \includegraphics[width=\textwidth]{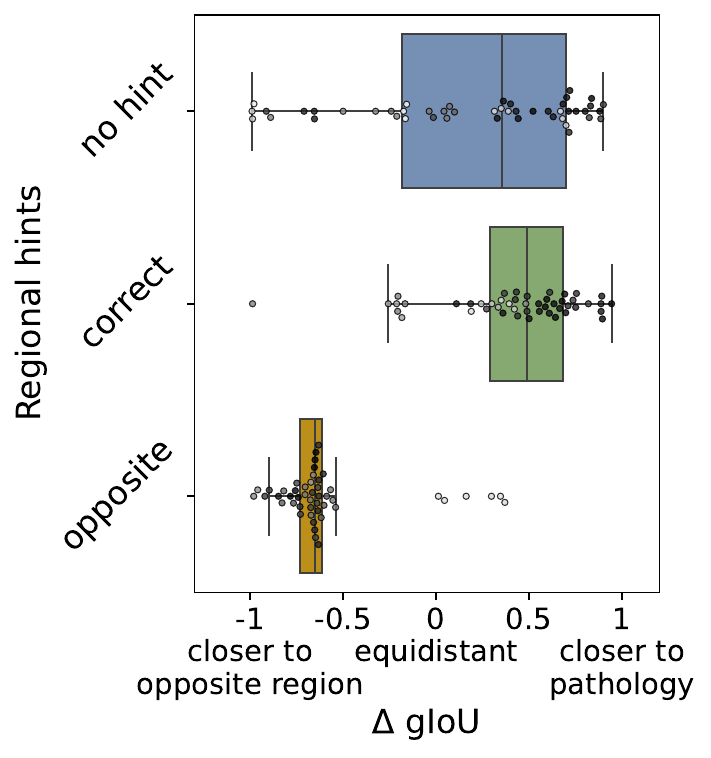}
         \caption{Relative distance (gIoU) between the pathology and the opposite lung region. Darker dots mark higher box scores.}
         \label{fig:inter_prompting_neg_quant}
    \end{subfigure}
    \caption{Effect of interactive prompting with regional hints to negative regions.
    In samples with a pathology in only one of the lungs (\eg, lung opacity in the right lung), 
    using no regional hint (\protect\say{lung opacity}) detects the pathology mostly correctly. Adding the correct regional hint (\protect\say{lung opacity in the right lung}) improves the localization while the regional hint for the opposite lung (\protect\say{lung opacity in the left lung}) steers the model towards the queried anatomical region (away from the pathology), 
    as expected. 
    }
    \label{fig:interactive_prompting_negative}
\end{figure}

ChEX aims to enable the interactive diagnosis of chest X-rays, going beyond the benchmarked tasks in \cref{sec:results_sota}. However, ChEX was trained only on simple prompts (\eg, the name of a pathology). We thus validate that ChEX generalizes to more complex prompts  by investigating the predicted bounding boxes in two scenarios:
(i) multi-region disambiguation via regional hints and (ii) regional hints for negative regions. 
For scenario (i), shown in \cref{fig:interactive_prompting}, we consider cases where a specific pathology is present in both lungs, \ie where there are two boxes for the same pathology.
We can observe that when using no regional hints in the textual query (\say{pleural effusion}), ChEX detects both pathology instances equally well while adding a course regional hint (\say{pleural effusion in the right lung}) or a fine regional hint (\say{pleural effusion in the right lower lung}), steers the models towards selecting the queried pathology instance. 
For scenario (ii), shown in \cref{fig:interactive_prompting_negative}, we consider samples where the queried pathology is only present once (\ie in one of the lungs). We again study the effect of using regional hints, either the correct hint (\protect\say{lung opacity in the right lung}, assuming there is a lung opacity only in the right lung) or the hint to the opposite lung (\protect\say{lung opacity in the left lung}). 
We found that using no regional hint (\protect\say{lung opacity}) detects the pathology mostly correctly, while adding the correct regional hint improves the localization. If, on the other hand, we provide the regional hint for the opposite lung, the model is steered towards checking the queried anatomical
region and thus away from the pathology. Overall, ChEX considers the user's intents of textual prompts very well and thus enables interactive usage patterns.

\subsection{Improvement through Precise Interactive Prompting}\label{sec:precise_prompts}
\begin{figure}[t]
    \centering
    \begin{subfigure}[t]{0.46\textwidth}
        \includegraphics[trim=0 -25 0 0, clip, width=\textwidth]{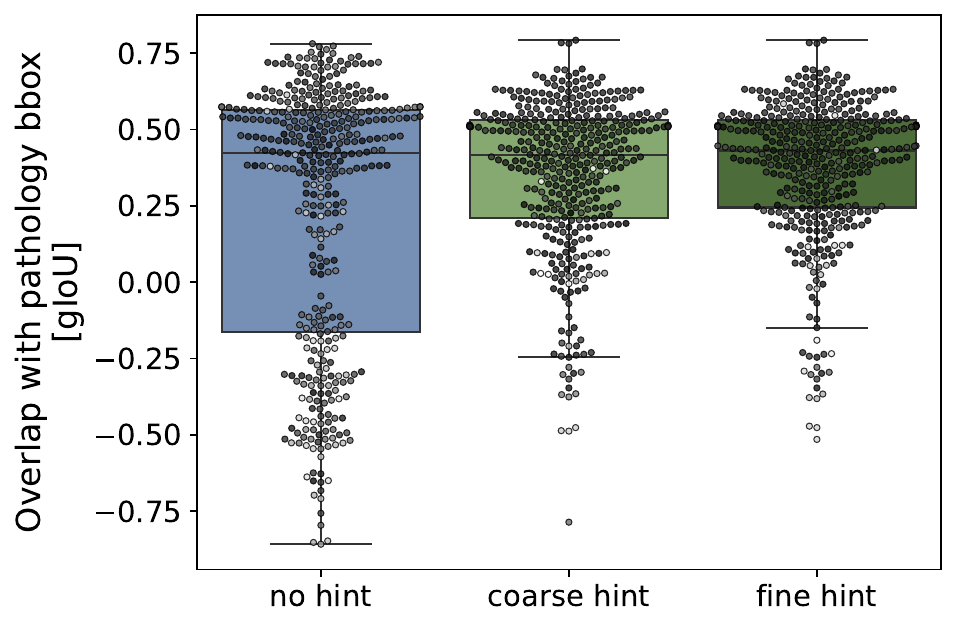}
    \caption{Effect of prompting strategies on pathology localization quality (gIoU). Compared to using no regional hint (\eg \protect\say{pneumonia}), a course hint (\protect\say{pneumonia in the left lung}) or a fine hint (\protect\say{pneumonia in the left upper lung}) improves pathology localization. Darker dots mark higher box scores.}
    \label{fig:precise_prompts_iou}
    \end{subfigure}
    \begin{subfigure}[t]{0.53\textwidth}
        \includegraphics[width=\textwidth]{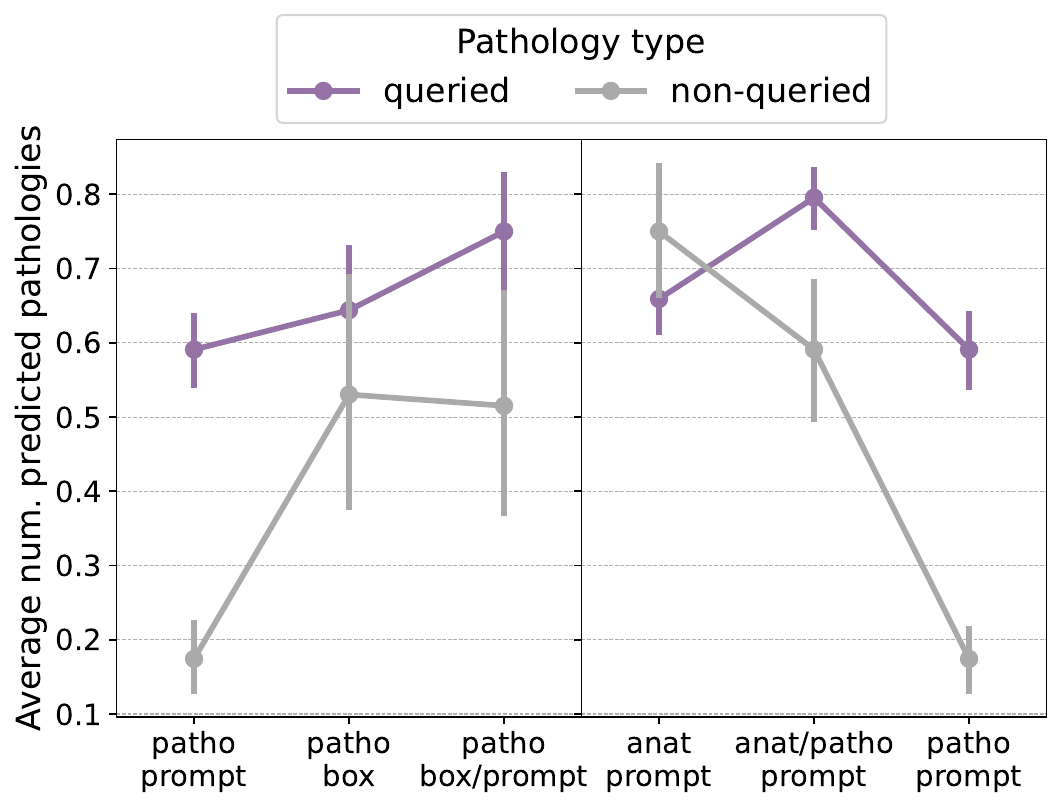}
        \caption{Effect of prompting strategies on the presence of queried pathologies (\ie defined in the prompt) and non-queried (\ie any other positive or negative) pathologies in the predicted sentence. \textbf{Left:} Pathologies queried by its textual prompt (patho prompt), by its bounding box (patho box), or both. \textbf{Right:} Prompting the associated anatomical region of the pathology (anat prompt, \eg \protect\say{left lung}), the pathology with a regional hint (\protect\say{penumonia in the left lung}), or only the pathology (\protect\say{penumonia}).}
        \label{fig:precise_prompts_text}
    \end{subfigure}
    \caption{Effect of precise prompting on localization (a) and sentence prediction (b). Preciser prompts improve localization and the description of queried pathologies, while more specific prompts reduce the description of additional, non-queried pathologies.}
    \label{fig:precise_prompt}
\end{figure}

ChEX facilitates the interactive involvement of medical expert users. In \cref{fig:precise_prompt}, 
we investigate the impact of 
providing preciser prompts
on the model's localization quality, its ability to accurately describe the queried pathology in the predicted sentence (prediction accuracy), and the inclusion of non-queried pathologies (query specificity). 
Providing coarse regional hints (\eg, \say{pneumonia in the left lung}) enhances localization and prediction accuracy compared to only providing the pathology name (\say{pneumonia}). Further refinement of regional hints (\eg, \say{[...] in the left \emph{upper} lung}) 
offers minimal additional benefit on localization. In addition to textual prompts, the model can be queried using bounding boxes of relevant regions. Utilizing the bounding box of a pathology (instead of a textual prompt) improves the prediction accuracy but also includes additional aspects, rendering the prediction less query-specific. Combining textual and bounding box queries yields the best prediction accuracy, enhancing query specificity compared to using only bounding boxes. When employing a textual query of the associated anatomical region (\eg, \say{left lung}), the present pathology is described adequately, but -- as expected -- descriptions are not specific to a single pathology. 
Using a more precise prompt (\eg, \say{pneumonia in the left lung}) enhances prediction accuracy and query specificity while relying solely on the pathology prompt increases query specificity further but marginally degrades the prediction accuracy. Despite ChEX demonstrating competitive performance even with simple prompts (\cref{sec:results_sota}), its responsiveness to more specific prompts enables even greater prediction accuracy when used interactively.

\subsection{Interpretable and Customizable Report Generation}\label{sec:custom_reportgen}
\begin{figure}[t]
    \centering
    \includegraphics[width=.9\textwidth]{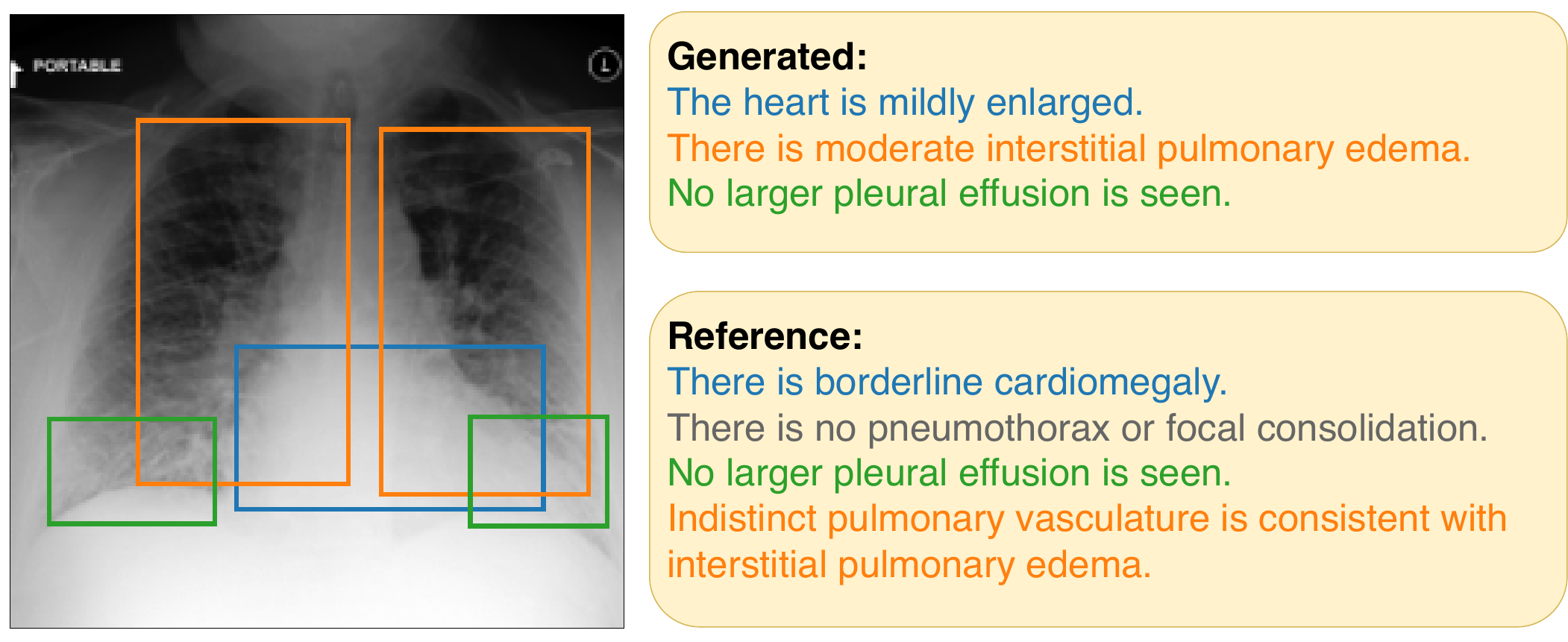}
    \caption{Example of a generated report with predicted bounding boxes. Our model generates a concise and accurate report. By predicting bounding boxes for descriptions, ChEX provides a high level of interpretability and promotes easy checking of the generated report through radiologists, enabling an optimal workflow for clinical practice. For further qualitative examples, we refer to the supp.\@ material.}
    \label{fig:qual_report}
\end{figure}
ChEX supports automatic full report generation using a pre-defined set of textual prompts based on which the model localizes and describes relevant regions.
Unlike typical report generation models, the reliance on prompt sets for report generation offers high flexibility as the prompt sets can be customized without the need for re-training.
We study (cf.\@ supp.\@ material) the effect of using only pathologies, anatomical regions, or both as prompt sets. The choice of the prompt set enables balancing precision and recall of the model. While all studied prompt sets lead to results competitive with the baselines, with Mic-F1-14 ranging from $50.08$ to $52.37$, using both prompt sets leads to optimal performance. 

In \cref{fig:qual_report}, we show an example of a generated report. 
The prediction of bounding boxes for predicted sentences enables a high degree of transparency and interpretability -- not provided by most report generation models -- thus simplifying correctness checking and enabling an optimal clinical workflow.

\subsection{Technical Insights}\label{sec:technical_insights}
We additionally conducted ablation studies on several technical design decisions of our model and training paradigm. We summarize the key findings in the following. For detailed results we refer to the supp.\@ material.

\subsubsection{The mixture of tokens for pathologies, anatomy, and report sentences enables multitask capabilities.}
We found that using all these three token types during training achieves the best multitask performance. Having both pathology and anatomy tokens is especially relevant for accurate localization (SG and OD tasks). Anatomy tokens are especially relevant for RC and RE tasks, likely due to the availability of pathology labels and sentences associated with bounding boxes.
While sentence tokens can slightly harm localization quality, they are relevant to achieve the best possible text generation performance.

\subsubsection{Localization targets are highly beneficial for all tasks, even for full report generation.}
We identified the importance of bounding box targets for all tasks. Pathology class labels are relevant mainly for OD tasks. 
Contrastive sentence supervision improves some OD tasks as well as MS-CXR-based RC and RE tasks, \ie it helps with the understanding of pathology regions, while generative text supervision is not significantly beneficial for non-generative tasks.

\subsubsection{Technical differences to baselines}
A key distinction between ChEX and most generative baselines lies in its aspect-level generation approach, where sentences are generated individually for specific findings and regions of the image. In contrast, models like MAIRA-1\cite{maira}, Med-PaLM M \cite{medpalmm}, or OmniFM-DR\cite{omnifm} generate the full report in a single shot. This approach is critical to ChEX's strong performance, as demonstrated by the good results of RGRG\cite{rgrg}, which employs a similar region-level strategy. Compared to RGRG, ChEX introduces two key innovations: (i) pathology and sentence tokens, which extend RGRG's solely anatomy-based approach; and (ii) contrastive alignment of region features with their sentences and pathology prompts. Our ablation studies confirm that these differences are essential to ChEX's superior performance over RGRG.

\section{Discussion}
\subsubsection{Importance of Semantically Meaningful Visual Features}
Many recent works on radiology report generation heavily focus on the utilization of large language models (LLMs) \cite{maira,medpalmm,omnifm,radialog}.
We demonstrate that even smaller language models can achieve competitive performance when prioritizing improvements in image understanding aspects.
Specifically, our approach places emphasis on semantic regions, leveraging bounding box supervision to enhance prediction quality. 
We argue that a synergistic combination of optimal image encoding with regional focus
along with the deductive powers and knowledge of LLMs presents a promising path for improving report generation. 
Furthermore, our proposed training strategy enables the integration of multiple tasks and datasets, offering a viable approach to realizing this path.

\subsubsection{Interpretability and Interactivity for Clinical Practice}
While the performance of report generation models is improving,
even occasional inaccuracies in their predictions can limit their clinical applicability.
We argue that a high degree of interpretability and the integration of medical experts in the generation process, \ie interactivity, offers a more promising path to clinical application than solely improving prediction quality. Our model ChEX showcases a unique combination of interpretability and interactivity -- not provided by other models -- by providing bounding boxes for generated descriptions while enabling user guidance through textual prompts and bounding box queries.

\subsubsection{Limitations}
Our training approach effectively utilizes the available datasets and eliminates the need for intricate data engineering. However, this comes at the cost of a more complex training process.
Furthermore, the used datasets do not provide question-answer pairs, so textual queries are either based on pre-defined prompts 
or report sentences, and answers are always based on report sentences. This limits the types of supported textual queries, mainly regional or pathology hints, or both. At the same time, answers are plain descriptions of the queried regions or pathologies, while specific answers to more complicated questions (\eg comparing regions) are not supported. Also, using report sentences as answers can lead to the hallucination of comparisons with previous images, although only a single image is used, a phenomenon common to report generation models \cite{maira,biovil_t,ramesh2022improving}.
Future work may use instruction tuning to tackle these issues.

Additionally, as we move towards clinical application, further evaluation from a radiologist's perspective is essential, and future work should include systematic studies on user experience to ensure seamless integration into clinical workflows.

\subsubsection{Conclusion}
We proposed ChEX, a model for predicting visually grounded textual descriptions of chest X-rays based on user queries. 
Our analysis underscores ChEX's competitive performance against SOTA models across 9 tasks, and its responsiveness to user prompts, therefore laying a foundation for future advancements in interactive and localized text generation models.

\section*{Acknowledgements}
The project was supported by ERC Grant Deep4MI (884622).

GK and DR received support from the German Federal Ministry of Education and Research and the Bavarian State Ministry for Science and the Arts under the Munich Centre for Machine Learning (MCML), from the German Ministry of Education and Research and the the Medical Informatics Initiative as part of the PrivateAIM Project, from the Bavarian Collaborative Research Project PRIPREKI of the Free State of Bavaria Funding Programme \say{Artificial Intelligence -- Data Science}, and from the German Academic Exchange Service (DAAD) under the Kondrad Zuse School of Excellence for Reliable AI (RelAI).

%
%
\bibliographystyle{splncs04}
\bibliography{ms}

\begin{thebibliography}{10}
\providecommand{\url}[1]{\texttt{#1}}
\providecommand{\urlprefix}{URL }
\providecommand{\doi}[1]{https://doi.org/#1}

\bibitem{meteor}
Banerjee, S., Lavie, A.: Meteor: An automatic metric for mt evaluation with improved correlation with human judgments. In: ACL workshop on intrinsic and extrinsic evaluation measures for machine translation and/or summarization. pp. 65--72 (2005)

\bibitem{biovil_t}
Bannur, S., Hyland, S., Liu, Q., Pérez-García, F., Ilse, M., Castro, D.C., Boecking, B., Sharma, H., Bouzid, K., Thieme, A., Schwaighofer, A., Wetscherek, M., Lungren, M.P., Nori, A., Alvarez-Valle, J., Oktay, O.: Learning to exploit temporal structure for biomedical vision-language processing. In: CVPR. pp. 15016--15027 (2023). \doi{10.1109/CVPR52729.2023.01442}

\bibitem{mscxr2}
Boecking, B., Usuyama, N., Bannur, S., Coelho~de Castro, D., Schwaighofer, A., Hyland, S., Wetscherek, M.T., Naumann, T., Nori, A., Alvarez~Valle, J., Poon, H., Oktay, O.: Ms-cxr: Making the most of text semantics to improve biomedical vision-language processing (version 0.1). PhysioNet  (2021). \doi{https://doi.org/10.13026/b90j-vb87}

\bibitem{biovil}
Boecking, B., Usuyama, N., Bannur, S., Castro, D.C., Schwaighofer, A., Hyland, S., Wetscherek, M., Naumann, T., Nori, A., Alvarez-Valle, J., Poon, H., Oktay, O.: Making the most of text semantics to improve biomedical vision--language processing. In: Avidan, S., Brostow, G., Ciss{\'e}, M., Farinella, G.M., Hassner, T. (eds.) ECCV. pp. 1--21. Springer Nature Switzerland, Cham (2022). \doi{10.1007/978-3-031-20059-5_1}

\bibitem{albumentations}
Buslaev, A., Iglovikov, V.I., Khvedchenya, E., Parinov, A., Druzhinin, M., Kalinin, A.A.: Albumentations: Fast and flexible image augmentations. Information  \textbf{11}(2) (2020). \doi{10.3390/info11020125}, \url{https://www.mdpi.com/2078-2489/11/2/125}

\bibitem{detr}
Carion, N., Massa, F., Synnaeve, G., Usunier, N., Kirillov, A., Zagoruyko, S.: End-to-end object detection with transformers. In: Vedaldi, A., Bischof, H., Brox, T., Frahm, J.M. (eds.) ECCV. pp. 213--229. Springer International Publishing, Cham (2020). \doi{10.1007/978-3-030-58452-8_13}

\bibitem{chen2023_miccai}
Chen, Z., Zhou, Y., Tran, A., Zhao, J., Wan, L., Ooi, G.S.K., Cheng, L.T.E., Thng, C.H., Xu, X., Liu, Y., Fu, H.: Medical phrase grounding with region-phrase context contrastive alignment. In: Greenspan, H., Madabhushi, A., Mousavi, P., Salcudean, S., Duncan, J., Syeda-Mahmood, T., Taylor, R. (eds.) MICCAI. pp. 371--381. Springer Nature Switzerland, Cham (2023). \doi{10.1007/978-3-031-43990-2_35}

\bibitem{transvgpp}
Deng, J., Yang, Z., Liu, D., Chen, T., Zhou, W., Zhang, Y., Li, H., Ouyang, W.: Transvg++: End-to-end visual grounding with language conditioned vision transformer. IEEE TPAMI  \textbf{45}(11),  13636--13652 (nov 2023). \doi{10.1109/TPAMI.2023.3296823}

\bibitem{transvg}
Deng, J., Yang, Z., Chen, T., Zhou, W., Li, H.: Transvg: End-to-end visual grounding with transformers. In: ICCV. pp. 1749--1759. {IEEE} (2021). \doi{10.1109/ICCV48922.2021.00179}, \url{https://doi.org/10.1109/ICCV48922.2021.00179}

\bibitem{vit}
Dosovitskiy, A., Beyer, L., Kolesnikov, A., Weissenborn, D., Zhai, X., Unterthiner, T., Dehghani, M., Minderer, M., Heigold, G., Gelly, S., Uszkoreit, J., Houlsby, N.: An image is worth 16x16 words: Transformers for image recognition at scale. ICLR  (2021)

\bibitem{du2022visual}
Du, Y., Fu, Z., Liu, Q., Wang, Y.: Visual grounding with transformers. In: ICME (2022)

\bibitem{eslami_2021}
Eslami, S., de~Melo, G., Meinel, C.: Does {CLIP} benefit visual question answering in the medical domain as much as it does in the general domain? ArXiv preprint  \textbf{abs/2112.13906} (2021), \url{https://arxiv.org/abs/2112.13906}

\bibitem{geis2019ethics}
Geis, J.R., Brady, A.P., Wu, C.C., Spencer, J., Ranschaert, E., Jaremko, J.L., Langer, S.G., Borondy~Kitts, A., Birch, J., Shields, W.F., et~al.: Ethics of artificial intelligence in radiology: summary of the joint european and north american multisociety statement. Radiology  \textbf{293}(2),  436--440 (2019)

\bibitem{physionet}
Goldberger, A., Amaral, L., Glass, L., Hausdorff, J., et~al.: Physiobank, physiotoolkit, and physionet: Components of a new research resource for complex physiologic signals. Circulation [Online]  \textbf{101}(23),  215--–220 (2000)

\bibitem{Gu_2024_comg}
Gu, T., Liu, D., Li, Z., Cai, W.: Complex organ mask guided radiology report generation. In: WACV. pp. 7995--8004 (January 2024)

\bibitem{vild}
Gu, X., Lin, T., Kuo, W., Cui, Y.: Open-vocabulary object detection via vision and language knowledge distillation. In: ICLR (2022), \url{https://openreview.net/forum?id=lL3lnMbR4WU}

\bibitem{guo_2023_miccai}
Guo, M., Yi, H., Qin, Z., Wang, H., Men, A., Lao, Q.: Multiple prompt fusion for zero-shot lesion detection using vision-language models. In: Greenspan, H., Madabhushi, A., Mousavi, P., Salcudean, S., Duncan, J., Syeda-Mahmood, T., Taylor, R. (eds.) MICCAI. pp. 283--292. Springer Nature Switzerland, Cham (2023). \doi{10.1007/978-3-031-43904-9_28}

\bibitem{he2024pefomed}
He, J., Li, P., Liu, G., Zhao, Z., Zhong, S.: Pefomed: Parameter efficient fine-tuning on multimodal large language models for medical visual question answering (2024)

\bibitem{resnet}
He, K., Zhang, X., Ren, S., Sun, J.: Deep residual learning for image recognition. In: CVPR. pp. 770--778 (2016)

\bibitem{gelu}
Hendrycks, D., Gimpel, K.: Gaussian error linear units (gelus) (2023)

\bibitem{hou_2023_organ}
Hou, W., Xu, K., Cheng, Y., Li, W., Liu, J.: {ORGAN}: Observation-guided radiology report generation via tree reasoning. In: Rogers, A., Boyd-Graber, J., Okazaki, N. (eds.) Proceedings of the 61st Annual Meeting of the Association for Computational Linguistics (Volume 1: Long Papers). pp. 8108--8122. Association for Computational Linguistics, Toronto, Canada (Jul 2023). \doi{10.18653/v1/2023.acl-long.451}, \url{https://aclanthology.org/2023.acl-long.451}

\bibitem{densenet}
Huang, G., Liu, Z., Van Der~Maaten, L., Weinberger, K.Q.: Densely connected convolutional networks. In: CVPR. pp. 2261--2269 (2017). \doi{10.1109/CVPR.2017.243}

\bibitem{droppath}
Huang, G., Sun, Y., Liu, Z., Sedra, D., Weinberger, K.Q.: Deep networks with stochastic depth. In: Leibe, B., Matas, J., Sebe, N., Welling, M. (eds.) ECCV. pp. 646--661. Springer International Publishing, Cham (2016). \doi{10.1007/978-3-319-46493-0_39}

\bibitem{gloria}
Huang, S., Shen, L., Lungren, M.P., Yeung, S.: Gloria: {A} multimodal global-local representation learning framework for label-efficient medical image recognition. In: ICCV. pp. 3922--3931. {IEEE} (2021). \doi{10.1109/ICCV48922.2021.00391}, \url{https://doi.org/10.1109/ICCV48922.2021.00391}

\bibitem{huang2024_mia}
Huang, Y., Yang, X., Liu, L., Zhou, H., Chang, A., Zhou, X., Chen, R., Yu, J., Chen, J., Chen, C., Liu, S., Chi, H., Hu, X., Yue, K., Li, L., Grau, V., Fan, D.P., Dong, F., Ni, D.: Segment anything model for medical images? Medical Image Analysis  \textbf{92},  103061 (2024). \doi{https://doi.org/10.1016/j.media.2023.103061}, \url{https://www.sciencedirect.com/science/article/pii/S1361841523003213}

\bibitem{maira}
Hyland, S.L., Bannur, S., Bouzid, K., Castro, D.C., Ranjit, M., Schwaighofer, A., Pérez-García, F., Salvatelli, V., Srivastav, S., Thieme, A., Codella, N., Lungren, M.P., Wetscherek, M.T., Oktay, O., Alvarez-Valle, J.: Maira-1: A specialised large multimodal model for radiology report generation (2023)

\bibitem{ichinose2023_miccai}
Ichinose, A., Hatsutani, T., Nakamura, K., Kitamura, Y., Iizuka, S., Simo-Serra, E., Kido, S., Tomiyama, N.: Visual grounding of whole radiology reports for 3d ct images. In: Greenspan, H., Madabhushi, A., Mousavi, P., Salcudean, S., Duncan, J., Syeda-Mahmood, T., Taylor, R. (eds.) MICCAI. pp. 611--621. Springer Nature Switzerland, Cham (2023). \doi{10.1007/978-3-031-43904-9_59}

\bibitem{chexpert}
Irvin, J., Rajpurkar, P., Ko, M., Yu, Y., Ciurea{-}Ilcus, S., Chute, C., Marklund, H., Haghgoo, B., Ball, R.L., Shpanskaya, K.S., Seekins, J., Mong, D.A., Halabi, S.S., Sandberg, J.K., Jones, R., Larson, D.B., Langlotz, C.P., Patel, B.N., Lungren, M.P., Ng, A.Y.: Chexpert: A large chest radiograph dataset with uncertainty labels and expert comparison. In: AAAI. pp. 590--597 (2019). \doi{10.1609/aaai.v33i01.3301590}

\bibitem{promptmrg}
Jin, H., Che, H., Lin, Y., Chen, H.: Promptmrg: Diagnosis-driven prompts for medical report generation (2024)

\bibitem{mimic-cxr}
Johnson, A., Pollard, T., Berkowitz, S., et~al.: Mimic-cxr, a de-identified publicly available database of chest radiographs with free-text reports. Sci Data  \textbf{6}(317) (2019). \doi{https://doi.org/10.1038/s41597-019-0322-0}

\bibitem{mimic-cxr-2}
Johnson, A., Pollard, T., Mark, R., Berkowitz, S., Horng, S.: Mimic-cxr database (version 2.0.0). PhysioNet (2019). \doi{https://doi.org/10.13026/C2JT1Q}

\bibitem{johnson2019mimicjpg}
Johnson, A.E.W., Pollard, T.J., Berkowitz, S.J., Greenbaum, N.R., Lungren, M.P., Deng, C.y., Mark, R.G., Horng, S.: Mimic-cxr-jpg, a large publicly available database of labeled chest radiographs. arXiv preprint arXiv:1901.07042  (2019)

\bibitem{sam}
Kirillov, A., Mintun, E., Ravi, N., Mao, H., Rolland, C., Gustafson, L., Xiao, T., Whitehead, S., Berg, A.C., Lo, W.Y., Dollár, P., Girshick, R.: Segment anything (2023)

\bibitem{llavamed}
Li, C., Wong, C., Zhang, S., Usuyama, N., Liu, H., Yang, J., Naumann, T., Poon, H., Gao, J.: Llava-med: Training a large language-and-vision assistant for biomedicine in one day (2023)

\bibitem{glip}
Li, L.H., Zhang, P., Zhang, H., Yang, J., Li, C., Zhong, Y., Wang, L., Yuan, L., Zhang, L., Hwang, J.N., Chang, K.W., Gao, J.: Grounded language-image pre-training. In: CVPR. pp. 10955--10965 (2022). \doi{10.1109/CVPR52688.2022.01069}

\bibitem{li_2023_cvpr}
Li, M., Lin, B., Chen, Z., Lin, H., Liang, X., Chang, X.: Dynamic graph enhanced contrastive learning for chest x-ray report generation. In: CVPR. pp. 3334--3343 (2023). \doi{10.1109/CVPR52729.2023.00325}

\bibitem{li2021referring}
Li, M., Sigal, L.: Referring transformer: {A} one-step approach to multi-task visual grounding. In: Ranzato, M., Beygelzimer, A., Dauphin, Y.N., Liang, P., Vaughan, J.W. (eds.) NeurIPS. pp. 19652--19664 (2021), \url{https://proceedings.neurips.cc/paper/2021/hash/a376802c0811f1b9088828288eb0d3f0-Abstract.html}

\bibitem{liao_2021}
Liao, R., Moyer, D., Cha, M., Quigley, K., Berkowitz, S., Horng, S., Golland, P., Wells, W.M.: Multimodal representation learning via maximization of local mutual information. In: de~Bruijne, M., Cattin, P.C., Cotin, S., Padoy, N., Speidel, S., Zheng, Y., Essert, C. (eds.) MICCAI. pp. 273--283. Springer International Publishing, Cham (2021). \doi{10.1007/978-3-030-87196-3_26}

\bibitem{focalloss}
Lin, T.Y., Goyal, P., Girshick, R., He, K., Dollár, P.: Focal loss for dense object detection. IEEE TPAMI  \textbf{42}(2),  318--327 (2020). \doi{10.1109/TPAMI.2018.2858826}

\bibitem{liu_2023_vqa}
Liu, J., Hu, T., Zhang, Y., Feng, Y., Hao, J., Lv, J., Liu, Z.: Parameter-efficient transfer learning for medical visual question answering. IEEE Transactions on Emerging Topics in Computational Intelligence pp. 1--11 (2023). \doi{10.1109/TETCI.2023.3311333}

\bibitem{liu2022dabdetr}
Liu, S., Li, F., Zhang, H., Yang, X., Qi, X., Su, H., Zhu, J., Zhang, L.: {DAB-DETR:} dynamic anchor boxes are better queries for {DETR}. In: ICLR (2022), \url{https://openreview.net/forum?id=oMI9PjOb9Jl}

\bibitem{liu2023grounding}
Liu, S., Zeng, Z., Ren, T., Li, F., Zhang, H., Yang, J., Li, C., Yang, J., Su, H., Zhu, J., Zhang, L.: Grounding dino: Marrying dino with grounded pre-training for open-set object detection (2023)

\bibitem{p-tuning}
Liu, X., Ji, K., Fu, Y., Du, Z., Yang, Z., Tang, J.: P-tuning v2: Prompt tuning can be comparable to fine-tuning universally across scales and tasks. CoRR  \textbf{abs/2110.07602} (2021), \url{https://arxiv.org/abs/2110.07602}

\bibitem{adamw}
Loshchilov, I., Hutter, F.: Decoupled weight decay regularization. In: ICLR (2019)

\bibitem{ma2024segment}
Ma, J., He, Y., Li, F., Han, L., You, C., Wang, B.: Segment anything in medical images. Nature Communications  \textbf{15}(1), ~654 (2024)

\bibitem{Maaz2022Multimodal}
Maaz, M., Rasheed, H., Khan, S., Khan, F.S., Anwer, R.M., Yang, M.H.: Class-agnostic object detection with multi-modal transformer. In: ECCV. Springer (2022)

\bibitem{conditional-detr}
Meng, D., Chen, X., Fan, Z., Zeng, G., Li, H., Yuan, Y., Sun, L., Wang, J.: Conditional detr for fast training convergence. In: ICCV. pp. 3631--3640 (2021). \doi{10.1109/ICCV48922.2021.00363}

\bibitem{miller2019explanation}
Miller, T.: Explanation in artificial intelligence: Insights from the social sciences. Artificial intelligence  \textbf{267},  1--38 (2019)

\bibitem{miura2021improving}
Miura, Y., Zhang, Y., Tsai, E., Langlotz, C., Jurafsky, D.: Improving factual completeness and consistency of image-to-text radiology report generation. In: NAACL. pp. 5288--5304 (2021)

\bibitem{lovt}
M{\"u}ller, P., Kaissis, G., Zou, C., Rueckert, D.: Joint learning of localized representations from medical images and reports. In: Avidan, S., Brostow, G., Ciss{\'e}, M., Farinella, G.M., Hassner, T. (eds.) ECCV. pp. 685--701. Springer Nature Switzerland, Cham (2022)

\bibitem{adpd}
M{\"u}ller, P., Meissen, F., Brandt, J., Kaissis, G., Rueckert, D.: Anatomy-driven pathology detection on chest x-rays. In: Greenspan, H., Madabhushi, A., Mousavi, P., Salcudean, S., Duncan, J., Syeda-Mahmood, T., Taylor, R. (eds.) MICCAI. pp. 57--66. Springer Nature Switzerland, Cham (2023). \doi{10.1007/978-3-031-43907-0_6}

\bibitem{wsrpn}
Müller, P., Meissen, F., Kaissis, G., Rueckert, D.: Weakly supervised object detection in chest x-rays with differentiable roi proposal networks and soft roi pooling (2024)

\bibitem{vindr2}
Nguyen, H.Q., Pham, H.H., Tuan~Linh, L., Dao, M., Khanh, L.: Vindr-cxr: An open dataset of chest x-rays with radiologist annotations (version 1.0.0). PhysioNet  (2021). \doi{https://doi.org/10.13026/3akn-b287}

\bibitem{vindr}
Nguyen, H.Q., Lam, K., Le, L.T., Pham, H.H., Tran, D.Q., Nguyen, D.B., Le, D.D., Pham, C.M., Tong, H.T., Dinh, D.H., et~al.: Vindr-cxr: An open dataset of chest x-rays with radiologist’s annotations. Scientific Data  \textbf{9}(1), ~429 (2022). \doi{https://doi.org/10.1038/s41597-022-01498-w}

\bibitem{nicolson_2023}
Nicolson, A., Dowling, J., Koopman, B.: Improving chest x-ray report generation by leveraging warm starting. Artificial Intelligence in Medicine  \textbf{144},  102633 (2023). \doi{https://doi.org/10.1016/j.artmed.2023.102633}, \url{https://www.sciencedirect.com/science/article/pii/S0933365723001471}

\bibitem{CPC}
van~den Oord, A., Li, Y., Vinyals, O.: Representation learning with contrastive predictive coding. arXiv preprint arXiv: 1807.03748  (2019)

\bibitem{radialog}
Pellegrini, C., Özsoy, E., Busam, B., Navab, N., Keicher, M.: Radialog: A large vision-language model for radiology report generation and conversational assistance (2023)

\bibitem{stanza}
Qi, P., Zhang, Y., Zhang, Y., Bolton, J., Manning, C.D.: Stanza: A {Python} natural language processing toolkit for many human languages. In: Proceedings of the 58th Annual Meeting of the Association for Computational Linguistics: System Demonstrations (2020), \url{https://nlp.stanford.edu/pubs/qi2020stanza.pdf}

\bibitem{clip}
Radford, A., Kim, J.W., Hallacy, C., Ramesh, A., Goh, G., Agarwal, S., Sastry, G., Askell, A., Mishkin, P., Clark, J., Krueger, G., Sutskever, I.: Learning transferable visual models from natural language supervision. In: Meila, M., Zhang, T. (eds.) ICML. Proceedings of Machine Learning Research, vol.~139, pp. 8748--8763. {PMLR} (2021), \url{http://proceedings.mlr.press/v139/radford21a.html}

\bibitem{radford2019language}
Radford, A., Wu, J., Child, R., Luan, D., Amodei, D., Sutskever, I., et~al.: Language models are unsupervised multitask learners. OpenAI blog  \textbf{1}(8), ~9 (2019)

\bibitem{chexnet}
Rajpurkar, P., Irvin, J., Zhu, K., Yang, B., Mehta, H., Duan, T., Ding, D., Bagul, A., Langlotz, C., Shpanskaya, K., et~al.: Chexnet: Radiologist-level pneumonia detection on chest x-rays with deep learning. arXiv preprint arXiv:1711.05225  (2017). \doi{10.48550/arXiv.1711.05225}

\bibitem{ramesh2022improving}
Ramesh, V., Chi, N.A., Rajpurkar, P.: Improving radiology report generation systems by removing hallucinated references to non-existent priors. In: Machine Learning for Health. pp. 456--473. PMLR (2022)

\bibitem{Hanoona2022Bridging}
Rasheed, H., Maaz, M., Khattak, M.U., Khan, S., Khan, F.S.: Bridging the gap between object and image-level representations for open-vocabulary detection. In: NIPS (2022)

\bibitem{frcnn}
Ren, S., He, K., Girshick, R., Sun, J.: Faster r-cnn: Towards real-time object detection with region proposal networks. NIPS  \textbf{28} (2015)

\bibitem{seibold_2022}
Seibold, C., Rei{\ss}, S., Sarfraz, M.S., Stiefelhagen, R., Kleesiek, J.: Breaking with fixed set pathology recognition through report-guided contrastive training. In: Wang, L., Dou, Q., Fletcher, P.T., Speidel, S., Li, S. (eds.) MICCAI. pp. 690--700. Springer Nature Switzerland, Cham (2022). \doi{10.1007/978-3-031-16443-9_66}

\bibitem{chexbert}
Smit, A., Jain, S., Rajpurkar, P., Pareek, A., Ng, A.Y., Lungren, M.P.: Chexbert: Combining automatic labelers and expert annotations for accurate radiology report labeling using bert (2020)

\bibitem{wbf}
Solovyev, R., Wang, W., Gabruseva, T.: Weighted boxes fusion: Ensembling boxes from different object detection models. Image and Vision Computing  \textbf{107},  104117 (2021). \doi{https://doi.org/10.1016/j.imavis.2021.104117}

\bibitem{sonsbeek_2023}
van Sonsbeek, T., Derakhshani, M.M., Najdenkoska, I., Snoek, C.G.M., Worring, M.: Open-ended medical visual question answering through prefix tuning of language models. In: Greenspan, H., Madabhushi, A., Mousavi, P., Salcudean, S., Duncan, J., Syeda-Mahmood, T., Taylor, R. (eds.) MICCAI. pp. 726--736. Springer Nature Switzerland, Cham (2023). \doi{10.1007/978-3-031-43904-9_70}

\bibitem{Sun_2022}
Sun, J., Wei, D., Wang, L., Zheng, Y.: Lesion Guided Explainable Few Weak-Shot Medical Report Generation, p. 615–625. Springer Nature Switzerland (2022). \doi{10.1007/978-3-031-16443-9_59}, \url{http://dx.doi.org/10.1007/978-3-031-16443-9_59}

\bibitem{rgrg}
Tanida, T., Müller, P., Kaissis, G., Rueckert, D.: Interactive and explainable region-guided radiology report generation. In: CVPR. pp. 7433--7442 (2023). \doi{10.1109/CVPR52729.2023.00718}

\bibitem{chexzero}
Tiu, E., Talius, E., Patel, P., Langlotz, C.P., Ng, A.Y., Rajpurkar, P.: Expert-level detection of pathologies from unannotated chest x-ray images via self-supervised learning. Nature Biomedical Engineering  \textbf{6}(12),  1399--1406 (2022)

\bibitem{medpalmm}
Tu, T., Azizi, S., Driess, D., Schaekermann, M., Amin, M., Chang, P.C., Carroll, A., Lau, C., Tanno, R., Ktena, I., Palepu, A., Mustafa, B., Chowdhery, A., Liu, Y., Kornblith, S., Fleet, D., Mansfield, P., Prakash, S., Wong, R., Virmani, S., Semturs, C., Mahdavi, S.S., Green, B., Dominowska, E., y~Arcas, B.A., Barral, J., Webster, D., Corrado, G.S., Matias, Y., Singhal, K., Florence, P., Karthikesalingam, A., Natarajan, V.: Towards generalist biomedical ai. NEJM AI  \textbf{1}(3),  AIoa2300138 (2024). \doi{10.1056/AIoa2300138}

\bibitem{wang_2022_neurips}
Wang, F., Zhou, Y., Wang, S., Vardhanabhuti, V., Yu, L.: Multi-granularity cross-modal alignment for generalized medical visual representation learning. In: Koyejo, S., Mohamed, S., Agarwal, A., Belgrave, D., Cho, K., Oh, A. (eds.) NeurIPS (2022)

\bibitem{wang2022inclusive}
Wang, L., Ning, M., Lu, D., Wei, D., Zheng, Y., Chen, J.: An inclusive task-aware framework for radiology report generation. In: MICCAI. pp. 568--577 (2022)

\bibitem{beit3}
Wang, W., Bao, H., Dong, L., Bjorck, J., Peng, Z., Liu, Q., Aggarwal, K., Mohammed, O.K., Singhal, S., Som, S., Wei, F.: Image as a foreign language: Beit pretraining for vision and vision-language tasks. In: CVPR. pp. 19175--19186 (2023). \doi{10.1109/CVPR52729.2023.01838}

\bibitem{nih8}
Wang, X., Peng, Y., Lu, L., Lu, Z., Bagheri, M., Summers, R.M.: Chestx-ray8: Hospital-scale chest x-ray database and benchmarks on weakly-supervised classification and localization of common thorax diseases. In: CVPR. pp. 2097--2106 (2017). \doi{10.1109/CVPR.2017.369}

\bibitem{wang_metrans_2023}
Wang, Z., Liu, L., Wang, L., Zhou, L.: Metransformer: Radiology report generation by transformer with multiple learnable expert tokens. In: CVPR. pp. 11558--11567 (2023). \doi{10.1109/CVPR52729.2023.01112}

\bibitem{wang2022medclip}
Wang, Z., Wu, Z., Agarwal, D., Sun, J.: {M}ed{CLIP}: Contrastive learning from unpaired medical images and text. In: Conference on Empirical Methods in Natural Language Processing. pp. 3876--3887. Association for Computational Linguistics, Abu Dhabi, United Arab Emirates (2022), \url{https://aclanthology.org/2022.emnlp-main.256}

\bibitem{cig}
Wu, J., Agu, N., Lourentzou, I., Sharma, A., Paguio, J.A., Yao, J.S., Dee, E.C., Kashyap, S., Giovannini, A., Celi, L.A., et~al.: Chest imagenome dataset for clinical reasoning. In: NIPS (2021)

\bibitem{cig2}
Wu, J.T., Agu, N.N., Lourentzou, I., Sharma, A., Paguio, J.A., Yao, J.S., Dee, E.C., Mitchell, W., Kashyap, S., Giovannini, A., et~al.: Chest imagenome dataset (version 1.0.0). PhysioNet  (2021). \doi{https://doi.org/10.13026/wv01-y230}

\bibitem{cora}
Wu, X., Zhu, F., Zhao, R., Li, H.: Cora: Adapting clip for open-vocabulary detection with region prompting and anchor pre-matching. In: CVPR. pp. 7031--7040 (2023). \doi{10.1109/CVPR52729.2023.00679}

\bibitem{wu_2023_miccai}
Wu, Y., Zhou, Y., Saiyin, J., Wei, B., Lai, M., Shou, J., Fan, Y., Xu, Y.: Zero-shot nuclei detection via visual-language pre-trained models. In: Greenspan, H., Madabhushi, A., Mousavi, P., Salcudean, S., Duncan, J., Syeda-Mahmood, T., Taylor, R. (eds.) MICCAI. pp. 693--703. Springer Nature Switzerland, Cham (2023). \doi{10.1007/978-3-031-43987-2_67}

\bibitem{omnifm}
Xu, L., Ni, Z., Liu, X., Wang, X., Li, H., Zhang, S.: Learning a multi-task transformer via unified and customized instruction tuning for chest radiograph interpretation (2023)

\bibitem{xu2023elixr}
Xu, S., Yang, L., Kelly, C., Sieniek, M., Kohlberger, T., Ma, M., Weng, W.H., Kiraly, A., Kazemzadeh, S., Melamed, Z., Park, J., Strachan, P., Liu, Y., Lau, C., Singh, P., Chen, C., Etemadi, M., Kalidindi, S.R., Matias, Y., Chou, K., Corrado, G.S., Shetty, S., Tse, D., Prabhakara, S., Golden, D., Pilgrim, R., Eswaran, K., Sellergren, A.: Elixr: Towards a general purpose x-ray artificial intelligence system through alignment of large language models and radiology vision encoders (2023)

\bibitem{yang2022_unitab}
Yang, Z., Gan, Z., Wang, J., Hu, X., Ahmed, F., Liu, Z., Lu, Y., Wang, L.: Unitab: Unifying text and box outputs for grounded vision-language modeling. In: Avidan, S., Brostow, G., Ciss{\'e}, M., Farinella, G.M., Hassner, T. (eds.) ECCV. pp. 521--539. Springer Nature Switzerland, Cham (2022). \doi{10.1007/978-3-031-20059-5_30}

\bibitem{ovdetr}
Zang, Y., Li, W., Zhou, K., Huang, C., Loy, C.C.: Open-vocabulary detr with conditional matching. In: Avidan, S., Brostow, G., Ciss{\'e}, M., Farinella, G.M., Hassner, T. (eds.) ECCV. pp. 106--122. Springer Nature Switzerland, Cham (2022). \doi{10.1007/978-3-031-20077-9_7}

\bibitem{ovr}
Zareian, A., Rosa, K.D., Hu, D.H., Chang, S.F.: Open-vocabulary object detection using captions. In: CVPR. pp. 14388--14397 (2021). \doi{10.1109/CVPR46437.2021.01416}

\bibitem{samdetr}
Zhang, G., Luo, Z., Yu, Y., Cui, K., Lu, S.: Accelerating detr convergence via semantic-aligned matching. In: CVPR. pp. 939--948 (2022). \doi{10.1109/CVPR52688.2022.00102}

\bibitem{glip2}
Zhang, H., Zhang, P., Hu, X., Chen, Y.C., Li, L., Dai, X., Wang, L., Yuan, L., Hwang, J.N., Gao, J.: Glipv2: Unifying localization and vision-language understanding. In: Koyejo, S., Mohamed, S., Agarwal, A., Belgrave, D., Cho, K., Oh, A. (eds.) NeurIPS. vol.~35, pp. 36067--36080. Curran Associates, Inc. (2022), \url{https://proceedings.neurips.cc/paper_files/paper/2022/file/ea370419760b421ce12e3082eb2ae1a8-Paper-Conference.pdf}

\bibitem{zhang2024biomedgpt}
Zhang, K., Yu, J., Adhikarla, E., Zhou, R., Yan, Z., Liu, Y., Liu, Z., He, L., Davison, B., Li, X., Ren, H., Fu, S., Zou, J., Liu, W., Huang, J., Chen, C., Zhou, Y., Liu, T., Chen, X., Chen, Y., Li, Q., Liu, H., Sun, L.: Biomedgpt: A unified and generalist biomedical generative pre-trained transformer for vision, language, and multimodal tasks (2024)

\bibitem{convirt}
Zhang, Y., Jiang, H., Miura, Y., Manning, C.D., Langlotz, C.P.: Contrastive learning of medical visual representations from paired images and text. In: Lipton, Z., Ranganath, R., Sendak, M., Sjoding, M., Yeung, S. (eds.) Machine Learning for Healthcare Conference. Proceedings of Machine Learning Research, vol.~182, pp. 2--25. PMLR (05--06 Aug 2022)

\bibitem{regionclip}
Zhong, Y., Yang, J., Zhang, P., Li, C., Codella, N., Li, L.H., Zhou, L., Dai, X., Yuan, L., Li, Y., Gao, J.: Regionclip: Region-based language-image pretraining. In: CVPR. pp. 16772--16782 (2022). \doi{10.1109/CVPR52688.2022.01629}

\bibitem{detic}
Zhou, X., Girdhar, R., Joulin, A., Kr{\"a}henb{\"u}hl, P., Misra, I.: Detecting twenty-thousand classes using image-level supervision. In: Avidan, S., Brostow, G., Ciss{\'e}, M., Farinella, G.M., Hassner, T. (eds.) Computer Vision -- ECCV 2022. pp. 350--368. Springer Nature Switzerland, Cham (2022). \doi{10.1007/978-3-031-20077-9_21}

\bibitem{zhu2020deformable}
Zhu, X., Su, W., Lu, L., Li, B., Wang, X., Dai, J.: Deformable {DETR:} deformable transformers for end-to-end object detection. In: ICLR (2021), \url{https://openreview.net/forum?id=gZ9hCDWe6ke}

\bibitem{codetr}
Zong, Z., Song, G., Liu, Y.: Detrs with collaborative hybrid assignments training. In: ICCV. pp. 6725--6735 (2023). \doi{10.1109/ICCV51070.2023.00621}

\end{thebibliography}

\newpage
\appendix
\section{Ablation Studies and Further Analysis}
\subsection{Multitask Training: Token and Target Types}
\begin{center}
    \captionof{table}{Ablation study on the forms of supervision (\ie token and target types) used for multitask training.  We study using only pathology tokens (P-only), only anatomy tokens (A-only), and only sentence tokens (S-only). Additionally, we study the exclusion of specific tokens types (No-P, No-A, No-S). We also study the exclusion of specific target types like bounding box targets (no box), pathology class targets (No cls), and sentence targets (No sent).
    Best results and those within one std are marked in bold. 
    The symbols \mimic, \vindr, and \nih  \ indicate that images from MIMIC-CXR, VinDr-CXR, and NIH8 are used for training, respectively.}
    \label{tab:abl_supervision}
    \tiny
    \setlength{\tabcolsep}{2pt}
    \begin{tabular}{llcccccccccc}
    \toprule
    & & ChEX & \multicolumn{6}{c}{Token Types} & \multicolumn{3}{c}{Target Types} \\ 
    \cmidrule(lr){3-3}\cmidrule(lr){4-9} \cmidrule(lr){10-12} &&& P-only & A-only & S-only & No-P & No-A & No-S & No box & No cls & No sent \\   
    \multicolumn{2}{l}{Pathology boxes\vindr} & \checkmark & \checkmark & &&& \checkmark & \checkmark & & \checkmark & \checkmark\\
    \multicolumn{2}{l}{Pathology classes\vindr} & \checkmark & \checkmark & &&& \checkmark & \checkmark & \checkmark & & \checkmark\\
    \multicolumn{2}{l}{Anat. boxes\mimic} & \checkmark & & \checkmark & & \checkmark && \checkmark & & \checkmark & \checkmark \\
    \multicolumn{2}{l}{Anat. patho classes\mimic} & \checkmark & & \checkmark & & \checkmark && \checkmark & \checkmark & & \checkmark \\
    \multicolumn{2}{l}{Anat. sentences\mimic} & \checkmark & & \checkmark & &\checkmark && \checkmark & \checkmark & \checkmark & \\ 
    \multicolumn{2}{l}{Sentences\mimic} & \checkmark & && \checkmark & \checkmark & \checkmark &  & \checkmark &\checkmark & \\ 
    \midrule
    \multicolumn{5}{l}{\textbf{Sentence Grounding (SG)}} \\
    MS-CXR\mimic & [mIoU] & 47.52 & 42.34 & 45.07 & 9.00 & 23.90 & 34.92 & \textbf{52.97} & 3.68 & 44.38 & \textbf{52.97} \\
    & [mAP] & 44.47 & 37.92 & 40.47 & 0.94 & 18.89 & 19.55 & \textbf{50.66} & 3.71 & 36.43 & \textbf{50.11} \\
    \midrule
    \multicolumn{5}{l}{\textbf{Pathology Detection (OD)}} \\
    VinDr\vindr & [mAP] & \textbf{14.12} & 13.25 & 4.60 & 0.06 & 1.16 & \textbf{14.15} & \textbf{14.38} & 0.22 & 10.35 & \textbf{14.38} \\
    NIH8\nih & [mAP] & \textbf{11.14} & 7.02 & 4.40 & 0.22 & 2.23 & 5.38 & 7.96 & 0.56 & 6.70 & 7.96 \\
    MS-CXR\mimic & [mAP] & \textbf{16.60} & 13.52 & 10.90 & 0.26 & 8.83 & 5.51 & 14.11 & 0.52 & 14.96 & 14.11 \\
    \midrule
    \multicolumn{5}{l}{\textbf{Region Classification (RC)}} \\
    MS-CXR\mimic & [AUROC] & \textbf{82.33} & 74.02 & \textbf{79.86} & 49.49 & 76.58 & 69.10 & \textbf{80.89} & 46.87 & \textbf{79.89} & 75.67 \\ 
    CIG\mimic & [wAUROC] & \textbf{70.46} & 59.36 & 69.61 & 50.08 & 68.58 & 62.28 & 69.61 & 58.03 & \textbf{70.16} & 69.89 \\ 
    \midrule
    \multicolumn{5}{l}{\textbf{Region Explanation (RE)}} \\
    MS-CXR\mimic & [Mic-F1-14] & \textbf{49.97} & 45.44\textsuperscript{\textdagger} & 39.47 & 10.94 & 22.49 & 35.37 & 42.72 & 5.53 & 35.02 & 43.16\textsuperscript{\textdagger} \\ 
    & [Mac-F1-14] & \textbf{20.50} & 17.99\textsuperscript{\textdagger} & \textbf{20.29} & 1.48 & 15.19 & 9.37 & \textbf{20.01} & 2.61 & 15.17 & 14.40\textsuperscript{\textdagger} \\
    & [METEOR] & \textbf{8.79} & 6.91\textsuperscript{\textdagger} & \textbf{9.02} & 0.00 & 6.13 & 5.69 & 7.74 & 2.96 & 7.77 & 5.73\textsuperscript{\textdagger} \\
    CIG\mimic & [Mic-F1-14] & \textbf{53.34} & 35.26\textsuperscript{\textdagger} & 51.25 & 10.96 & 47.51 & 28.76 & 51.84 & 23.47 & \textbf{53.22} & 48.93\textsuperscript{\textdagger} \\ 
    & [Mac-F1-14] & \textbf{29.13} & 16.00\textsuperscript{\textdagger} & 27.65 & 1.66 & 24.14 & 14.82 & 28.20 & 10.73 & \textbf{29.18} & 24.36\textsuperscript{\textdagger} \\
    & [METEOR] & \textbf{10.18} & 5.48\textsuperscript{\textdagger} & 9.57 & 0.00 & 7.95 & 4.70 & 9.79 & 3.47 & \textbf{10.08} & 8.89\textsuperscript{\textdagger} \\
    \midrule
    \multicolumn{5}{l}{\textbf{Full report generation (RG)}} \\
    MIMIC-CXR\mimic & [Mic-F1-14] & \textbf{52.32} & 50.10\textsuperscript{\textdagger} & 49.64 & 16.73 & 43.44 & 46.64 & 49.91 & 40.68 & 51.69 & 51.18\textsuperscript{\textdagger}  \\ 
    & [Mac-F1-14] & \textbf{32.56} & 26.76\textsuperscript{\textdagger} & 29.87 & 3.22 & 27.60 & 25.35 & 31.32 & 22.83 & 30.75 & 29.81\textsuperscript{\textdagger} \\
    & [Ex-F1-14] & \textbf{58.76} & \textbf{59.04}\textsuperscript{\textdagger} & 57.60 & 50.25 & 50.10 & 55.02 & 57.39 & 55.93 & 57.39 & 58.44\textsuperscript{\textdagger} \\
    & [Mic-F1-5+] & \textbf{61.03} & 58.16\textsuperscript{\textdagger} & 57.09 & 44.28 & 52.10 & 54.43 & 57.77 & 49.79 & \textbf{60.58} & 58.63\textsuperscript{\textdagger} \\
    & [Mac-F1-5+] & \textbf{55.85} & 48.65\textsuperscript{\textdagger} & 52.31 & 22.81 & 49.96 & 50.20 & 52.55 & 40.75 & 54.54 & 51.62\textsuperscript{\textdagger} \\
    & [METEOR] & 13.26 & 11.13\textsuperscript{\textdagger} & 12.45 & 0.00 & 12.69 & 10.97 & 13.05 & 8.01 & \textbf{13.55} & 11.99\textsuperscript{\textdagger} \\
    \bottomrule
    \end{tabular} \\
    \textsuperscript{\textdagger} We additionally trained the sentence generator afterwards (with anatomy and report sentences). \\
\end{center}

\subsubsection{Token Types}
We study the effect of different forms of supervision (\ie token and target types) during multitask training (\cref{tab:abl_supervision}). In the default case, ChEX is trained with all token types (pathology, anatomy, sentence) and all target types (bounding boxes, pathology classification, sentences). Using only pathology tokens (with bounding box and class targets), \ie only samples from VinDr-CXR, leads to performance reductions on SG, OD, and RC tasks. When additionally using sentence supervision for training only the sentence generator, the performance on the RE task on MS-CXR and the RG task is only slightly degraded compared to ChEX, while there is a huge performance drop on the RE task on CIG. Using only anatomy tokens (with bounding boxes, pathology labels, and sentences) leads to a further performance drop on OD, RE on MS-CXR, and RG, while performance on SG and RE on CIG is improved. Using only sentence tokens leads to very poor performance on all tasks. Similarly, using no pathology tokens or no anatomy tokens leads to very poor performance on most tasks, with the exception that using no anatomy tokens leads to good performance on VinDR-CXR OD. Using no sentence tokens only leads to small performance drops on most tasks and improves performance on SG as well as on VinDr-CXR OD, while only leading to meaningfully performance drops on the other two OD tasks.

\subsubsection{Target Types}
We also study the use of all token types while excluding specific target types. As expected, excluding bounding box targets leads to very poor performance on all tasks involving localization, \ie all SG and OD tasks. Huge performance drops can also be observed on RE and RG tasks, while drops are smaller on RC tasks. Excluding pathology classification labels (for pathology and anatomy tokens) reduces the performance mainly on OD tasks and RE on MS-CXR, while smaller performance drops can be observed on SG and RG. Almost no performance drops are observed on the RC tasks and RE on CIG. Excluding sentence targets (\ie excluding sentence targets for anatomy tokens and all sentence tokens) leads to performance drops on most tasks, except on SG and OD on VinDr-CXR, where performance improves. 

\FloatBarrier

\subsection{Sentence Generator}\label{sec:abl_sentgen}
In \cref{tab:abl_modality_map}, we provide an ablation study on design choices of the sentence generator. More precisely, we studied (i) the relevance of the post decoder, a component of the sentence decoder; (ii) variations of the query-key-value (qky) features used in P-tuning (linear projection instead of MLP, shared projections for all decoder layers, shorter prefix length); and (iii) the benefits of decoder language model pre-training on MIMIC-CXR. Overall, we found that most design decisions, except the post decoder, have little impact. 

\begin{table}[h!]
    \centering\scriptsize
    \setlength{\tabcolsep}{2pt}
    \begin{tabular}{lccc}
    \toprule
    \textbf{Sentence Generator} & MS-CXR {\tiny[Mic-F1-14]} & CIG {\tiny[Mic-F1-14]}& MIMIC-CXR {\tiny[Mic-F1-14]} \\
    \midrule
    ChEX & \textbf{49.97} & \textbf{53.34} & \textbf{52.32} \\ 
    no post decoder & 45.86 & 43.90 & 46.07 \\
    linear qkv proj. & \textbf{48.64} & 52.80 & \textbf{52.41} \\
    layer-shared qkv & \textbf{48.25} & \textbf{53.38} & \textbf{52.46} \\
    prefix length 1 & \textbf{49.65} & \textbf{53.14} & \textbf{52.25} \\
    no gen.\@ pretraining & \textbf{48.56} & 52.89 & \textbf{52.46} \\
    \bottomrule
    \end{tabular}
    \caption{Ablation studies of the sentence generator.}
    \label{tab:abl_modality_map}
\end{table}

We argue that the post decoder is so beneficial, because it enables further processing of ROI features even if only bounding boxes where given as queries (where huge parts of the detector are skipped).
The choice of the projections during P-tuning has little influence, such that we opted for the default configuration (MLP projections without sharing). While the prefix length has minimal influence, a slightly longer prefix length of 5 (as used in CHeX), provide some minor improvements across all generation tasks.
Skipping pre-training of the decoder on MIMIC-CXR and directly using the decoder language model pre-trained only on PubMed, leads to small performance degradation on some tasks. However, the main benefit of this pre-training is a faster convergence in the final end-2-end training.

\subsection{Query Type for Region Classification and Explanation}

\begin{table}[ht!]
    \centering\scriptsize
    \caption{Effect of query types on region classification (RC) and region explanation (RE) on the CIG dataset. We compare the setting using box queries (the default setting) and the setting using textual prompts (based on the names of the anatomical regions).}
    \label{tab:abl_prompting}
    \setlength{\tabcolsep}{5pt}
    \begin{tabular}{lcc}
    \toprule
     & \multicolumn{2}{c}{Chest ImaGenome (CIG)} \\
     \cmidrule(lr){2-3} & Region Classification (RC) & Region Explanation (RE) \\
     Query type & [wAUROC] & [Mic-F1-14] \\
    \midrule \\
    Boxes (default task) & 70.46\std{0.36} & 53.34\std{0.43}  \\
    Textual prompt & 70.76\std{0.37} & 54.70\std{0.41} \\
    \bottomrule
    \end{tabular}
\end{table}
\cref{tab:abl_prompting} shows the effect of using bounding boxes versus textual prompts as queries for RC and RE on the CIG dataset, \ie for anatomical regions. When using textual prompts, \ie, the names of anatomical regions, instead of their bounding boxes (the default setting), the performance of ChEX slightly improves. We assume that these improvements are caused by the additional context provided by the region prompts. Overall, the differences are marginal, showing that our model ChEX is robust to the choice of user query.

\FloatBarrier

\newpage

\subsection{Prompt Sets for Report Generation}\label{sec:prompt_sets}
\begin{table}[h]
    \centering\scriptsize
    \caption{Effect of different prompt sets (anatomy or pathology prompts or both) and filtering (using only regions with positive pathologies) for full report generation. The default setting used by ChEX is highlighted in grey, best results are marked in bold.}
    \label{tab:abl_reportgen}
    \setlength{\tabcolsep}{2pt}
    \begin{tabular}{llcccccc}
    \toprule
     \multicolumn{2}{c}{Prompt sets} & \multicolumn{6}{c}{MIMIC-CXR} \\
    \cmidrule(lr){1-2} \cmidrule(lr){3-8} Anatomy & Pathology & [Mic-F1-14] & [Mac-F1-14] & [Ex-F1-14] & [Mic-P-14] & [Mic-R-14] & [METEOR] \\
    \midrule 
    yes & yes & 51.04 & 30.61 & 57.05 & 42.99 & \textbf{62.80} & \textbf{13.90} \\
    yes & positive & 50.13 & 30.67 & 56.91 & 42.46 & 61.22 & 13.60 \\
    yes & no & 50.08 & 30.07 & 56.97 & 42.66 & 60.64 & 12.67 \\
    \rowcolor[gray]{.9}positive & yes & \textbf{52.37} & 32.69 & 58.78 & 45.09 & 62.44 & 13.28 \\
    positive & positive & 52.02 & \textbf{32.88} & 58.36 & 44.64 & 62.33 & 12.24\\
    positive & no & 51.27 & 32.18 & 58.82 & 45.16 & 59.31 & 10.77 \\
    no & yes & 50.35 & 25.82 & 59.35 & 47.88 & 53.10 & 9.82 \\
    no & positive & 50.39 & 26.69 & \textbf{60.14} & \textbf{49.38} & 51.44 & 8.26 \\
    \bottomrule
    \end{tabular}
\end{table}

ChEX uses pre-defined prompt sets to enable full report generation (RG).
In \cref{tab:abl_reportgen}, we study the effect of using pathology and anatomical region prompts. We additionally study the effect of excluding regions without pathologies (based on contrastive classification of the CheXpert\cite{chexpert} pathologies). 
Overall, all studied prompt sets lead to results competitive with the baselines, with Mic-F1-14 ranging from $50.08\%$ to $52.37\%$, but the selection of the prompt set influences the relation between precision (Mic-P-14) and recall (Mic-R-14).  
Using only pathology prompts leads to a high precision but achieves much smaller macro F1 scores than also using anatomy prompts (with positive filtering). This is explained by the poor performance on a few pathologies (adding an anatomical prompt with positive filtering to the prompt set achieves $>4$-fold improvement for enlarged cardiomediastinum and $>11$-fold improvement for lung lesions). Therefore, using anatomical prompts can spot additional pathologies that may be missed by pathology prompts. Similarly, using only anatomy prompts leads to a small recall because pathologies may again be missed. Using both prompt sets leads to the best results. However, considering all anatomical regions can lead to overprediction (lower precision) and, therefore, lower overall F1 scores. We found that using positive filtering with anatomy prompts (\ie considering only anatomical regions with positive pathologies for generation) improves performance. The best setting for Mic-F1-14 uses positive filtering for anatomy prompts and no filtering for pathology prompts. We consider this the default setting of ChEX, and all reported results from the main paper follow this setting.
The best Mac-F1-14 can be achieved using positive filtering for both prompt sets. The best METEOR score and recall can be achieved using no positive filtering with both prompt sets, while the highest precision is achieved using only pathology prompts with positive filtering. 
We highlight that changing the prompt set does not require re-training.

\newpage

\subsection{Sensitivity to Prompt Modifications}
\begin{figure}[ht!]
    \centering
    \includegraphics[width=1.0\textwidth]{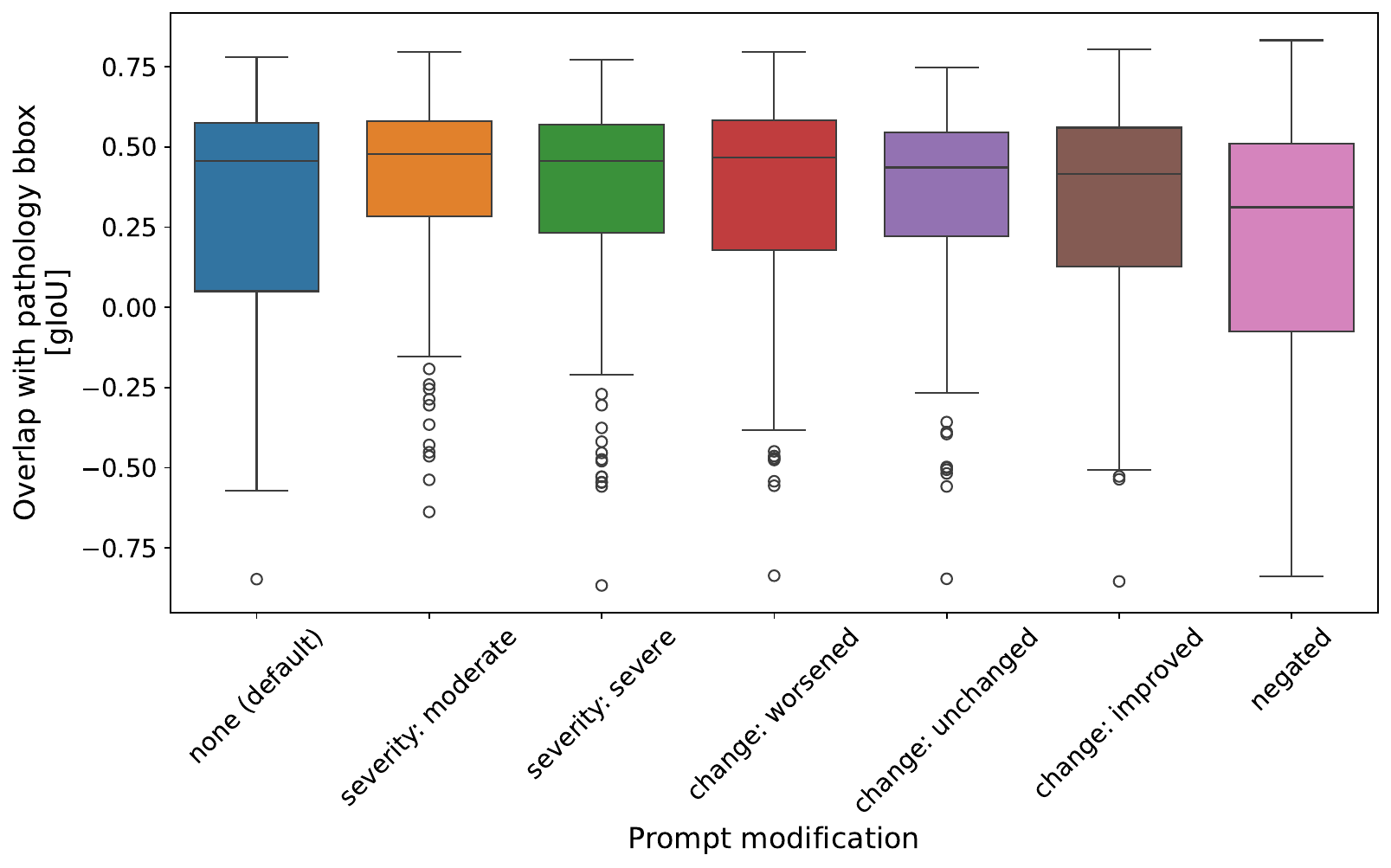}
    \caption{Effect of different prompting strategies on pathology localization performance}
    \label{fig:prompt_effect}
\end{figure}
\cref{fig:prompt_effect} shows the effect of different prompting strategies on pathology localization performance. Modifying the default prompt (\ie just the pathology name) by adding severity modifications reduces the overall variance of localization quality, where \say{moderate} severity leads to the best performance. Using change indications also leads to a reduction in variance, albeit smaller than the severity modifications. The best performance can be achieved with \say{worsened} while \say{improved} leads to slightly poorer performance, probably because this indicates negative findings. However, all of these prompt modifications have little influence on the median results. Negating the prompt, on the other hand, leads to a noticeable drop in localization performance. Overall, this indicates that our model ChEX is robust to most modifications of prompts.

\FloatBarrier

\newpage
\subsection{Learned Regional Bias for Pathology Prompts}

\begin{figure}[ht!]
    \centering
    \includegraphics[width=0.7\textwidth]{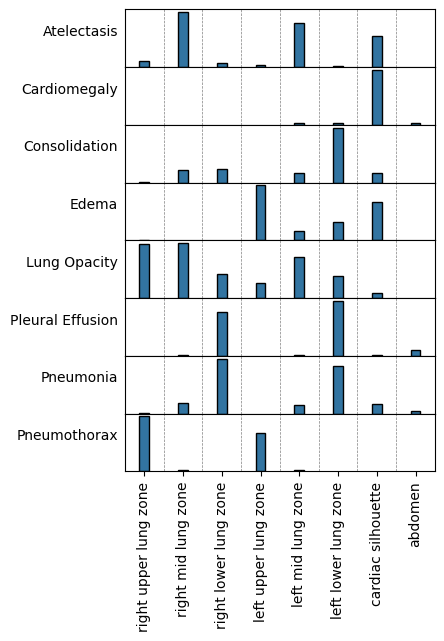}
    \caption{Regional bias of pathology prompts in negative images. For each pair of pathology and anatomical region, the bar indicates the number of cases where the given region has the highest overlap, among all shown regions, with the bounding box of the given pathology prompt.}
    \label{fig:regional_bias}
\end{figure}

In \cref{fig:regional_bias}, we study the regional bias of pathology prompts in negative images. Cardiomegaly prompts typically overlap with the cardiac silhouette, while most other pathologies mostly overlap with lung regions, as expected. More noteworthy, pleural effusion is typically searched in the lower lung zones, while pneumothorax is searched in the upper lung zones, which aligns well with the typical locations of these pathologies in chest X-rays. This indicates that our model has learned a meaningful regional bias for pathologies.

\FloatBarrier

\newpage

\section{Qualitative Examples and Failure Cases}
\begin{figure}[h!]
    \centering
    \includegraphics[width=.85\textwidth]{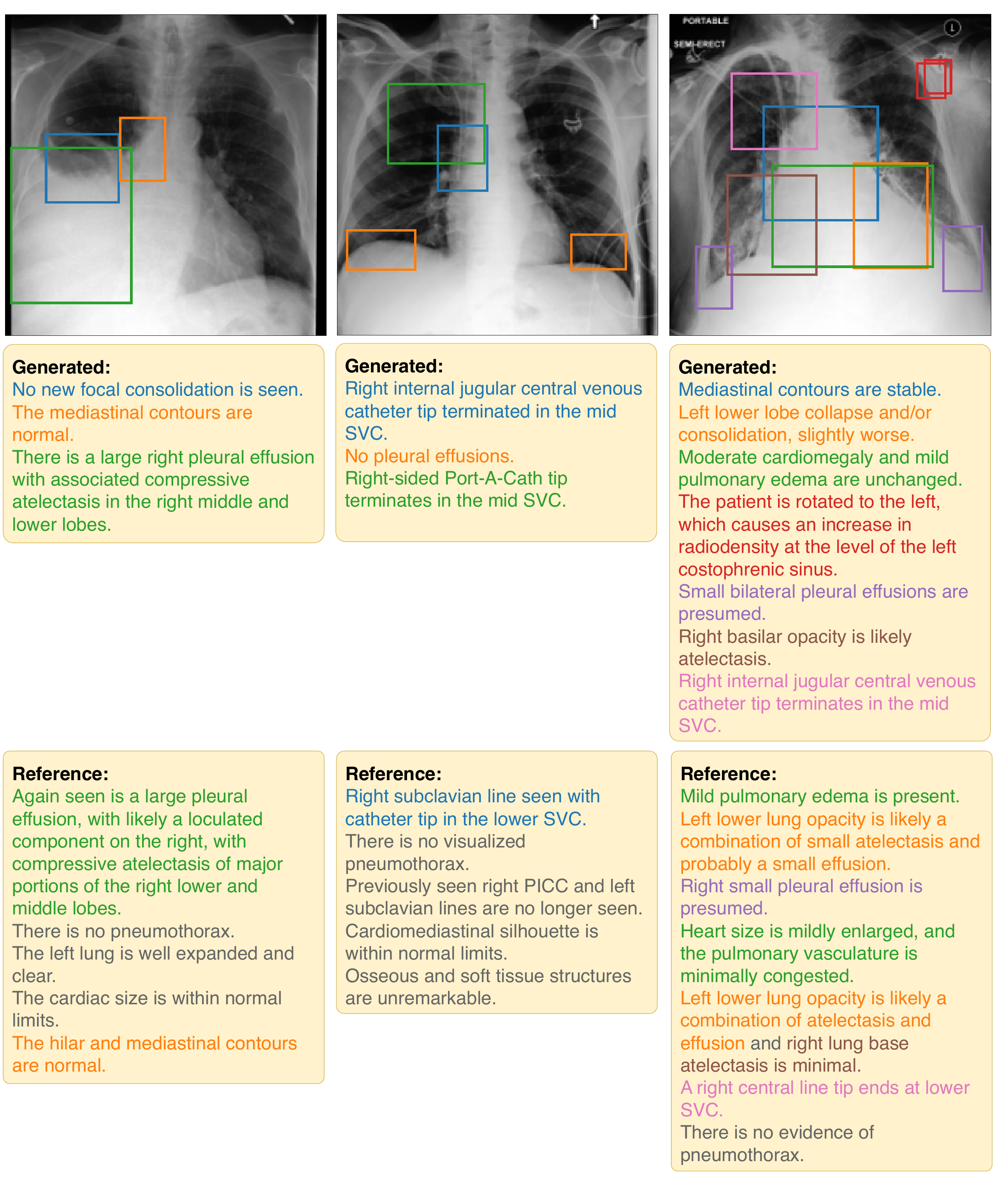}
    \caption{Full report generation with predicted bounding boxes on three example chest X-rays. All positive findings in the reference reports have been correctly captured in the generated reports, while reported negative findings can deviate. Changes like the removal of the right PICC (middle) cannot be captured by our model as it only considers a single image but not the temporal sequence. Terms relating to changes like \protect\say{new} (left, blue) or \protect\say{unchanged} (right, green) in the generated sentences are hallucinated due to the used training data, a phenomenon common to report generation models \cite{maira,biovil_t,ramesh2022improving}.}
\end{figure}
\FloatBarrier
\begin{figure}[h!]
    \centering
    \includegraphics[width=.82\textwidth]{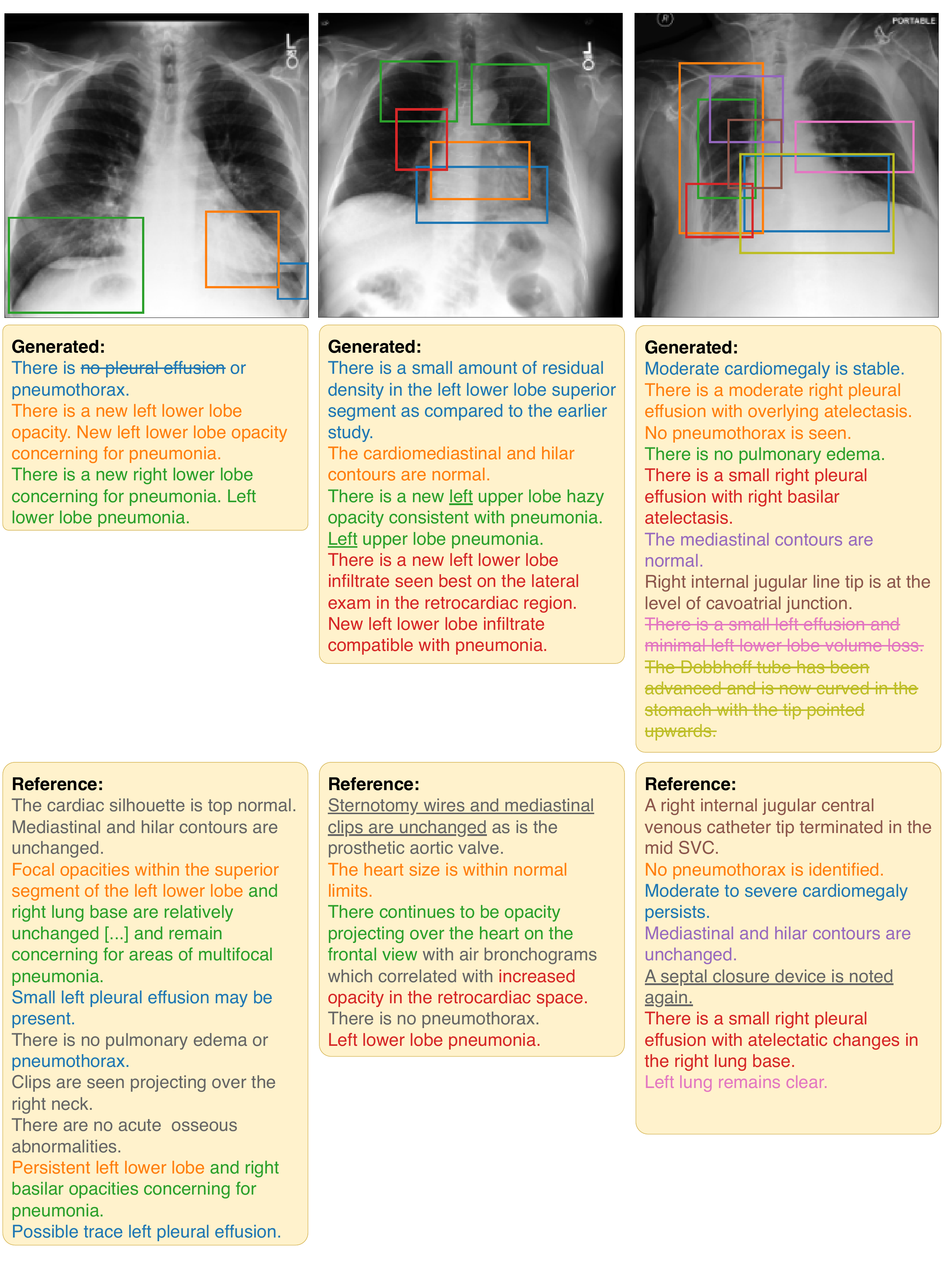}
    \caption{Example failure cases in full report generation. \textbf{Left:} The model incorrectly predicts that no pleural effusion is present, while there may be a trace of a left pleural effusion (blue). However, it predicts a bounding box for a potential left pleural effusion, thereby enabling correctness checking by a radiologist. \textbf{Middle:} The model correctly predicts that there is left lower lobe pneumonia (red). However, it (inaccurately) predicts that there might be additional left upper lobe pneumonia while predicting bounding boxes in both lungs, \ie one of the bounding boxes does not match the prediction (green). \textbf{Right:} The model incorrectly predicts that there is a small left effusion (pink), which can again be checked and corrected using the bounding box. It also incorrectly predicts the presence of a Dobhoff tube (yellow) which it may confuse with the correctly detected catheter tip (brown). Overall, the predictions cover most positive findings correctly. The provided bounding boxes for positive and negative findings enable simple checking of nuanced cases and cases where the model is uncertain.}
\end{figure}

\FloatBarrier
\newpage

\section{Detailed Benchmark Results}
\subsection{Sentence Grounding (SG) Task}
\begin{table}[h!]
\caption{Sentence grounding (SG) results on the MS-CXR dataset. We indicate the use of training or evaluation images from VinDr-CXR by \vindr, from MIMIC-CXR by \mimic, and from other sources by \other. 
All results are based on our own experiments, except OmniFM-DR for which we show the results reported by the original paper.
We provide the std based on bootstrapping with $N=250$. Best results and those within 1-std are marked in bold.}
\label{tab:results_sg}
\vspace{\floatsep}
\begin{center}
\centering\scriptsize
\begin{tabular}{lcc}
    \toprule
      & \multicolumn{2}{c}{Sentence Grounding (SG)}  \\
     \cmidrule(lr){2-3} & \multicolumn{2}{c}{MS-CXR\mimic} \\
     & [mIoU] & [mAP]\\ 
     \midrule
     ChEX\vindr\mimic & 47.52\std{1.45} & \textbf{44.47\std{2.21}} \\ 
     \midrule
     \multicolumn{2}{l}{\textbf{Zeroshot Contrastive Models}} \\
     BioVIL\cite{biovil} (masks) \mimic & 22.75\std{1.26} & -- \\ 
     BioVIL\cite{biovil} (boxes) \mimic & 28.57\std{1.31} & 18.62\std{1.37} \\
     CheXzero\cite{chexzero} (masks) \mimic & 11.94\std{0.59} & -- \\ 
     CheXzero\cite{chexzero} (boxes) \mimic & 15.45\std{0.67} & 5.94\std{0.64} \\
     \midrule
    \multicolumn{2}{l}{\textbf{Multitask Generative Models}} \\
     OmniFM-DR\cite{omnifm} \vindr\mimic\other & 46.2 & -- \\ 
     \midrule
     \multicolumn{2}{l}{\textbf{Supervised Visual Grounding}} \\
     TransVG\cite{transvg} \mimic & 48.81\std{1.45} & 37.65\std{2.61} \\
     TransVG\cite{transvg} (BioVIL\cite{biovil}) \mimic & \textbf{52.13\std{1.73}} & 41.24\std{2.73} \\
     TransVG\cite{transvg} (CheXzero\cite{chexzero}) \mimic & \textbf{53.51\std{1.53}} & \textbf{44.05\std{2.63}} \\
  \bottomrule
\end{tabular} 
\end{center}
\end{table}

In \emph{sentence grounding (SG)} (\cref{tab:results_sg}), ChEX demonstrates similar performance to the only generative model OmniFM-DR\cite{omnifm} and the best SupVG model TransVG\cite{transvg} (with CheXzero\cite{chexzero} backbone), where TransVG shows an advantage on the mIoU metric. Notably, TransVG was trained explicitly on this task. ChEX outperforms TransVG without the CheXzero backbone by $18\%$ mAP and is within 1-std on mIoU. 
The superiority of ChEX over contrastive models
underscores the importance of bounding box supervision for accurate localization.

\FloatBarrier
\newpage

\subsection{Pathology Detection (OD) Tasks}
\captionof{table}{Pathology detection (OD) results on the VinDr-CXR, NIH8, and MS-CXR datasets. We indicate the use of training or evaluation images from VinDr-CXR by \vindr, from NIH8 by \nih, and from MIMIC-CXR by \mimic. All results are based on our own experiments and we provide the std based on bootstrapping with $N=250$. Best results and those within 1-std are marked in bold.}
\label{tab:restuls_loccls}
\vspace{\floatsep}
\begin{center}
\centering\scriptsize
\begin{tabular}{lrrr}
        \toprule
        & \multicolumn{3}{c}{Pathology Detection (OD)} \\
         \cmidrule(lr){2-4}  
        & VinDr\vindr & NIH8\nih & MS-CXR\mimic  \\
         &[mAP] & [mAP] & [mAP] \\
         \midrule
         ChEX\vindr\mimic & 14.12\std{0.95} & \textbf{11.14}\std{1.05} & \textbf{16.60}\std{1.38} \\ 
         \midrule
         \multicolumn{4}{l}{\textbf{Supervised Object Detection}} \\
        FRCNN\cite{frcnn} \vindr & \textbf{18.21}\std{1.20} & 6.69\std{0.82} &  15.13\std{1.22} \\ 
        FRCNN\cite{frcnn} (BioVIL\cite{biovil}) \vindr\mimic & 16.17\std{1.31}  & 6.56\std{0.90} & \textbf{15.64}\std{1.16} \\
        FRCNN\cite{frcnn} (CheXzero\cite{chexzero}) \vindr\mimic  & \multicolumn{3}{c}{\emph{no convergence}}  \\
        DETR-R50\cite{detr} \vindr & 11.91\std{0.96} & 5.90\std{0.85} & 8.94\std{0.35} \\
        DETR-BioVIL\cite{detr,biovil} \vindr\mimic & 3.82\std{0.59} & 2.46\std{0.46} & 1.94\std{0.48} \\ 
        DETR-CheXzero\cite{detr,chexzero} \vindr\mimic & 1.26\std{0.39} & 1.18\std{0.34} & 0.18\std{0.15} \\ 
        DETR-DC5-R50\cite{detr} \vindr & 11.94\std{0.89} & 4.21\std{0.61} & 10.30\std{0.87} \\ 
        DETR-DC5-BioVIL\cite{detr,biovil} \vindr\mimic & 13.40\std{1.43} & 4.48\std{0.61} & 8.12\std{0.69}  \\ 
        Conditional DETR-DC5-R50\cite{conditional-detr} \vindr & 13.35\std{0.82} & 3.93\std{0.57} & 13.50\std{1.09}  \\ 
        Conditional DETR-DC5-BioVIL\cite{conditional-detr,biovil} \vindr\mimic & 14.86\std{1.33} & 4.97\std{0.67} & 11.99\std{0.98} \\ 
        Deformable DETR-R50\cite{zhu2020deformable} \vindr & 11.49\std{1.13} & 2.94\std{0.50} & 10.65\std{0.88} \\ 
        Deformable DETR-BioVIL\cite{zhu2020deformable,biovil} \vindr\mimic & 12.27\std{0.94} & 4.89\std{0.63} & 10.26\std{0.82} \\ 
        $\mathcal{C}$o-Deformable-DETR-R50\cite{codetr} \vindr & 16.39\std{1.13} & 4.26\std{0.63} & 14.86\std{1.20} \\
        $\mathcal{C}$o-Deformable-DETR-BioVIL\cite{codetr,biovil} \vindr\mimic & 15.97\std{1.00} & 5.38\std{0.75} & \textbf{15.83\std{1.42}}\\
        \midrule
        \multicolumn{4}{l}{\textbf{Weakly-Supervised Object Detection}} \\
        CheXnet\cite{chexnet} \nih & 1.83\std{0.25} & 7.12\std{0.74} & 7.47\std{0.60} \\
        CheXnet\cite{chexnet} (BioVIL\cite{biovil}) \nih\mimic & 1.62\std{0.17} & 4.93\std{0.41} & 4.66\std{0.51} \\
        CheXnet\cite{chexnet} (CheXzero\cite{chexzero}) \nih\mimic & 0.30\std{0.04} & 0.68\std{0.18} & 1.40\std{0.53} \\
        MIL-ADPD\cite{adpd} \mimic & 4.30\std{0.47} & 8.14\std{0.77} & 7.21\std{0.34}  \\
        MIL-ADPD\cite{adpd} (BioVIL\cite{biovil}) \mimic & 2.19\std{0.27} & 3.29\std{0.34} & 2.20\std{0.34} \\
         Loc-ADPD\cite{adpd} \mimic & 6.69\std{0.42} & \textbf{11.33\std{0.90}} & 14.71\std{0.82} \\
         Loc-ADPD\cite{adpd} (BioVil\cite{biovil}) \mimic & 7.44\std{0.41} & \textbf{11.89\std{0.88}} & \textbf{16.56\std{1.06}} \\
        \midrule
        \multicolumn{4}{l}{\textbf{Zero-Shot Contrastive}} \\
        BioVIL\cite{biovil} \mimic & 2.82\std{0.25} & 2.63\std{0.26} & 7.15\std{0.52} \\
        CheXzero\cite{chexzero} \mimic & 0.54\std{0.06} & 2.39\std{0.24} & 2.81\std{0.25} \\
        \bottomrule
\end{tabular}
\end{center}

In \emph{pathology detection (OD)} (\cref{tab:restuls_loccls}), 
ChEX is competitive on 2 of 3 tasks, namely NIH8 and MS-CXR. On VinDr-CXR, where bounding box supervision is available, the SupOD model Faster R-CNN \cite{frcnn}, trained on this dataset, performs best, outperforming ChEX by $29\%$. 
$\mathcal{C}$o-DETR\cite{codetr} performs slightly worse but still outperforms ChEX, while other DETR-style models are on-par or inferior to ChEX (ChEX outperforms DETR-DC5\cite{detr} by $18\%$).
ChEX doubles the performance of the best WSupOD model Loc-ADPD\cite{adpd} (with BioVIL backbone). 
On NIH8, all SupOD models perform very poor (ChEX almost doubles their best performance),
while the WSupOD model Loc-ADPD is competitive with ChEX.
On MS-CXR, the best SupOD baseline $\mathcal{C}$o-DETR (with BioVIL backbone), trained on VinDr-CXR, and the best WSupOD model Loc-ADPD are both within 1-std of ChEX. 

\FloatBarrier
\newpage

\subsection{Region Classification (RC) Tasks}
\begin{table}[h!]
\caption{Region Classification (RC) results on the MS-CXR and CIG datasets. We indicate the use of training or evaluation images from VinDr-CXR by \vindr, from NIH8 by \nih, and from MIMIC-CXR by \mimic. All results are based on our own experiments and we provide the std based on bootstrapping with $N=250$. Best results and those within 1-std are marked in bold.}
\label{tab:results_rc}
\vspace{\floatsep}
\begin{center}
\centering\scriptsize
    \begin{tabular}{lcc}
        \toprule
           & \multicolumn{2}{c}{Region Classification (RC)} \\
         \cmidrule(lr){2-3} 
         & \multicolumn{1}{c}{MS-CXR\mimic} & Chest ImaGenome\mimic  \\
         &  [AUROC] & [wAUROC] \\
         \midrule
         ChEX \vindr\mimic & \textbf{82.33\std{2.80}} & \textbf{70.46\std{0.36}} \\
         \midrule
         \multicolumn{3}{l}{\textbf{Supervised Image Classifiers (ROI-pooled)}} \\
         CheXnet\cite{chexnet} \nih & 61.46\std{3.41} & 59.03\std{0.21} \\
         CheXnet\cite{chexnet} (BioVIL\cite{biovil}) \nih\mimic & 55.19\std{3.29} & 58.12\std{0.21} \\
         CheXnet\cite{chexnet} (CheXzero\cite{chexzero}) \nih\mimic & 57.54\std{2.91} & 60.02\std{0.21} \\
         \midrule
         \multicolumn{3}{l}{\textbf{ROI-Pool Object Detectors (Classification Heads)}} \\
         Faster R-CNN\cite{frcnn} \vindr & 75.66\std{2.64} & 58.28\std{0.22} \\
         Faster R-CNN\cite{frcnn} (BioVIL\cite{biovil}) \vindr\mimic & 76.13\std{2.55} & 56.64\std{0.19} \\ 
         \midrule
         \multicolumn{3}{l}{\textbf{Zeroshot Contrastive Models}} \\
         BioVIL\cite{biovil} ROI Pool \mimic & 67.41\std{2.80} & 66.96\std{0.32} \\
         BioVIL\cite{biovil} Cropped \mimic & 59.56\std{3.09} & 60.41\std{0.26} \\
         ChexZero\cite{chexzero} ROI Pool \mimic & 59.97\std{1.14} & 65.73\std{0.35} \\
         ChexZero\cite{chexzero} Cropped \mimic & 60.53\std{3.45} & 61.80\std{0.28}  \\
      \bottomrule
    \end{tabular}
\end{center}
\end{table}

On \emph{region classification (RC)} (\cref{tab:results_rc}) tasks, ChEX clearly outperforms all baselines. The best baseline on MS-CXR is the Faster R-CNN with BioVIL backbone, trained on VinDr-CXR, and is outperformed by $8\%$, while on CIG the best baseline is the contrastive model BioVIL, which is outperformed by $5 \%$.

\subsection{Region Explanation (RE) Tasks}

\emph{Region explanation (RE)} can only be performed by a single generative model, namely RGRG\cite{rgrg}. Other report generation models can only predict the full report but cannot provide descriptions for specified regions. 
On MS-CXR (\cref{tab:results_re_mscxr}), our model ChEX performs similar to RGRG on Mic-F1-14 and METEOR but improves by $25\%$ on Mac-F1-14. 
On CIG (\cref{tab:results_re_cig}), ChEX outperforms RGRG by $18\%$ on Mic-F1-14, by $40\%$ on Mac-F1-14, and by $29\%$ on METEOR.
This result is particularly noteworthy considering RGRG's explicit training for this task 
and highlights the importance of the contrastive losses in ChEX. 
Contrastive baselines exhibit poor performance on these generative tasks, as they rely on sentence retrieval based on region similarity.

\captionof{table}{Region explanation (RE) results on the MS-CXR dataset. We indicate the use of training or evaluation images from VinDr-CXR by \vindr and from MIMIC-CXR by \mimic. All results are based on our own experiments and we provide the std based on bootstrapping with $N=10$. Best results and those within 1-std are marked in bold.}
\label{tab:results_re_mscxr}
\begin{center}
\centering\scriptsize
\begin{tabular}{lrrrrr}
    \toprule
      & \multicolumn{5}{c}{Region Explanation (RE)}  \\
     \cmidrule(lr){2-6} 
     & \multicolumn{5}{c}{MS-CXR\mimic}  \\
     &  [Mic-F1-14] & [Mac-F1-14] & [Mic-F1-5+] & [Mac-F1-5+] & [METEOR] \\
     \midrule
     ChEX \vindr\mimic & \textbf{49.97\std{2.24}} & \textbf{20.50\std{1.54}} & \textbf{62.54\std{1.50}} & \textbf{44.95\std{2.23}} & \textbf{8.79\std{0.54}} \\
     \midrule
     \multicolumn{6}{l}{\textbf{Generative Models}} \\
     RGRG\cite{rgrg} \mimic & \textbf{48.97\std{2.50}} & 16.37\std{2.00} & 60.39\std{2.21} & 38.34\std{2.23} & 8.15\std{0.78}  \\
     \midrule
     \multicolumn{6}{l}{\textbf{Zeroshot Contrastive Models (Nearest Neighbor, ROI pooled)}} \\
     BioVIL\cite{biovil} \mimic & 5.86\std{1.41} & 3.69\std{0.67} & 12.79\std{2.25} & 10.95\std{1.93} & 3.82\std{0.03} \\
     CheXzero\cite{chexzero} \mimic & 5.41\std{1.30} & 3.40\std{0.78} & 11.00\std{2.25} & 9.02\std{1.83} & 3.47\std{0.45}\\
  \bottomrule
\end{tabular}
\end{center}

\captionof{table}{Region explanation (RE) results on the CIG dataset. We indicate the use of training or evaluation images from VinDr-CXR by \vindr and from MIMIC-CXR by \mimic. All results are based on our own experiments and we provide the std based on bootstrapping with $N=10$. Best results and those within 1-std are marked in bold.}
\label{tab:results_re_cig}
\begin{center}
\scriptsize
\begin{tabular}{lrrrrr}
    \toprule
      & \multicolumn{5}{c}{Region Explanation (RE)}  \\
     \cmidrule(lr){2-6} & \multicolumn{5}{c}{Chest ImaGenome\mimic}  \\
     &  [Mic-F1-14] & [Mac-F1-14] & [Mic-F1-5+] & [Mac-F1-5+] & [METEOR] \\
     \midrule
     ChEX \vindr\mimic & \textbf{53.34\std{0.43}} & \textbf{29.13\std{0.35}} & \textbf{44.53\std{0.40}} & \textbf{39.19\std{0.37}} & \textbf{10.18\std{0.13}} \\
     \midrule
     \multicolumn{6}{l}{\textbf{Generative Models}} \\
     RGRG\cite{rgrg} \mimic & 45.26\std{0.44} & 20.88\std{0.19} & 31.93\std{0.31} & 27.86\std{0.32} & 7.88\std{0.10}\textsuperscript{\textdagger} \\
     \midrule
     \multicolumn{6}{l}{\textbf{Zeroshot Contrastive Models (Nearest Neighbor, ROI pooled)}} \\
     BioVIL\cite{biovil} \mimic & 24.40\std{0.38} & 7.15\std{0.10} & 9.42\std{0.22} & 7.45\std{0.16} & 3.82\std{0.03} \\
     CheXzero\cite{chexzero} \mimic & 21.86\std{0.35} & 8.93\std{0.15} & 7.78\std{0.21} & 6.99\std{0.17} & 3.68\std{0.04} \\
  \bottomrule
\end{tabular}\\
\textsuperscript{\textdagger}Results differ to the RGRG paper \cite{rgrg}, because RGRG uses a different test split and computes scores only for regions that where selected for prediction by the model.

\end{center}
\FloatBarrier

\subsection{Full Report Generation (RG) Task}
\begin{table}[ht]
    \centering
    \caption{Clinical efficacy (CE) results on the full report generation (RG) task on MIMIC-CXR. We indicate the use of training or evaluation images from VinDr-CXR by \vindr, from MIMIC-CXR by \mimic, and from other sources by \other. Except for ChEX, results are taken from the original papers if not marked otherwise. For ChEX, we provide the std based on bootstrapping with $N=10$. The best results are marked in bold. We note that test splits, pre-, and post-processing can differ between models, leading to limitations in the comparison of exact results as also acknowledged by \cite{maira,rgrg}. }
    \label{tab:results_rg_ce}
\centering\scriptsize
\setlength{\tabcolsep}{0.5pt}
\begin{tabular}{lccccc}
    \toprule
    & \multicolumn{5}{c}{Full Report Generation (RG)} \\
    \cmidrule(lr){2-6} & \multicolumn{5}{c}{MIMIC-CXR\mimic} \\
    & [Mic-F1-14] & [Ex-F1-14] & [Mac-F1-14] & [Mic-F1-5+] & [Mac-F1-5+] \\
    \midrule
    ChEX \vindr\mimic & 52.32\std{0.51} & \textbf{58.76\std{0.42}} & 32.56\std{0.51}  & \textbf{61.03\std{0.56}} & \textbf{55.85\std{0.57}} \\
    \midrule
    RGRG\cite{rgrg} \mimic & 49.8 & 44.7 & $31.8^*$ & 54.7 &   46.44 \\ 
    CvT-212DistilGPT2\cite{nicolson_2023} \mimic & 44.2 & 38.4 & 30.7 &  -- \\
    M2 Trans w/NLL+BS+fCEN\cite{miura2021improving} \mimic & -- & -- & -- & 56.7 &  --   \\
    METransformer \cite{wang_metrans_2023} \mimic & -- & 31.1 & -- &  -- & -- \\ 
    Med-PaLM M (12B) \cite{medpalmm} \mimic\other & 51.41 & -- & 37.31 & 56.54\textsuperscript{\textdagger} & 50.57\textsuperscript{\textdagger} \\
    Med-PaLM M (84B) \cite{medpalmm} \mimic\other  & 53.56 & -- & \textbf{39.83} & 57.88\textsuperscript{\textdagger} &  51.60\textsuperscript{\textdagger} \\
    Med-PaLM M (562B) \cite{medpalmm} \mimic\other  & 51.60 & -- & 37.81 & 56.28\textsuperscript{\textdagger} &  49.86\textsuperscript{\textdagger}
    \\
    MAIRA-1\cite{maira} \mimic & \textbf{55.7} & -- & 38.6 & 58.8 & 51.7 \\
    Prompt-MRG\cite{promptmrg} \mimic  & -- & 47.6 & 38.1 &  -- & --\\ 
    OmniFM-DR\cite{omnifm} \mimic\other & -- & 33.3 & -- & -- \\
    RaDialog-INS\cite{radialog} \mimic\other & -- & -- & 38.6 & -- & --  \\ 
    RaDialog-RG\cite{radialog} \mimic\other & -- & -- & 39.4 & -- & -- \\
    COOMG\cite{Gu_2024_comg} &  -- & 34.5 & -- & -- & -- \\
    ORGAN\cite{hou_2023_organ} & -- & 38.5 & -- & -- & -- \\
  \bottomrule
\end{tabular}\\
\textsuperscript{\textdagger} Limited comparability as F1-5 was reported instead of F1-5+, i.e.\ uncertain cases were treated as positive instead of negative. \\
* Reported number taken from the Prompt-MRG\cite{promptmrg} paper.
\end{table}

For \emph{full report generation (RG)} on MIMIC-CXR (\cref{tab:results_rg_ce,tab:results_rg_nlg}), ChEX sets a new state-of-the-art on the Ex-F1-14, Mic-F1-5+, and Mac-F1-5+ metrics. Considering the commonly used metric Mic-F1-14, the SOTA model MAIRA-1\cite{maira} -- having 7B parameters -- outperforms ChEX -- having only 1B parameters -- by only $6\%$. More notably, ChEX outperforms the more than ten times larger 12B parameter Med-PaLM M by $2\%$ and is only outperformed by $2\%$ by the 80 times larger 84B parameter version. Limitations of ChEX can be observed on the Mac-F1-14 metric, where ChEX is outperformed by the SOTA model Med-PaLM M 84B by $22\%$ but also by smaller models like Prompt-MRG\cite{promptmrg} (by $17\%$) and RaDialog-RG\cite{radialog} (by $21\%$). ChEX, however, improves over RGRG by $2\%$, indicating that this is a limitation of training based on anatomical regions and is caused by the poor performance of some rarer pathologies (we refer to the supp.\@ material).
On the Ex-F1-14 metric, ChEX outperforms the best baseline Prompt-MRG by $23\%$. Most larger models do however not report this metric. On the Mic-F1-5+ and Mac-F1-5+ metrics, ChEX outperforms the large SOTA model MAIRA-1 by $4\%$ and $8\%$, respectively.  
In language-based metrics such as METEOR, ChEX shows relatively low performance ($60\%$ lower than the best and $21\%$ lower than the second-best model) due to not being explicitly trained for full report generation but generating reports based on a set of queries. 
While this approach offers advantages (see Sec.\@ 5.4 of the main paper), it may result in reports differing from reference reports. 
However, it's important to note that language-based metrics offer limited insights into the factual correctness of generated reports \cite{miura2021improving}.

In \cref{tab:results_rg_classes}, we provide 
the CE metric F1 per pathology/finding and compare the results between ChEX and the two baselines RGRG and MAIRA-1.

\FloatBarrier
\newpage

\captionof{table}{Natural language generation (NLG) results on the full report generation (RG) task on MIMIC-CXR. We indicate the use of training or evaluation images from VinDr-CXR by \vindr, from MIMIC-CXR by \mimic, and from other sources by \other. Except for ChEX, results are taken from the original papers if not marked otherwise. For ChEX, we provide the std based on bootstrapping with $N=10$. The best results are marked in bold. As in \cref{tab:results_rg_ce}, we again emphasize the limitations when comparing results.}
    \label{tab:results_rg_nlg}
\begin{center}
\centering\scriptsize
\begin{tabular}{lccccc}
    \toprule
    & \multicolumn{5}{c}{Full Report Generation (RG)} \\
    \cmidrule(lr){2-6} & \multicolumn{5}{c}{MIMIC-CXR\mimic} \\
     & [BLEU-1] & [BLEU-4] & [METEOR] & [ROUGE-L] & [CIDEr] \\
    \midrule
    ChEX \vindr\mimic & 29.58\std{0.24} & 6.52\std{0.11} & 13.26\std{0.10} & 19.49\std{0.10} & 18.49\std{0.79}\\
    \midrule
    RGRG\cite{rgrg} \mimic & 37.3 & 12.6 & 16.8 & 26.4 & \textbf{49.5}  \\
    CvT-212DistilGPT2\cite{nicolson_2023} \mimic & 39.2 & 12.4 & 15.3 & 28.5 & 36.1\\
    M2 Trans w/ NLL+BS+fCEN\cite{miura2021improving} \mimic & -- & 11.4 & -- & -- & 50.9 \\
    METransformer\cite{wang_metrans_2023} \mimic & 38.6 & 12.4 & 15.2 & 29.1 & 36.2\\
    Med-PaLM M (12B)\cite{medpalmm} \mimic\other & 30.90 & 10.43 & -- & 26.16 & 23.43 \\
    Med-PaLM M (84B)\cite{medpalmm} \mimic\other & 32.31 & 11.31 & -- & 27.29 & 26.17 \\
    Med-PaLM M (562B)\cite{medpalmm} \mimic\other & 31.73 & 11.50 & -- & 27.49 & 25.27  \\
    MAIRA-1\cite{maira} \mimic & 39.2 & \textbf{14.2} & \textbf{33.3} & 28.9 & -- \\
    Prompt-MRG\cite{promptmrg} \mimic & 39.8 & 11.2 & 15.7 & 26.8 & -- \\ 
    OmniFM-DR\cite{omnifm} \mimic\other & -- & 11.0 & 14.0 & 26.5 & -- \\ 
    RaDialog-INS\cite{radialog} \mimic\other & 34.0 & 9.7 & 13.6 & 27.0 & -- \\
    RaDialog-RG\cite{radialog} \mimic\other & 34.6 & 9.5 & 14.0 & 27.1 & -- \\
    ITA\cite{wang2022inclusive} \mimic & 39.5 & 12.1 & 14.7 & 28.4 & -- \\
    COOMG\cite{Gu_2024_comg} \mimic & 34.6 &  10.4 & 13.7 & 27.9 & -- \\
    COOMG-RL \cite{Gu_2024_comg} \mimic & 36.3 & 12.4 & 12.8 & 29.0 & -- \\
    ORGAN\cite{hou_2023_organ} & \textbf{40.7} & 12.3 & 16.2 & 29.3 & -- \\
    BioViL-T\cite{biovil_t} \mimic & -- & 9.2 & -- & \textbf{29.6} & -- \\
  \bottomrule
\end{tabular}
\end{center}

\captionof{table}{
CE results (F1) per pathology/finding for ChEX, RGRG, and MAIRA-1 on the full report full report generation (RG) task on MIMIC-CXR. Except for ChEX, results are taken from the original papers. Our method ChEX performs best on 5 of the 14 classes, even outperforming the much larger MAIRA-1. On 9 of the 14 classes, it outperforms RGRG, where it is able to describe findings and pathologies like lung lesions or enlarged cardiomediastinum, which RGRG is unable to detect at all.}
\label{tab:results_rg_classes}
\begin{center}
\centering
\scriptsize
\begin{tabular}{lrrr}
\toprule
& \multicolumn{3}{c}{F1} \\
\cmidrule(lr){2-4}  Pathology/Finding & ChEX & RGRG\cite{rgrg} & MAIRA-1\cite{maira} \\ 
\midrule
Cardiomegaly & \textbf{66.62\std{1.08}} & 62.4 & 64.0 \\ 
Edema & \textbf{54.81\std{1.48}} & 51.4 & 44.0\\ 
Consolidation & 15.27\std{2.10} & 7.8 & \textbf{20.0} \\ 
Atelectasis & 45.99\std{1.43} & \textbf{54.6} & 41.3\\   
Pleural Effusion & \textbf{69.11\std{1.04}} & 56.0 & \textbf{68.9} \\ 
\midrule
Enlarged Cardiomediastinum & 6.81\std{1.74} & 0.3 & \textbf{11.9} \\ 
Fracture & 0.00\std{0.00} & 0.0 & \textbf{24.9} \\ 
Lung Lesion & 14.24\std{3.05} & 0.7 & \textbf{18.8} \\ 
Lung Opacity & \textbf{52.32\std{1.06}} & 26.8 & 49.8 \\ 
Pleural Other & 0.00\std{0.00} & 0.2 & \textbf{14.7} \\ 
Pneumonia & \textbf{19.30\std{1.55}} & 16.2 & \textbf{18.3} \\ 
Pneumothorax & 17.59\std{3.21} & 15.9 & \textbf{40.8} \\ 
Support Devices & 69.40\std{0.92} & 70.9 & \textbf{84.5} \\  
No Finding & 24.36\std{2.97} & \textbf{63.2} & 38.6\\ 
\bottomrule
\end{tabular}
\end{center}

\newpage
\section{Model Implementation Details}
\subsection{Model Architecture and Components}

\subsubsection{Prompt Detector}
\begin{figure}[ht!]
    \centering
    \includegraphics[width=.9\textwidth]{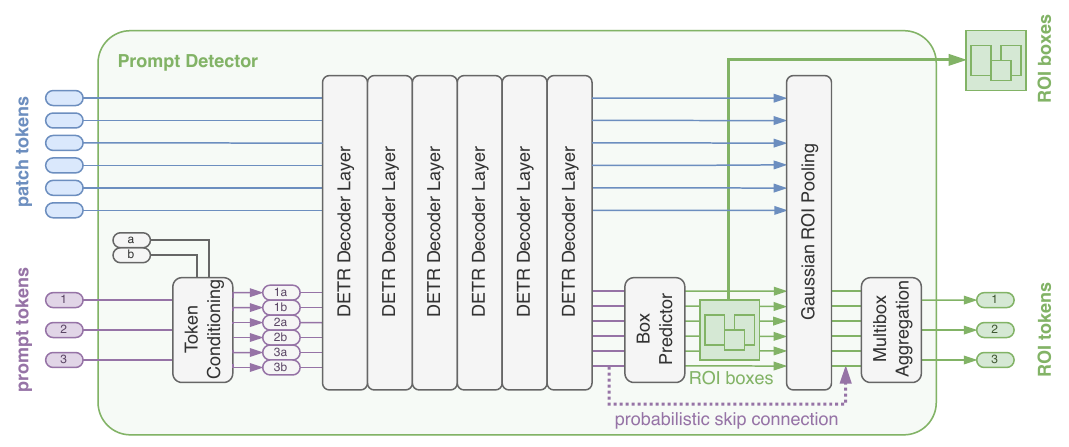}
    \caption{Prompt detector}
    \label{fig:detector}
\end{figure}
\cref{fig:detector} provides an overview of the prompt detector. 
Given is an input of $Q$ prompt tokens $\bm{q}_k$ with $k \in \{1, \dots, Q\}$. In order to enable the prediction of up to $M = 3$ bounding boxes per prompt token $\bm{q}_k$, we utilize $M$ learned, randomly initialized, box tokens $\bm{t}_m$.
We then additively condition these tokens on the $Q$ queries resulting in a total of $Q\cdot M$ query tokens $\bm{t}'_{k, m}$:
\begin{align}
    \bm{t}'_{k, m} = \bm{q}_k + \bm{t}_m \,.
\end{align}
These tokens serve as the initial queries for the DETR decoder layers, while the patch tokens from the image encoder are used for keys and values.
We use a total of 6 DETR-style transformer decoder layers. Self-attention is restricted, allowing attention only between tokens from the same query, \ie with the same $k$. The query tokens $\bm{t}'_{k, m} $ are only used as queries for the first layers, later layers use the output of the previous layer as queries. We additively combine the patch tokens with the corresponding positional encodings from the ViT before using them in the prompt detector.
The outputs of the last decoder layer are used as preliminary ROI tokens $\tilde{\bm{r}}_{k, m}$. They are then used by the MLP-based box predictor to predict the corresponding bounding boxes, where we apply sigmoid to return relative image coordinates.

Next, we use Gaussian ROI pooling following \cite{wsrpn}. For each bounding box, we define a 2D multivariate Gaussian with independent $x$
and $y$ components (\ie with zero covariance) using the box center as mean and the box size as standard deviation. The probability density function (pdf) of this Gaussian is used in a weighted aggregation of the patch tokens (which have been projected by an MLP), resulting in a single feature vector per ROI $(k, m)$.
We randomly add the preliminary ROI tokens $\tilde{\bm{r}}_{k, m}$ in $25\%$ of the cases but apply dropout to them before adding. Finally, these features are projected using an MLP with two hidden layers resulting in the ROI tokens $\bm{r}_{k, m}$. We use an additional MLP followed by sigmoid on these ROI tokens to predict box scores $s_{k, m}$. Using these scores, we now aggregate the ROI tokens for each prompt $k$ to get the final (aggregated) ROI tokens $\bm{r}_{k}$:
\begin{align}
    \bm{r}_{k} = \frac{\sum_m s_{k, m} \cdot \bm{r}_{k, m}}{\sum_m s_{k, m}} \,.
\end{align}

\FloatBarrier

\subsubsection{Sentence Generator}
\begin{figure}[ht!]
    \centering
    \includegraphics[width=.8\textwidth]{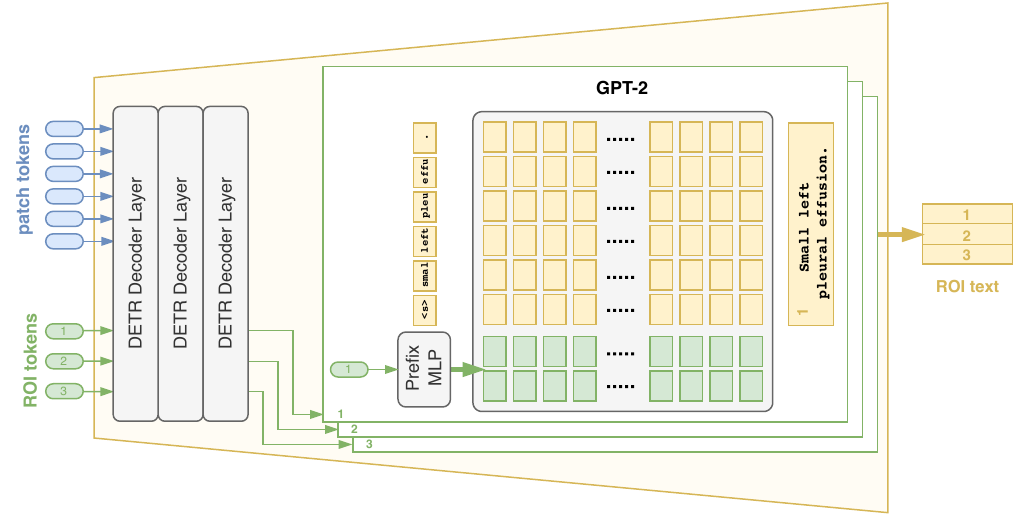}
    \caption{Sentence generator with post decoder on the left and the conditioned language model on the right}
    \label{fig:generator}
\end{figure}
\cref{fig:detector} provides an overview of the sentence generator. It consists of two major components, the post decoder and the language model (GPT-2) with conditioning.
The post decoder consists of three DETR decoder layers. 
For each ROI $k$, we use the per-bounding-box ROI tokens $\bm{r}_{k, m}$ and the aggregated ROI token $\bm{r}_{k}$ as queries, leading to a total of $M+1 = 4$ queries per ROI.
We randomly drop the per-bounding-box ROI tokens $\bm{r}_{k, m}$ with a $30\%$ probability while always keeping the aggregated tokens $\bm{r}_{k}$.
We use the patch tokens and the (projected) CLS token from the image encoder as keys and values. Self-attention is restricted, allowing attention only between queries for the same ROI $k$. After the final decoder layer, we only keep the output features associated with the aggregated ROI token of each ROI $k$, discarding the features of the per-bounding-box ROI tokens. 
The final output of the post decoder is computed by summing the outputs of the final decoder layer with the input features, introducing a skip connection. Before summing, we multiply the output features with a learned gating factor, which is initialized by zero. 

For text generation, we condition the GPT-2 medium\cite{radford2019language} language model on the output features of the post decoder. The descriptions of each ROI $k$ are predicted individually, treating each $k$ as an independent sample.
We follow a similar approach as P-tuning v2 \cite{p-tuning} but do not freeze any parameters of the language model. Specifically, we use an MLP to project the feature vector for each $k$ into a prefix for the language model. 
We use a prefix of five tokens and compute key and value features for each language model layer. During generation, we use a maximum length of 128 generated tokens.

\subsubsection{Futher Implementation Details}
For all components of the prompt detector and post decoder of the sentence generator, if not otherwise specified, we use the model dimension $512$ with hidden dimension $2048$ for MLPs, dropout $0.3$, attention dropout $0.1$, drop-path\cite{droppath} probability $0.2$, and GELU\cite{gelu} non-linearities. For attention layers we apply the layer norm before each operation and use layer scaling with initialization $0.1$.

\subsubsection{Number of Parameters}
ChEX has an overall 1051M parameters of which we keep 123M frozen throughout our whole training. The ViT image encoder contains 88M, the text encoder 63M, the prompt detector 28M, and the sentence generator 872M parameters. 

\subsection{Supervision and Loss Functions}
\subsubsection{Pathology Tokens}
\begin{wrapfigure}[14]{r}{.45\textwidth}
\centering
    \vspace{-2em}
    \includegraphics[trim=0 0 40 0, clip, width=.45\textwidth]{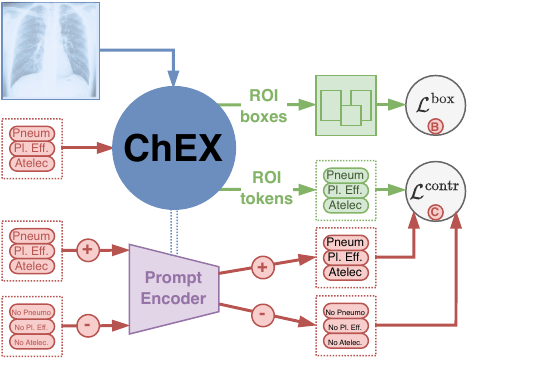}
    \caption{Training with pathology tokens}
    \label{fig:train_patho}
\end{wrapfigure}

\cref{fig:train_patho} visualizes training with pathology tokens.
Given the set of pathology classes $\mathcal{C}$ from VinDr-CXR (22 classes with bounding boxes), for each pathology class $c \in \mathcal{C}$, we encode its positive prompt (\eg, \say{Pleural effusion}) and its negative prompt (\eg, \say{No pleural effusion}) using the prompt encoder. We denote the resulting prompt tokens as $\bm{q}_c^+$ and $\bm{q}_c^-$, respectively.
Next, we pass the positive tokens $\bm{q}_c^+$ to the prompt decoder, which predicts three ROI bounding boxes $\hat{b}_{c, m}$ with box scores $s_{c, m} \in [0, 1]$ (with $m \in \{1, 2, 3\}$) and the aggregated ROI token $\bm{r}_c$.

\paragraph{Pathology Detection (Bounding Box Loss)}
For the bounding boxes, we apply the modified DETR loss. The class match probability $p_{c, m}$ of a bounding box is computed as
\begin{align}
    p_{c, m} = \frac{\exp\left(\bm{r}_{c, m} \bullet \bm{q}_c^+\right)}{\exp\left(\bm{r}_{c, m} \bullet \bm{q}_c^+\right) + \exp\left(\bm{r}_{c, m} \bullet \bm{q}_c^-\right)} \,,
\end{align}
where $\bullet$ denotes the dot product and $\bm{r}_{c, m}$ denotes the (non-aggregated) ROI token for each bounding box. Using the standard formulation of the box loss 
\begin{align}
\mathcal{L}_\text{box}\left(\hat{b}, b\right) = 5 \cdot \text{L}_1\left(\hat{b}, b\right) + 2 \cdot \text{gIoU}\left(\hat{b}, b\right) \,,
\end{align}
we compute the matching cost for class $c$, predicted box $m$, and target box $m'$ based on the box cost, class match probability of the box, and box score as
\begin{align}
    \mathcal{L}_\text{match}(c, m, m') = \mathcal{L}_\text{box}\left(\hat{b}_{c, m}, b_{c, m'}\right) + p_{c, m} + 3 \cdot s_{c, m} \,.
\end{align}
For each positive class $c$, \ie each class that has target bounding boxes, we iteratively apply Hungarian matching using the matching cost $\mathcal{L}_\text{match}$, where matching is applied individually for each $c$. We start with an initial matching. If non-matched predicted boxes remain for that class, we apply matching again with the remaining predicted boxes and all target boxes until all predicted boxes are matched, allowing a target box to be matched with several predicted boxes. We apply the box loss $\mathcal{L}_\text{box}$ to all pairs of matched predicted/target boxes. Additionally, we apply the focal loss $\mathcal{L}_\text{focal}$ to the box scores $s_{c, m}$ with positive targets for boxes matched in the first iteration and negative targets for other boxes and all negative classes. Note that we compute the $\alpha$ parameter of the focal loss based on the frequency of positive targets per class in the current batch. The final pathology detection loss is computed as $\mathcal{L}_\text{patho-detect} = \mathcal{L}_\text{box} + 3 \cdot \mathcal{L}_\text{focal}$.

\paragraph{Pathology Classification (Contrastive Loss)} We additionally apply a contrastive loss on the aggregated ROI token $\bm{r}_c$. Here, we consider the pathology targets $y_c \in \{0, 1\}$, which define whether the pathology $c$ is present in the image. The ROI token for class $c$ is paired with its positive or negative pathology prompt $\bm{q}_c^+$ or $\bm{q}_c^-$ depending on $y_c$. As negative pairs, we include the opposing prompt alongside all positive and negative prompts of all other pathologies. This leads to the following loss function:
\begin{align}
    \mathcal{L}_\text{patho-cls} = - \frac{1}{\lvert\mathcal{C}\rvert}\sum_c \log \frac{y_c \cdot \exp(\cos(\bm{r}_c,\bm{q}_c^+) / \tau) + (1 - y_c) \cdot \exp(\cos(\bm{r}_c,\bm{q}_c^-) / \tau)}{\sum_{c'} \exp(\cos(\bm{r}_c,\bm{q}_{c'}^+) / \tau) + \exp(\cos(\bm{r}_c,\bm{q}_{c'}^-) / \tau)} \,,
\end{align}
where we set the temperature $\tau$ to $0.2$.

\subsubsection{Anatomy Tokens}
\begin{wrapfigure}[14]{r}{.45\textwidth}
\centering
    \vspace{-2em}
    \includegraphics[trim=0 0 160 0, clip, width=.45\textwidth]{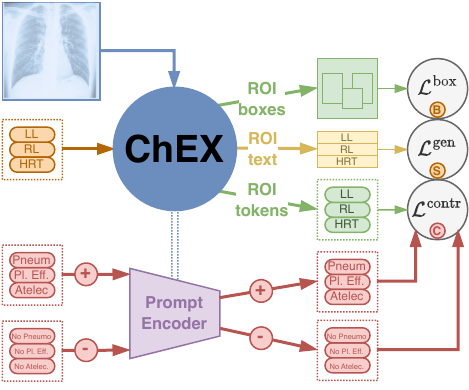}
    \caption{Training with anatomy tokens}
    \label{fig:train_anat}
\end{wrapfigure}

\cref{fig:train_anat} visualizes training with anatomy tokens.
Given is the set of anatomical regions $\mathcal{A}$ from CIG (29 regions from CIG and additionally 5 anatomical regions each consisting of 2 regions from CIG, \eg lungs consisting of the right and left lungs).
For each region $a \in \mathcal{A}$, we encode its prompt using the prompt encoder and denote 
the resulting prompt tokens as $\bm{q}_a$. We pass these tokens to the prompt decoder, which predicts three ROI bounding boxes $\hat{b}_{a, m}$ and the aggregated ROI token $\bm{r}_a$. Additionally, we encode positive and negative prompts for pathologies assigned to regions (we use 53 pathologies and findings from CIG) and denote the resulting prompt tokens as $\bm{q}_c^+$ and $\bm{q}_c^-$, respectively.

\paragraph{Anatomy Detection (Bounding Box Loss)}
For anatomical regions with a single target bounding box, we match all three predicted boxes for that region to the target bounding box. For regions with multiple target boxes, we apply the same matching strategy as used for pathology bounding boxes but using only $\mathcal{L}_\text{box}$ as the matching cost. Finally, we apply the box loss $\mathcal{L}_\text{box}$ to all matched bounding boxes. Box scores are not trained for anatomy tokens.

\paragraph{Anatomy Classification (Contrastive Loss)} We again apply a contrastive loss to the ROI tokens. Here we use the positive and negative pathology tokens $\bm{q}_c^+$ and $\bm{q}_c^-$ and align the ROI token $\bm{r}_a$ of anatomy $a$ based on whether the pathologies are present in this region. The presence of pathology $c$ in anatomy $a$ is given as $y_{a, c}$ and is provided by CIG. This leads to the following loss function:
\vspace{3em}
\begin{align}
\begin{split}
    &\mathcal{L}_\text{anat-cls} = \\ &- \frac{1}{\lvert\mathcal{A}\rvert}\sum_a\log
    \frac{\exp\left\{\frac{1}{\lvert\mathcal{C}\rvert}\sum_c y_{a, c} \cdot \cos(\bm{r}_a,\bm{q}_c^+) / \tau) + (1 - y_{a, c}) \cdot \cos(\bm{r}_a,\bm{q}_c^-) / \tau)\right\}}
    {\sum_{c'} \exp(\cos(\bm{r}_a,\bm{q}_{c'}^+) / \tau) + \exp(\cos(\bm{r}_a,\bm{q}_{c'}^-) / \tau)}
\end{split}
\end{align}
where we set the temperature $\tau$ to $0.25$ and randomly subsample a maximum of 10 negative classes (\ie classes not present in any of the regions in the current batch). 

\paragraph{Anatomy Sentence Generation}
For sentence generation, we condition the sentence generator on the ROI tokens $\bm{r}_a$ and train it to autoregressively predict the target sentences assigned to this region using the standard cross-entropy loss $\mathcal{L}_\text{anat-gen}$. Note that we concatenate sentences if multiple sentences are assigned to a region and ignore regions without sentences. Additionally, the MSE loss $\mathcal{L}_\text{anat-mse}$ is applied between the output of the post decoder and the anatomy sentences encoded by the prompt encoder.

\subsubsection{Sentence Tokens}
\begin{wrapfigure}[10]{R}{.45\textwidth}
\centering
    \vspace{-2em}
    \includegraphics[trim=0 0 40 0, clip, width=.45\textwidth]{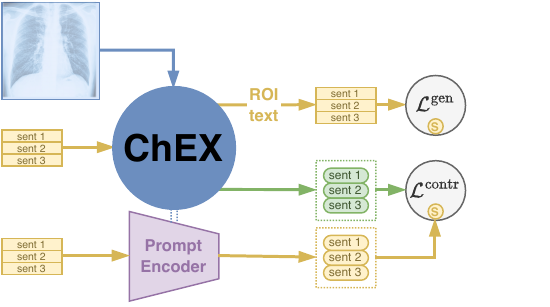}
    \caption{Training with sentence tokens}
    \label{fig:train_sent}
\end{wrapfigure}

\cref{fig:train_sent} visualizes training with sentence tokens.
Given are the sentences $\mathcal{S}^{(i)}$ of sample $i$ from MIMIC-CXR.
We encode each sentence $s$ using the prompt encoder and denote 
the resulting prompt tokens as $\bm{q}^{(i)}_s$. We pass each of these tokens to the prompt decoder, which predicts the aggregated ROI token $\bm{r}^{(i)}_s$ for each of them. 
\paragraph{Sentence Contrastive Loss}
We again apply a contrastive loss between the ROI tokens $\bm{r}^{(i)}_s$ and the prompt tokens $\bm{q}^{(i)}_s$. Unlike for pathology and anatomy tokens, where the contrastive loss was applied for each sample and token independently, we utilize sentence prompt tokens from the same and other samples in the batch as negatives. This leads to the following loss function:
\begin{align}
    \mathcal{L}_\text{sent-contr} = 
    - \frac{1}{N}\sum_i
    \frac{1}{\lvert\mathcal{S}^{(i)}\rvert}\sum_{s \in \mathcal{S}^{(i)}}
    \log 
    \frac{\exp\left(\cos\left(\bm{r}^{(i)}_s, \bm{q}^{(i)}_s\right) / \tau\right)}
    {\sum_{j,s'} \exp\left(\cos\left(\bm{r}^{(i)}_s,\bm{q}^{(j)}_{s'}\right) / \tau\right)} \,,
\end{align}
where $N$ is the batch size and we set the temperature $\tau$ to $0.25$.

\paragraph{Sentence Generation}
For sentences generation, \ie sentence reconstruction, we condition the sentence generator on the ROI tokens $\bm{r}_s$ and train it to autoregressively reconstruct the report sentences used as prompts, using the standard cross-entropy loss $\mathcal{L}_\text{sent-gen}$. Additionally, the MSE loss $\mathcal{L}_\text{sent-mse}$ is applied between the output of the post decoder and the encoded sentence prompts $\bm{q}_s$.

\paragraph{Global Contrastive Loss}
To fine-tune and stabilize the image encoder, we additionally apply standard CLIP loss $\mathcal{L}_\text{global-contr}$ between the (global) average pooled projected patch features and the average pooled sentences tokens $\bm{q}^{(i)}_s$. We use $0.2$ as the temperature.

\subsection{Multistage Training}
We train ChEX in three different stages where we add new components with each stage while freezing some already trained components. 
Additionally, we prepare the language model for MIMIC-CXR reports in a text-only pre-training stage.

\subsubsection{Stage 1}
The goal of stage 1 is to train the prompt detector and fine-tune the image encoder. We initialize the image and text encoder as well as their projection heads from the CheXzero\cite{chexzero} weights of one of their checkpoints \footnote{\texttt{best\_64\_5e-05\_original\_22000\_0.864.pt}} which we retrieved from their provided cloud drive\footnote{\url{https://drive.google.com/drive/folders/1makFLiEMbSleYltaRxw81aBhEDMpVwno}}. The text encoder and its projection head are completely frozen throughout all training stages, and we freeze only the embedding layer and the first eight transformer layers of the image encoder for stage 1. 
The prompt detector is initialized randomly and will be trained in this stage. The sentence decoder is not yet used in this stage.
We train using both datasets (MIMIC-CXR and VinDr-CXR) and all types of tokens -- pathology tokens for samples from VinDr-CXR, anatomy tokens for the CIG targets for MIMIC-CXR samples, and sentence tokens for MIMIC-CXR samples. We use all bounding box and contrastive losses but do not yet train text generation. Overall, the loss function for stage 1 is defined as
\begin{align}
\begin{split}
    \mathcal{L}_\text{stage1} = & 10 \cdot \mathcal{L}_\text{patho-detect} + 1 \cdot \mathcal{L}_\text{patho-cls} \\ & \quad + 0.1 \cdot \mathcal{L}_\text{anat-detect} + 0.005 \cdot \mathcal{L}_\text{anat-cls} \\ & \quad + 0.005 \cdot \mathcal{L}_\text{sent-contr} + 1 \cdot \mathcal{L}_\text{global-contr}
\end{split}
\end{align}

Note that for anatomy tokens we only use the textual prompts in this stage, bounding box-based queries are not yet used.
To improve generalization and limit the GPU memory consumption, we randomly select a subset from the possible tokens (\ie queries for pathologies and anatomical regions) to be used in the current batch.
We subsample 10 pathologies for bounding box training, 10 pathologies for the contrastive loss, and 20 anatomical regions.  

We train with a batch size of 256 for 50 epochs on a single Nvidia A40, which takes around 36 hours. We train with AdamW\cite{adamw} and use a cosine annealing schedule with a linear warmup of 1000 steps, initial learning rate \SI{1e-4}, and minimum learning rate \SI{1e-7}. Additionally, we use a weight decay of \SI{1e-5}, clip gradients with a norm larger than \SI{1.0}.
During training, we use the transformations from the Albumentations library \cite{albumentations}. We randomly crop and resize all images to a resolution of $224\times224$ using a crop scale range of $[0.5, 1.0]$. We then apply random horizontal flips with $50\%$ probability, random affine transformations with rotation angle range $[-10, 10]$, translation range $[-0.02, 0.02]$, and scaling range $[0.95, 1.0]$. We also apply random contrast/brightness jittering with contrast/brightness range $[0.6, 1.4]$ and random Gaussian blurring with $50\%$ probability and sigma range $[0.1, 5.0]$.

\subsubsection{Stage 2}
The goal of stage 2 is to train the post decoder of the sentence generator. We use the trained components from stage 1 but freeze all of them, \ie we fully freeze the prompt encoder, image encoder, and prompt detector. We now add the (randomly initialized) post decoder of the sentence generator, however, without using the language model yet. 
We train only on the MIMIC-CXR dataset (including targets from CIG) and only use sentence and anatomy tokens. The VinDr-CXR dataset and pathology tokens are not used. 
We use the contrastive losses for anatomy and sentence tokens and additionally use MSE losses to reconstruct the sentence embeddings of target sentences based on ROI tokens from either anatomy or sentence prompts. We do not train bounding boxes or sentence generation.
Overall, the loss function for stage 2 is defined as
\begin{align}
\begin{split}
    \mathcal{L}_\text{stage2} = & 0.01 \cdot \mathcal{L}_\text{anat-cls} + 0.04 \cdot \mathcal{L}_\text{anat-mse} \\ & \quad + 0.005 \cdot \mathcal{L}_\text{sent-contr} + 0.02 \cdot \mathcal{L}_\text{sent-mse}
\end{split}
\end{align}

For anatomical regions we randomly decide for each sample whether to use bounding-box-based or text-prompt-based queries.
We train with a batch size of 256 and initial learning rate \SI{1e-3} for 200 epochs on a single Nvidia A40, which takes around 24 hours. All other hyperparameters are kept the same as in stage 1.

\subsubsection{Text-only Pre-Training}
The goal of text-only pre-training is to prepare the decoder language model for MIMIC-CXR report sentences. 
We initialize the GPT-2 medium language model from a checkpoint\footnote{\url{https://huggingface.co/healx/gpt-2-pubmed-medium}} pre-trained on PubMed abstracts.
We only use the frozen prompt encoder together with the language model and a (randomly initialized) MLP projection for P-tuning, \ie we do not use the post decoder. Using the sentences from MIMIC-CXR, we encode each sentence, apply the MLP projection, and reconstruct the original sentence using the language model. We train the language model and MLP projection using autoregressive training. 

We use a batch size of $32$ reports (each with one or more sentences) but randomly drop sentences if there are more than 64 in a batch and use 8 gradient accumulation steps.
We train with an initial learning rate of \SI{3e-4} and 10,000 warmup steps for 50 epochs on a single Nvidia A40, which takes around 3 days.

\subsubsection{Stage 3}
The goal of stage 3 is to train the whole sentence generator. We use the pre-trained post decoder from stage 2 and the decoder language model from text-only pre-training, both of which will be trained in this stage. Additionally, we use the prompt encoder, image encoder, and prompt detector from stage 1 but keep all of them frozen.
We train only on the MIMIC-CXR dataset (including targets from CIG) and only use sentence and anatomy tokens. We autoregressive train the sentence generator to predict the target sentences of each anatomical region and to reconstruct the sentences for sentence tokens. Additionally, we use MSE losses between the output of the post decoder and the sentences encoded by the prompt encoder. 
Overall, the loss function for stage 3 is defined as
\begin{align}
\begin{split}
    \mathcal{L}_\text{stage3} = & 1 \cdot \mathcal{L}_\text{anat-gen} + 0.04 \cdot \mathcal{L}_\text{anat-mse} \\ & \quad + 0.5 \cdot \mathcal{L}_\text{sent-gen} + 0.02 \cdot \mathcal{L}_\text{sent-mse}
\end{split}
\end{align}

During training, we randomly drop the per-bounding-box ROI tokens for a query with a $10\%$ probability but always keep the aggregated ROI token. 
We use a batch size of $8$ reports (with $32$ gradient accumulation steps) but restrict the total number of generated sentences per batch to 32 for anatomical regions and 32 for sentence tokens, randomly subsampling if a batch contains more.
We train with an initial learning rate of \SI{3e-4} and 10,000 warmup steps for 20 epochs on a single Nvidia A40, which takes around 4 days.

\subsection{Inference}
\subsubsection{Sentence Grounding (SG) and Object Detection (OD)}
For SG and OD, we use textual prompts as queries (the given input sentences in SG and predefined prompts, the names of pathology classes, for OD). The encoded prompts are given to the prompt detector, which predicts bounding boxes with box scores and ROI token vectors. For SG, we directly predict these bounding boxes with their associated box scores. For OD, we compute cosine similarities between the ROI tokens and positive as well as negative pathology prompts to compute box class probabilities. We then multiplicative combine these box class probabilities with the predicted box scores to get the final box scores. Finally, we apply task-specific post-processing as described in \cref{sec:eval_details}.

\subsubsection{Region Classification (RC) and Explanation (RE)}
For RC and RE, bounding boxes are given as inputs. We use these bounding boxes as inputs for the Gaussian ROI pooling layer of the prompt detector to compute ROI tokens for each input based on the patch features. Other parts of the prompt detector and the prompt encoder are not used for this task. We then further process the ROI tokens by the post decoder of the sentence decoder to further condition them on the patch features. 

For RC, we now compute the cosine similarity between the post decoder outputs and encoded pathology prompts. Finally, we classify each region based on the cosine similarity. For the multiclass task MS-CXR, we compute the probability using softmax over all pathologies and predict the pathology class with maximum similarity. For the multilabel binary task CIG, we compute the softmax over the positive and negative prompts for each pathology class independently and predict all classes where the similarity with the positive pathology prompt is larger than the negative prompt.

For RE, we use the post decoder outputs to condition the language model and generate the descriptions for each region independently.

\subsubsection{Full Report Generation (RG)}
For RG, we use pre-defined sets of textual prompts. These prompts may include aspects like pathologies or anatomical regions as studied in \cref{sec:prompt_sets}. 
As pathology prompts, we use the set of 14 findings and pathologies from CheXpert\cite{chexpert}. As anatomy prompts we use a subset of 15 anatomical region names from the regions used in CIG, namely the lower, mid, and upper lung zones, hilar structures, apical zones, as well as the clavicle, each for left and right independently, and additional the mediastinum, the cardiac silhouette, and the abdomen.
We encode the prompts with the prompt encoder and then apply the prompt detector to predict bounding boxes and ROI tokens for each prompt. 
The ROI tokens are then used by the sentence generator to predict textual descriptions for each prompt. 

Next, we filter the descriptions to include in the final report. 
We compute the positive probability for each prompt (and the associated description) by classifying each ROI token similarly as in the region classification task using the CheXpert classes excluding no-finding. 
If any pathology is positive, we consider the prompt and associated description to be positive. We decided to keep all descriptions of pathology prompts but only positive anatomy descriptions (we refer to \cref{sec:prompt_sets}). 

Finally, we filter the descriptions based on their associated box scores. 
We use a dynamic approach where we identify similar descriptions from different prompts by computing the cosine similarity between the descriptions encoded by the prompt encoder. For descriptions with high similarity, the lower-scoring ones are excluded. By setting the similarity threshold low enough, we allow the model to not only remove synonymous descriptions but also contradicting descriptions of similar aspects or several negative findings, where we only keep the most confident and relevant ones.

\section{Evaluation Details}
\subsection{Benchmark Datasets}
\subsubsection{MIMIC-CXR}
We use the MIMIC Chest X-ray (MIMIC-CXR) Database v2.0.0\cite{mimic-cxr,mimic-cxr-2,physionet}, which consists of chest X-ray images with corresponding free-text radiology reports from 227,835 radiographic studies of 65,079 patients performed at Beth Israel Deaconess Medical Center in Boston, US.
The dataset authors de-identified the dataset satisfying the US Health Insurance Portability and Accountability Act of 1996 (HIPAA) Safe Harbor requirements and assuring Protected health information (PHI) has been removed.
For all MIMIC-CXR-based datasets, \ie MIMIC-CXR, CIG, and MS-CXR, we only use the frontal chest X-ray images from the already pre-processed MIMIC-CXR-JPG\cite{johnson2019mimicjpg} version.

We download the free-text radiology reports from the original MIMIC-CXR dataset and extract the text from the \texttt{Findings} and \texttt{Impression} sections. Reports containing none of these sections, where both of these sections are empty, or which contain less than two words are removed. We split each of these sections into sentences using Stanza\cite{stanza} and use all of these sentences (from both sections) as the report sentences of that sample. If a full report is required, as in the full report generation task, we concatenate all sentences to a single string.
We follow the official MIMIC-CXR train/val/test split. This leads to a total of 1,879 images/report pairs for validation and 3,082 for testing.

\subsubsection{Chest ImaGenome (CIG)}
For anatomical regions, including bounding boxes with associated sentences and pathologies, we utilize the Chest ImaGenome dataset (CIG)\cite{cig,cig2,physionet}. It provides scene graphs automatically constructed from the free-text radiology reports from MIMIC-CXR.
Each scene graph includes bounding boxes for 29 unique anatomical regions with associated attributes, where we consider 43 \texttt{anatomical finding} and 10 \texttt{disease} attributes as binary anatomy-level classification labels. Additionally, each region is associated with sentences from the report and we use these sentences as target sentences for anatomical regions. Note that not all sentences from the report may be associated with a region, sentences can be associated with several regions, and each region can be associated with zero, one, or multiple sentences. We only keep samples with scene graphs containing at least five valid region bounding boxes.
We use the images from the MIMIC-CXR dataset (see above) and follow the official MIMIC-CXR train/val/test split\footnote{Note that this means that we do not follow the official split from CIG.}, leading to 1,959 images for validation and 3402 for testing.
For multitask training of ChEX we combine the supervision from CIG with the sentences from MIMIC-CXR, leading to a total of 227,382 training images.

\subsubsection{MS-CXR}
The MS-CXR dataset \cite{biovil,mscxr2,physionet}, which we use only for evaluation, provides additional annotation for a subset of MIMIC-CXR. It contains 1,162 image–sentence pairs of bounding boxes and corresponding phrases for 8 pathologies annotated by board-certified radiologists. Each sentence is associated with a pathology and one or more bounding boxes. Images in MS-CXR contain at least one sentence (and pathology) but may also contain several sentences with possibly different pathologies.
We use the images from the MIMIC-CXR dataset. However, only very few images of MS-CXR are part of the MIMIC-CXR test split. For MS-CXR tasks, we thus use all MS-CXR images from the validation and test split of MIMIC-CXR for testing but exclude all images from the training split to avoid any data leakage. This leads to a total of 169 test images.

\subsubsection{VinDr-CXR}
We use the VinDr-CXr\cite{vindr,vindr2,physionet} dataset, containing 18,000 postero-anterior (PA) view, \ie frontal, chest X-ray images retrospectively collected from the Hospital 108 (H108) and the Hanoi Medical University Hospital (HMUH), two of the largest hospitals in Vietnam. The images were
manually annotated by radiologists with 22 localized labels (with bounding boxes) and 6 global (image-level classification) labels of diseases (pathologies). Each image may be associated with zero, one, or more pathologies, and for each (localized) pathology there may be one or more provided bounding boxes.
In order to protect the patient’s privacy and meet the U.S. HIPAA, the European GDPR, and the local privacy laws, all personally identifiable information associated with the images has been removed or replaced with random values\cite{vindr2}. 
The images are provided in the DICOM format, and we converted them into PNG files using the \texttt{pydicom}\footnote{\url{https://pydicom.github.io/}} library. The dataset authors provide a train/test split consisting of 15,000 and 3,000 samples, respectively. The samples in the train split contain bounding box annotations from individual radiologists that have not been merged by the dataset authors. We fuse bounding box annotations from different radiologists corresponding to the same pathology using weighted box fusion\cite{wbf} with IoU threshold $0.1$. Samples from the provided test split come with merged annotations, such that no further processing is required. We use all samples from the training split for training and randomly split the 3,000 test split samples into 1,500 samples for validation and 1,500 samples for testing.

\subsubsection{NIH ChestXray8 (NIH8)}
We use the ChestXray-8 (NIH8) dataset\cite{nih8}, which consists of
108,948 chest X-ray images retrospectively collected from the National Institutes of Health Clinical Center in the US. The authors do not provide information about data anonymization. The images are provided as pre-processed PNG-images (extracted from DICOM files) such that no further pre-processing is applied.

The dataset contains a subset of images with bounding boxes for eight
different disease types hand-labeled by a board-certified radiologist. We randomly split these images into 432 samples for validation and 448 samples for testing.

\subsection{Benchmark Task Evaluation}\label{sec:eval_details}
\subsubsection{Task-specific Inference Hyperparameters}
While no task-specific fine-tuning is applied, we use task-specific inference and post-processing hyperparameters for ChEX and all baselines to
consider differences in annotation practices of datasets.
We tune those hyperparameters on samples from the validation splits of the corresponding datasets.

For the sentence grounding (SG) task, we scale all predicted boxes by a constant factor which we optimize on the validation set. Additionally, we apply NMS with an IoU threshold of $0.25$.
For the pathology detection (OD) tasks, we scale boxes independently for each pathology class. We apply NMS with an IoU threshold of $0.25$ for the VinDr-CXR and MS-CXR datasets. For the NIH8 dataset, we merge all predicted boxes per pathology (using the super bounding box covering all predicted boxes) and report the maximum of the predicted score as its box score.

\subsubsection{Metrics and Evaluation}
For SG, we use the sentences from MS-CXR as input and the corresponding bounding boxes as targets. We compute mIoU using the thresholds $[0.1, 0.2, 0.3, 0.4, 0.5]$. For each sentence and threshold, we select the predicted bounding boxes based on the box scores and merge them into a single mask (corresponding to one or more boxes). Similarly, we merge the target boxes of that sentence before computing the IoU between the predicted and target mask. Finally, the mIoU is computed by first averaging the IoU per threshold over all sentences before averaging over all thresholds. For mAP computation, we treat the task as object detection, where each sentence corresponds to its own class, and then use the IoU thresholds $[0.1, 0.2, 0.3, 0.4, 0.5, 0.6, 0.7]$.

For OD, we use the pathology classes provided in the three datasets (VinDR-CXR, MS-CXR, NIH8), where for VinDr-CXR, we only use the top 15 most frequent classes.
We report the standard mAP metric with IoU thresholds $[0.1, 0.2, 0.3, 0.4, 0.5, 0.6, 0.7]$. 

For RC, we provide bounding boxes of regions (pathologies for MS-CXR, anatomical regions for CIG) as input and expect the model to classify these regions.
On MS-CXR this is a multiclass classification problem over 8 classes. Thus, we use the macro-averaged AUROC metric (one-vs-rest). On CIG, this is a multilabel binary classification problem with 53 highly imbalanced classes. Thus, we use the weighted AUROC (wAUROC) metric, which considers the frequency of classes. 

For RE tasks, we provide bounding boxes of regions (pathologies for MS-CXR, anatomical regions for CIG) as input and expect the model to predict textual descriptions for them. On the CIG dataset, regions without associated sentences are ignored, and for regions with multiple sentences, we concatenate their sentences.
For metric computation, we consider each region as its individual sample.

For RG, we only provide the image and expect the model to predict the full report, where each image/report pair is considered a single sample. Baseline results are taken from their publications. Note that some baselines have been evaluated on different test splits and that pre- or post-processing can differ as well, leading to limitations in the interpretation of exact results as also acknowledged by \cite{maira,rgrg}.

Final results are reported based on samples from the test split. We use bootstrapping with $N=250$ for the SG, OD, and RC tasks while using $N=10$ for the RE and RG tasks (due to expansive metric computation).

\subsection{Evaluation of Interactive Capabilities}
We evaluated the interactive capabilities as studied in \cref{sec:precise_prompts,sec:inter_prompting}
on a small subset of the test sets of CIG and MS-CXR.
We use only frontal images that have both CIG and MS-CXR targets and exclude samples where the positive classes present in the MS-CXR targets are not present in the CIG targets, \ie, where CIG and MS-CXR targets are not consistent. This results in a total of 98 images.

We only consider the eight pathology classes from MS-CXR, including their associated bounding boxes. If a pathology is present in an image, as defined by MS-CXR targets, we find the associated anatomical regions based on the scene graph from CIG. Additionally, we include all subregions of the associated regions, \eg, subregions of \say{left lung} include regions like \say{left upper lung}. Given the set of all associated regions and subregions, we further match each bounding box for the given positive pathology with a single region based on the highest IoU. We refer to these regions as \emph{fine regions} and utilize them as \emph{fine hints} in \cref{fig:inter_prompting_quant,fig:precise_prompts_iou}. 
To find the coarse associated regions, we use the parent of each fine region, \eg, the parent of \say{left upper lung} is \say{left lung}. These coarse regions are utilized as \emph{coarse hints} in \cref{fig:inter_prompting_quant,fig:precise_prompts_iou}, 
as \emph{correct hints} in \cref{fig:inter_prompting_neg_quant}, 
and as  \emph{anat prompts} in \cref{fig:precise_prompts_text}.
For the \emph{opposite hint} and \emph{opposite region} in \cref{fig:inter_prompting_neg_quant}, 
we use the anatomical region on the opposite side of the chest, \ie \say{left} is replaced by \say{right}.

For \cref{fig:precise_prompts_text}, 
we apply the CheXbert classifier on predicted sentences and identify the number of positive predicted pathology classes. We refer to the pathology used in the prompt as \emph{queried} and to all other pathologies as \emph{non-queried}.

\subsection{Baseline Details}
\subsubsection{Supervised Visual Grounding Baselines}
We train the supervised visual grounding baselines on the 878 MS-CXR samples corresponding to the MIMIC-CXR training split.
For TransVG\cite{transvg} we use their official implementation\footnote{\url{https://github.com/djiajunustc/TransVG}}. 
We additionally study modified versions of TransVG, where we replace its default image backbone (ResNet50\cite{resnet}) with backbones pre-trained on chest X-ray images (MIMIC-CXR). Specifically, we use the image encoders from BioVIL\cite{biovil} and CheXzero\cite{chexzero}. BioVIL provides a pre-trained ResNet50 backbone, such that the default backbone can be initialized with BioVIL weights without requiring further modifications. CheXzero only provides a  ViT backbone, such that the whole backbone is completely replaced by the CheXzero backbone. Here, we use the patch tokens of the final transformer layer to replace the patch features of the ResNet.

Chen \etal\cite{chen2023_miccai} proposed a supervised visual grounding model for the MS-CXR dataset. However, their model shows only marginal improvements above TransVG, and no public source code is provided, such that we did not include it in our studies. Also, they evaluated their model only on samples with a single bounding box per sentence, such that their results are not comparable with our benchmark.

\subsubsection{Supervised Object Detection Baselines}
We train all supervised object detection baselines on the VinDr-CXR dataset. For evaluation on other OD tasks, we map the VinDr-CXR pathology classes to the corresponding classes of the downstream task. With this strategy, we managed to map all classes of MS-CXR and NIH8, except for pneumonia.

For Faster R-CNN\cite{frcnn} we use the torchvision\footnote{\url{https://pytorch.org/vision}} implementation. For DETR\cite{detr}, Conditional DETR\cite{conditional-detr}, and Deformable DETR\cite{zhu2020deformable} we use the implementations from the Huggingface transformers library\footnote{\url{https://huggingface.co/docs/transformers}}. For $\mathcal{C}$o-Deformable-DETR\cite{codetr} we use the implementation provided by the MMDetection library\footnote{\url{https://mmdetection.readthedocs.io}}. We again study modified versions of these models, where we replace their default backbones with BioVIL\cite{biovil} and CheXzero\cite{chexzero}, following the same approach as for TransVG. Note that we do not study the CheXzero backbone with advanced DETR-style models (Conditional DETR, Deformable DETR, and Deformable DETR) as the standard DETR model with the CheXzero backbone performs very poorly.

We additionally evaluate the Faster R-CNN models on the region classification (RC) tasks. Here, we modify their ROI pooling layer and replace the predicted box proposals with the bounding boxes provided as input. With this modification, their ROI classification heads can be used to classify the given regions. Again, VinDr-CXR classes are mapped to MS-CXR and CIG classes, where we manage to map 7 of 8 MS-CXR classes and 17 of 53 (including many of the most frequent) CIG classes. Note that DETR-based models cannot be utilized for region classification as they do not have an ROI pooling layer.

\subsubsection{Weakly-Supervised Object Detection Baselines}
We also compare against weakly-supervised object detection (WSupOD) models. Such models use only limited supervision without bounding boxes for pathologies during training but can predict pathology bounding boxes during inference. 

We study CheXnet\cite{chexnet}, a model commonly used on chest X-rays. We use our own implementation with the DenseNet121\cite{densenet} backbone and train it on the 91,883 NIH8 training samples using only image-level classification targets. Additionally, we again replace the default backbone with BioVIL\cite{biovil} and CheXzero\cite{chexzero}. For evaluation on other datasets, we map the NIH8 pathologies to the corresponding classes of the downstream task, where we managed to map all MS-CXR classes, 7 of the 15 VinDr-CXR classes used for evaluation, and 10 of the 53 CIG classes.
We also evaluate the CheXnet models on the region classification (RC) tasks by replacing the average-pooling layer before the classifier by ROI pooling with the bounding box provided for the region.

We also study the two models MIL-ADPD\cite{adpd} and Loc-ADPD\cite{adpd}. Both use anatomical region bounding boxes and pathology classification targets, where MIL-ADPD uses image-level classification targets while Loc-ADPD uses region-level targets. We use their official implementation\footnote{\url{https://github.com/philip-mueller/adpd}} and trained it on the 237,917 samples from the CIG training dataset (\ie the CIG samples from the MIMIC-CXR training split). We again replace their default backbone with BioVIL\cite{biovil} but do not study the use of the CheXzero backbone as it performs very poorly on WSupOD with CheXnet. We again mapped pathology classes from CIG to the target task, where we were able to successfully map all classes.

\subsubsection{Contrastive Baselines}
We study two contrastive models pre-trained on the MIMIC-CXR dataset, namely BioVIL\cite{biovil} and CheXzero\cite{chexzero}. We use the official implementations and checkpoints for both BioVIL\footnote{\url{https://github.com/microsoft/hi-ml/tree/main/hi-ml-multimodal}} and CheXzero\footnote{\url{https://github.com/rajpurkarlab/CheXzero}}, where for CheXzero we use only one of their checkpoints\footnote{\texttt{best\_64\_5e-05\_original\_22000\_0.864.pt}} which we retrieved from their provided cloud drive\footnote{\url{https://drive.google.com/drive/folders/1makFLiEMbSleYltaRxw81aBhEDMpVwno}}.

For sentence grounding, we follow the approach of Boecking \etal\cite{biovil}. We first encode the input sentences using their text encoder and the patch features using their image encoder. Next, we project both into the shared image-text space and compute the cosine similarity between the sentence and each patch. The resulting similarity map is interpolated to the image resolution and then thresholded to generate \emph{masks}, which are then used to compute the mIoU with the target bounding boxes. Additionally, we apply the thresholding procedure from CheXnet\cite{chexnet} on the similarity maps to predict bounding boxes. We compute box scores based on the (average) similarity map values overlapping with the bounding box.
The box prediction enables a better comparison with box-based models and allows the computation of both mIoU and mAP scores.

For object detection, we follow a similar approach to predict bounding boxes but use predefined textual prompts for pathologies instead of sentences. 

For region classification, we first compute embeddings for textual prompts of each of the classes. We then compute image features for the given region bounding boxes and compute the cosine similarity between the projected image and text features. For the multiclass task on MS-CXR, we apply softmax over the similarities of all classes, for the multilabel task on CIG, we use positive and negative prompts and apply softmax over the positive and negative similarities for each class independently. We experiment with two approaches to compute the region image features, ROI pooling and cropping.
In the ROI pooling approach, we encode the whole image and then simply pool the patch features overlapping with the given region bounding boxes. For cropping, we feed each region as its individual sample, cropping the image to the given region.

For region explanation, we utilize the image-text retrieval capabilities of the contrastive models. We encode all sentences from the training set independently and store them in a vector database. During inference, we encode each region using the ROI pooling approach from region classification and find the sentence with the highest cosine similarity. 

\subsubsection{Generative Baselines}
We evaluate the RGRG\cite{rgrg} model for region explanation using their official implementation and checkpoint\footnote{\url{https://github.com/ttanida/rgrg}}. For full report generation on MIMIC-CXR, we report the values from the original publications.

\end{document}